%% file: main.tex
\documentclass[10pt]{article}
\usepackage{rsi2style}
\usepackage{xspace}

\newcommand{\oursbase}{\mbox{\novasqname{MetaRSI}}}
\newcommand{\oursplain}{\oursbase\xspace}
\newcommand{\ours}{\mbox{\novasqname{MetaRSI-v1}}\xspace}

\newcommand{\rsi}{\textsc{rsi}\xspace}
\newcommand{\dataop}{Data-RSI\xspace}
\newcommand{\harnessop}{Harness-RSI\xspace}
\newcommand{\modelop}{Model-RSI\xspace}
\newcommand{\rsihar}{\mbox{\novasqname{RSI-Harness}}\xspace}
\newcommand{\rsiag}{\mbox{\novasqname{RSI\textsuperscript{2}}}\ \mbox{\novasqname{Agent-v1}}\xspace}
\newcommand{\rsisub}{\mbox{\novasqname{RSI\textsuperscript{2}}}\ \mbox{\novasqname{Sub-Agent-v1}}\xspace}
\newcommand{\metaag}{\mbox{\novasqname{MetaRSI\textsuperscript{2}}}\ \mbox{\novasqname{Agent-v1}}\xspace}
\newcommand{\transag}{\mbox{\novasqname{Transition Agent-v1}}\xspace}

\newcommand{\opD}{\mathsf{D}}
\newcommand{\opH}{\mathsf{H}}
\newcommand{\opM}{\mathsf{M}}
\newcommand{\Kernel}{\mathcal{K}}
\newcommand{\sig}{\sigma}
\newcommand{\Sys}{S}
\newcommand{\Harn}{\mathcal{H}}
\newcommand{\Dat}{\mathcal{D}}
\newcommand{\Traj}{\tau}
\newcommand{\Budget}{B}
\newcommand{\Qsealed}{Q^{\star}}

\newcommand{\dcell}[1]{\textcolor{OpDataC}{\textbf{#1}}}
\newcommand{\hcell}[1]{\textcolor{OpHarnC}{\textbf{#1}}}
\newcommand{\mcell}[1]{\textcolor{OpModC}{\textbf{#1}}}

\newcommand{\ab}[1]{\textbf{\textcolor{accentdk}{#1}}}
\newcommand{\dhl}[1]{\textcolor{OpDataC}{\textbf{#1}}}
\newcommand{\hhl}[1]{\textcolor{OpHarnC}{\textbf{#1}}}
\newcommand{\mhl}[1]{\textcolor{OpModC}{\textbf{#1}}}
\newcommand{\xcl}{\textbf{\textcolor{OpDataC}{x}}}
\newcommand{\ocl}{\textbf{\textcolor{MetaC}{o}}}

\begin{document}

\reporttitleblock
{\raisebox{-2.90pt}[12.20pt][0pt]{\includegraphics[height=18pt]{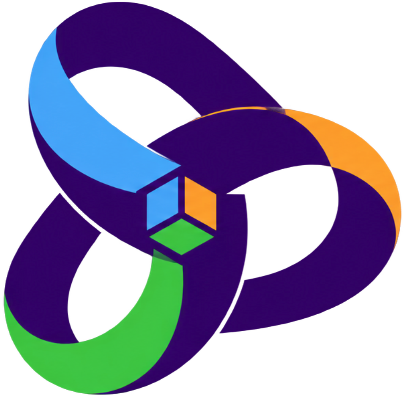}}\hspace{4pt}%
\raisebox{-0.69pt}[12.20pt][0pt]{\includegraphics[height=18pt]{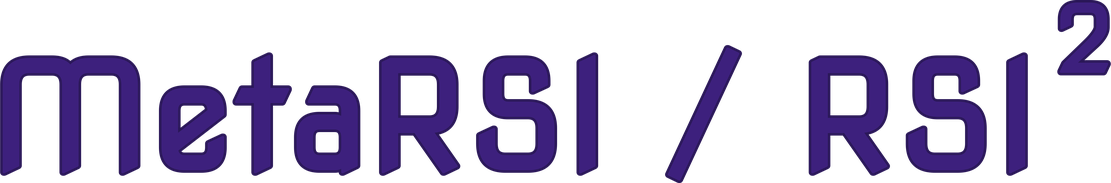}}: A Meta-Recursive Self-Improving\\[-4pt]
System for Recursive Self-Improving Systems Themselves\\[-4pt]
%
{\normalsize\color{accentlt}--- One Kernel, Two Axes, Three Operators, Every Domain}}
%
{Zihan Tan$^{\ast}$\quad Leixin Sun$^{\ast\dagger}$\quad Zitong Shi\quad Yitao Liu\quad
 Jiajun Wu\quad Nathaniel Brooks\quad Jiaru Qian\quad Xiaoran Shang\\[1.6pt]
 Suyuan Huang\quad Yi Ding\quad Yangxu Liao\quad Mukai Li\quad
 Qiushi Sun\quad Shudong Liu\quad Xuankun Rong\quad Xiaohang Yu\\[1.6pt]
 Zhuo Chen\quad Hejia Geng\quad Chenxin Li\quad Aozhou Wang\quad
 Zengji Tu\quad Robert Tang\quad Yuxin Zhan\quad Eric Jiang\\[1.6pt]
 Yuxin Wu\quad Jianqing Zhang\quad Xiao Liang\quad Fang Wu\quad
 Haochi Zhang\quad Alexander Marlow\quad Guancheng Wan$^{\ddagger}$\\[2.2pt]
 {\normalfont\scriptsize\color{black!62}$^{\ast}$Equal contribution\quad
  $^{\dagger}$Project leads\quad $^{\ddagger}$Corresponding author}}
{\instblock}
%
%
{\input{sections/00_abstract}}
%
%
{\input{sections/fig_palette}%
\centering
\linkbutton[figA]{\faGithub}{\rsihar}%
  {https://github.com/CosmosMind-ai/RSI-Harness}\hspace{3mm}%
\linkbutton[figB]{\raisebox{-0.42ex}{\includegraphics[height=1.9ex]%
  {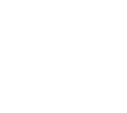}}}{Hugging Face}%
  {https://huggingface.co/CosmosMind/RSI-Harness}\par}
%
%
\vspace{-6pt}
\input{sections/00_teaser_v2}

\input{sections/00_hero}

\input{sections/01_introduction}
\input{sections/02_related_work}
\input{sections/03_preliminaries}
\input{sections/04_method}
\input{sections/05_experiments}
\input{sections/06_discussion}
\input{sections/07_conclusion}

{\small
\bibliography{references}
\bibliographystyle{unsrtnat}}

\newpage
\appendix
\input{sections/99_appendix}

\end{document}

%% file: sections/00_abstract.tex
\looseness=-1 Recursive self-improvement (RSI) lets a system improve the model-building machinery from its own failures, so every later model inherits the gain. Yet RSI has been validated almost exclusively on coding and formal benchmarks such as science QA and mathematics. This \ab{format bound} limits RSI to improvement within a \ab{machine-checkable slice}, not \ab{general capability} where questions are open and correctness is settled by argument, replication, or measurement. We argue RSI must next operate across \ab{real, diverse scientific, engineering, and meta-scientific domains}, not where formal evaluation is merely tractable. To that end we present \ours, where improvement is the \ab{scheduled composition} of \ab{three typed operators} over \ab{one unified paradigm}. \dhl{\dataop} amplifies existing competence and marks its boundary; \hhl{\harnessop} edits a five-slot scaffold without touching weights; \mhl{\modelop} internalizes capability into parameters through bounded training. Sharing one \emph{loop kernel} and artifact vocabulary, they make data, scaffold, and model changes \ab{composable rather than exclusive}. A \ab{two-axis optimizer} jointly decides operator order and each operator's proposal policy, while a \ab{meta-level policy} revises the schedule across terms. We validate \ours under the field's standard evaluations, on code and closed-form science, with \ab{no external teacher}: the target model plays every role in its own loop. \ours reframes self-improvement from a single-surface edit to a composition across the \ab{full model-production pipeline}, opening \ab{two paths}: a \mhl{model route} internalizing capability through training, and a \hhl{harness route} leaving weights untouched and thus extending self-improvement to \ab{any model reachable through an interface}, with \dhl{\dataop} redefined as the \ab{shared substrate feeding both}. The framework further yields \ab{refutable laws} on where loops exist, how operators compose, and what supervision buys.

%% file: sections/fig_palette.tex
\providecommand{\figscheme}{2}
\ifnum\figscheme=2
  \definecolor{figA}{HTML}{35A860}\colorlet{figAink}{white}
  \definecolor{figB}{HTML}{F9A245}\colorlet{figBink}{black!82}
  %
  %
  \definecolor{figC}{HTML}{DFB524}\colorlet{figCink}{black!82}
  \definecolor{figD}{HTML}{C7BA28}\colorlet{figDink}{black!82}
  \definecolor{figE}{HTML}{35A860}\colorlet{figEink}{white}
\else\ifnum\figscheme=3
  \definecolor{figA}{HTML}{FE8841}\colorlet{figAink}{black!82}
  \definecolor{figB}{HTML}{61A1DE}\colorlet{figBink}{black!82}
  \definecolor{figC}{HTML}{B6DAFC}\colorlet{figCink}{black!82}
  \definecolor{figD}{HTML}{FCD7A7}\colorlet{figDink}{black!82}
  \definecolor{figE}{HTML}{FE8841}\colorlet{figEink}{black!82}
\else
  \definecolor{figA}{HTML}{9B60A8}\colorlet{figAink}{white}
  \definecolor{figB}{HTML}{FCE070}\colorlet{figBink}{black!82}
  \definecolor{figC}{HTML}{F6DFFB}\colorlet{figCink}{black!82}
  \definecolor{figD}{HTML}{FDF3CA}\colorlet{figDink}{black!82}
  \definecolor{figE}{HTML}{9B60A8}\colorlet{figEink}{white}
\fi\fi
\colorlet{figAline}{figA!62!black}
\colorlet{figBline}{figB!62!black}
\colorlet{figCline}{figC!62!black}
\colorlet{figDline}{figD!62!black}
\colorlet{figEline}{figE!62!black}
\def\figplate#1#2#3#4#5{%
  \fill[white, rounded corners=1.2pt] (#1,#2) rectangle (#3,#4);
  \draw[rounded corners=1.2pt, fill=fig#5, fill opacity=0.06, text opacity=1,
        draw=fig#5, line width=2.0pt] (#1,#2) rectangle (#3,#4);
  \draw[rounded corners=0.8pt, draw=fig#5, opacity=0.45, line width=0.4pt]
    (#1+0.75,#2+0.75) rectangle (#3-0.75,#4-0.75);}%
%
\providecommand{\figheadfont}{\fontsize{9.4}{9.4}\selectfont\bfseries}
\providecommand{\fignumfont}{\fontsize{6.8}{6.8}\selectfont\bfseries}
\providecommand{\figloopfont}{\fontsize{9.0}{9.0}\selectfont}
%
%
%
%
\def\fighead#1#2#3#4#5#6{%
  \fill[fig#4, rounded corners=1.2pt]
    (#1,#3-6.0) rectangle (#2,#3);
  \fill[fig#4] (#1,#3-6.0) rectangle (#2,#3-2.4);
  \if\relax\detokenize{#5}\relax\else
    \node[circle, fill=fig#4line, inner sep=0pt, minimum size=4.4mm,
          font=\fignumfont, text=white] at (#1+3.6,#3-3.0) {#5};
  \fi
  \node[font=\figheadfont, text=white]
    at ({(#1+#2)/2+3.2},#3-3.0) {\textls[70]{#6}};}%
%
%
%
%
\providecommand{\figcyclefont}{\fontsize{5.2}{5.8}\selectfont}
\def\figcycle#1#2#3#4#5{%
  \begin{scope}[shift={(#1,#2)}]
    \draw[line width=1.2mm, draw=black!14] (36:#3) arc (36:-200:#3);
    \draw[line width=1.2mm, draw=#4!85]    (160:#3) arc (160:44:#3);
    \draw[draw=black!32, line width=0.3pt] (0,0) circle ({#3+0.75mm});
    \draw[draw=black!32, line width=0.3pt] (0,0) circle ({#3-0.75mm});
    \foreach \a in {160,44}
      {\draw[white, line width=0.6pt] (\a:{#3-0.85mm}) -- (\a:{#3+0.85mm});}
    \fill[white] (0,0) circle ({#3-1.05mm});
    \draw[draw=#4!55, line width=0.5pt] (0,0) circle ({#3-1.05mm});
    \if\relax\detokenize{#5}\relax\else
      \node[font=\figcyclefont, text=black!68, align=center] at (0,0) {#5};
    \fi
    \draw[-{Stealth[length=1.7mm,width=1.4mm]}, line width=0.5pt, draw=#4]
      (150:{#3+1.6mm}) arc (150:60:{#3+1.6mm});
  \end{scope}}%
%
%
%
%
\def\metaag{\mbox{\novasqname{MetaRSI\textsuperscript{2}}}\ \mbox{\novasqname{Agent}}\xspace}%
\def\rsiag{\mbox{\novasqname{RSI\textsuperscript{2}}}\ \mbox{\novasqname{Agent}}\xspace}%
\def\rsisub{\mbox{\novasqname{RSI\textsuperscript{2}}}\ \mbox{\novasqname{Sub-Agent}}\xspace}%
 

%% file: sections/00_teaser_v2.tex
\begingroup
\def\figheadfont{\fontsize{9.4}{9.4}\selectfont\bfseries}%
\def\fignumfont{\fontsize{6.8}{6.8}\selectfont\bfseries}%
\input{sections/fig_palette}
\centering
%
\resizebox{0.97\linewidth}{!}{%
\begin{tikzpicture}[
  x=1mm, y=1mm,
  tick/.style={font=\fontsize{6.4}{6.4}\selectfont, text=black!58},
  note/.style={font=\fontsize{7.0}{7.0}\selectfont\itshape, text=black!58},
  ar/.style={-{Stealth[length=1.9mm,width=1.5mm]}, line width=0.9pt},
  edge/.style={-{Stealth[length=2.0mm,width=1.6mm]}, line width=1.2pt},
  elab/.style={circle, fill=white, draw=black!25, line width=0.4pt,
               inner sep=0.7pt, font=\fontsize{5.4}{5.4}\selectfont\bfseries,
               text=black!58},
  opc/.style={circle, draw=#1, fill=#1!14, line width=0.9pt, minimum size=6.4mm,
              inner sep=0pt, font=\fontsize{7.0}{7.0}\selectfont\bfseries,
              text=#1!80!black},
  wmcell/.style={rounded corners=0.6pt, minimum width=3.6mm,
                 minimum height=3.2mm, inner sep=0pt, line width=0.4pt,
                 font=\fontsize{5.2}{5.2}\selectfont}
]
%
\begin{scope}[on background layer]
  \node[anchor=south west, inner sep=0] at (0,2.6)
    {\includegraphics[width=212mm, height=77.4mm]{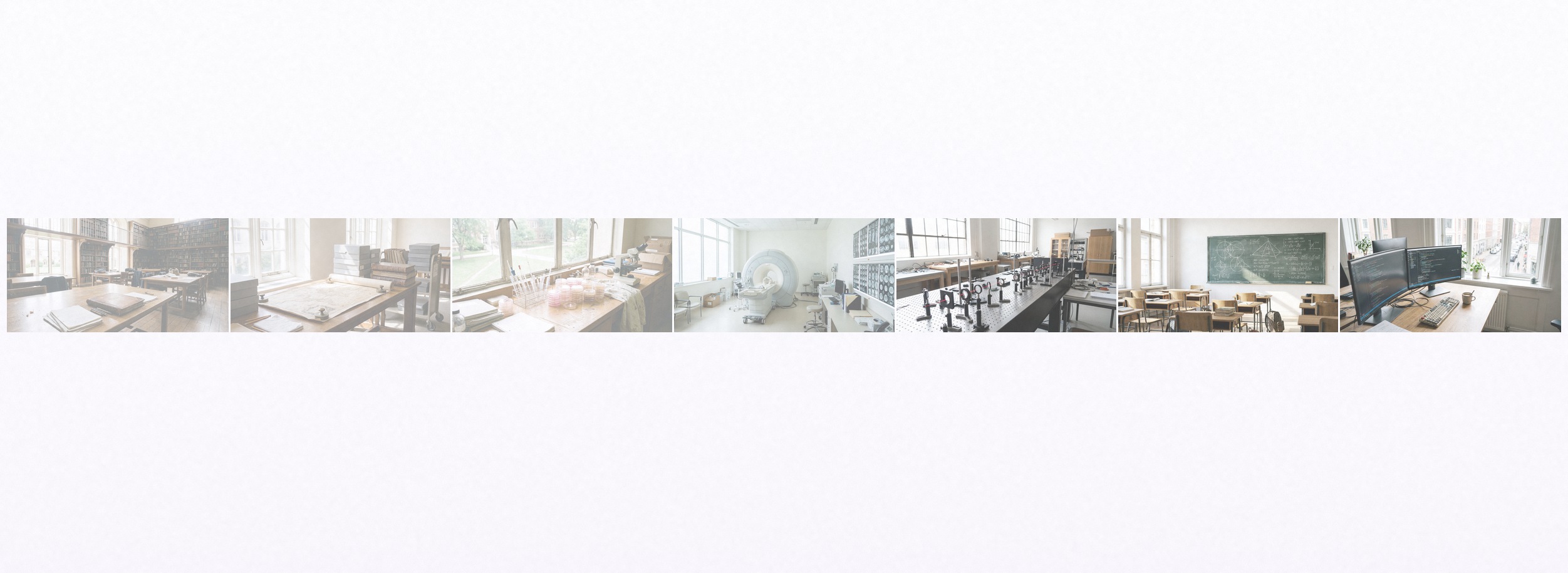}};
\end{scope}
\path (0,2.6) rectangle (212,80);

\figplate{1}{53.5}{69.3}{79}{A}
\fighead{1}{69.3}{79}{A}{1}{WHERE LOOPS CLOSE}

%
\newcommand{\px}[1]{10+(#1)*0.54}
\newcommand{\py}[1]{58.6+((#1)-5)*0.101}
\fill[MetaC!13] ({\px{0}},{\py{60}}) rectangle ({\px{45}},{\py{100}});
\draw[draw=MetaC!60, line width=0.5pt, dash pattern=on 1.1pt off 1.0pt]
  ({\px{0}},{\py{60}}) rectangle ({\px{45}},{\py{100}});
\draw[draw=black!40, line width=0.6pt]
  ({\px{0}},{\py{5}}) -- ({\px{100}},{\py{5}});
\draw[draw=black!40, line width=0.6pt]
  ({\px{0}},{\py{5}}) -- ({\px{0}},{\py{100}});
\draw[draw=figAline, line width=1.0pt, dash pattern=on 1.5pt off 1.1pt]
  ({\px{45}},{\py{5}}) -- ({\px{45}},{\py{100}});
%
%
\foreach \v/\c/\l in {%
  95/97/1, 90/97/1, 90/96.1/1, 85/95.5/1, 85/84.6/1, 70/83/1,
  60/64.4/1, 40/52.8/1, 30/67.8/1, 15/12.8/1,
  50/54/0, 40/83/0, 35/71/0, 20/25.8/0, 15/59.2/0,
  15/76/0, 10/35/0, 10/68/0}
  {\ifnum\l=1
     \fill[figAline] ({\px{\v}},{\py{\c}}) circle (0.92mm);
     \draw[draw=figAline!60!black, line width=0.3pt]
       ({\px{\v}},{\py{\c}}) circle (0.92mm);
   \else
     \fill[white] ({\px{\v}},{\py{\c}}) circle (0.92mm);
     \draw[draw=figAline!65, line width=0.7pt]
       ({\px{\v}},{\py{\c}}) circle (0.92mm);
   \fi}
\node[font=\fontsize{6.6}{6.6}\selectfont\bfseries, text=MetaC!85!black,
%
      anchor=north east] at ({\px{45}},{\py{100}+3.8})
  {5 fields, no key};
\node[tick, rotate=90] at ({\px{0}-2.4},{\py{52}}) {capability};
\fill[figAline] (11.4,56.7) circle (0.92mm);
\node[tick, anchor=west] at (13.1,56.7) {\strut closed};
\draw[draw=figAline!65, line width=0.7pt] (23.4,56.7) circle (0.92mm);
\node[tick, anchor=west] at (25.1,56.7) {\strut never};
\node[tick, anchor=east] at ({\px{100}},56.7)
  {\strut a machine can check it $\rightarrow$};

\figplate{71.8}{53.5}{140.2}{79}{B}
\fighead{71.8}{140.2}{79}{B}{2}{ONE KERNEL, THREE OPERATORS}

%
\begin{scope}[shift={(86.6,62.7)}]
  \draw[line width=2.0mm, draw=black!12]     (128:6.0mm) arc (128:-232:6.0mm);
  \draw[line width=2.0mm, draw=figBline!55]     (128:6.0mm) arc (128:46:6.0mm);
  \draw[draw=black!40, line width=0.35pt] (0,0) circle (7.1mm);
  \draw[draw=black!40, line width=0.35pt] (0,0) circle (4.9mm);
  \foreach \a in {128,46}
    {\draw[white, line width=0.7pt] (\a:4.7mm) -- (\a:7.3mm);}
  \draw[-{Stealth[length=1.6mm,width=1.3mm]}, line width=0.8pt,
        draw=figBline!75!black] (95:6.0mm) arc (95:70:6.0mm);
  \foreach \a in {-14,-134}
    {\draw[-{Stealth[length=1.6mm,width=1.3mm]}, line width=0.8pt,
           draw=black!42] (\a:6.0mm) arc (\a:\a-16:6.0mm);}
  \node[circle, fill=white, draw=figBline!65, line width=0.7pt,
        minimum size=9.4mm] at (0,0) {};
  \node[font=\fontsize{5.2}{5.6}\selectfont\bfseries, text=black!70,
        align=center] at (0,1.2mm) {one\\kernel};
  \node[font=\fontsize{5.0}{5.0}\selectfont, align=center] at (0,-2.2mm)
    {\textcolor{OpDataC}{$\Dat$}\,\textbar\,\textcolor{OpHarnC}{$\Harn$}%
     \,\textbar\,\textcolor{OpModC}{$\theta$}};
\end{scope}

\foreach \i/\s/\hue in {0/{$\Dat$}/OpDataC, 1/{$\Harn$}/OpHarnC,
                        2/{$\theta$}/OpModC}
  {\node[font=\fontsize{6.0}{6.0}\selectfont, text=\hue!80!black]
     at (101.8+\i*4.0,71.5) {\s};}
\foreach \r/\hue/\nm/\on in {%
  0/OpDataC/{\dataop}/0, 1/OpHarnC/{\harnessop}/1, 2/OpModC/{\modelop}/2}
  {\pgfmathsetmacro{\ry}{67.0 - \r*4.7}
   \foreach \i in {0,1,2}
     {\ifnum\i=\on
        \node[wmcell, fill=\hue, draw=\hue!65!black, text=white]
          at (101.8+\i*4.0,\ry) {\faCheck};
      \else
        \node[wmcell, fill=black!4, draw=black!25, text=black!28]
          at (101.8+\i*4.0,\ry) {--};
      \fi}
   \node[anchor=west, font=\fontsize{6.6}{6.6}\selectfont\bfseries, text=\hue]
     at (111.4,\ry) {\nm};}

\figplate{142.7}{53.5}{211}{79}{C}
\fighead{142.7}{211}{79}{C}{3}{FIVE OF SIX ADMISSIBLE}

\node[opc=OpDataC] (gD) at (157.0,68.4) {$\opD$};
\node[opc=OpHarnC] (gH) at (197.0,68.4) {$\opH$};
\node[opc=OpModC]  (gM) at (177.0,61.4) {$\opM$};
\draw[edge, draw=OpDataC!85] (gD) to[bend left=9]  node[elab,pos=0.5]{1} (gH);
\draw[edge, draw=OpHarnC!85] (gH) to[bend left=9]  node[elab,pos=0.5]{3} (gD);
\draw[edge, draw=OpDataC!85] (gD) to[bend right=10] node[elab,pos=0.5]{2} (gM);
\draw[edge, draw=OpModC!85]  (gM) to[bend right=10] node[elab,pos=0.5]{4} (gD);
\draw[edge, draw=OpModC!85]  (gM) to[bend right=10] node[elab,pos=0.5]{5} (gH);
\draw[edge, draw=MetaC, dash pattern=on 1.4pt off 1.1pt, line width=1.1pt]
  (gH) to[bend right=20] node[elab,pos=0.3,text=MetaC]{\faTimes} (gM);
\draw[edge, draw=OpDataC!85] (gD) to[out=172,in=224,looseness=5.5] (gD);
\draw[edge, draw=OpHarnC!85] (gH) to[out=8,in=-44,looseness=5.5] (gH);
\node[font=\fontsize{6.6}{6.6}\selectfont, text=MetaC!85!black, anchor=north]
  at (177.0,58.7)
  {\textcolor{MetaC}{$\opH\!\to\!\opM$ forbidden}: the dataset predates the
   change};

\foreach \x/\lab in {%
  16/{philosophy}, 46/{history}, 76/{wet-lab biology},
  106/{clinical medicine}, 136/{optical physics}, 166/{mathematics},
  196/{software}}
  {\node[font=\fontsize{6.4}{6.4}\selectfont, text=black!66] at (\x,32.5)
     {\lab};}
\draw[draw=figAline, line width=1.2pt, dash pattern=on 1.6pt off 1.2pt]
  (121.0,30.7) -- (121.0,50.3);
\node[rounded corners=1.0pt, fill=white, fill opacity=0.94, text opacity=1,
      draw=figAline!70, line width=0.5pt, inner xsep=2.2pt, inner ysep=1.2pt,
      font=\fontsize{6.6}{6.6}\selectfont\bfseries, text=figAline, anchor=north]
  at (121.0,49.6) {the affordance line};
%
\node[rounded corners=1.2pt, fill=white, fill opacity=0.90, text opacity=1,
      draw=figEline!45, line width=0.6pt, inner xsep=2.4pt, inner ysep=1.8pt,
      anchor=north west, align=left,
      font=\fontsize{7.4}{8.4}\selectfont\bfseries, text=figEline]
  at (2.2,49.6) {verification\\unavailable};
\node[rounded corners=1.2pt, fill=white, fill opacity=0.90, text opacity=1,
      draw=figEline!45, line width=0.6pt, inner xsep=2.4pt, inner ysep=1.8pt,
      anchor=north east, align=center,
      font=\fontsize{7.4}{8.4}\selectfont\bfseries, text=figEline]
  at (209.8,49.6) {verification\\free};

\begin{scope}[shift={(0,1.6)}]

\figplate{1}{2}{89.5}{27.5}{D}
\fighead{1}{89.5}{27.5}{D}{4}{THREE NESTED LEVELS}

\foreach \r/\bg/\ln/\tk/\ind/\tag/\txt/\cad in {%
  0/{figDline!16}/{figDline}/{figDline!90!black}/0/{L3}/%
    {\metaag: revises the scheduler}/{once a term},
  1/{figDline!11}/{figDline!78}/{figDline!80!black}/3.6/{L2}/%
    {\rsiag: picks one axis per step}/{once a step},
  2/{figDline!7}/{figDline!56}/{figDline!70!black}/7.2/{L1}/%
    {The Kernel: runs one cycle}/{once a cycle}}
  {\pgfmathsetmacro{\by}{16.9 - \r*4.6}
   \draw[rounded corners=1.0pt, fill=\bg, draw=\ln, line width=0.9pt]
     ({4.2+\ind},\by) rectangle (86.3,{\by+3.6});
   \node[anchor=west, font=\fontsize{7.0}{7.0}\selectfont\bfseries,
         text=\tk] at ({5.4+\ind},{\by+1.8}) {\tag};
   \node[anchor=west, font=\fontsize{7.0}{7.0}\selectfont, text=black!72]
     at ({12.2+\ind},{\by+1.8}) {\txt};
   \node[anchor=east, font=\fontsize{6.4}{6.4}\selectfont\itshape,
         text=\tk] at (85.1,{\by+1.8}) {\cad};}
\foreach \r/\ind in {0/0, 1/3.6}
  {\pgfmathsetmacro{\by}{16.9 - \r*4.6}
   \draw[draw=black!42, line width=0.6pt, rounded corners=0.8pt]
     ({5.9+\ind},\by) -- ({5.9+\ind},{\by-2.8}) -- ({7.9+\ind},{\by-2.8});}
\node[rounded corners=1.0pt, fill=black!6, draw=black!45, line width=0.9pt,
      inner xsep=2.6pt, inner ysep=1.3pt, anchor=west,
      font=\fontsize{7.0}{7.0}\selectfont, text=black!70] at (4.2,5.4)
  {\faLock\; sealed evaluator, \textbf{outside all three}};

\figplate{92}{2}{211}{27.5}{E}
\fighead{92}{211}{27.5}{E}{5}{THE TWO AXES IT SCHEDULES ON}

\node[anchor=west, font=\fontsize{7.8}{7.8}\selectfont\bfseries, text=figEline]
  at (95.0,19.6) {\textls[50]{HORIZONTAL}};
\node[note, anchor=west] at (95.0,16.7) {the order of application};
\foreach \i/\lab/\hue in {0/{$\opD$}/OpDataC, 1/{$\opM$}/OpModC,
                          2/{$\opH$}/OpHarnC}
  {\node[circle, draw=\hue, fill=\hue!14, line width=0.8pt, minimum size=5.8mm,
         inner sep=0pt, font=\fontsize{6.6}{6.6}\selectfont\bfseries,
         text=\hue!80!black] at (98.4+\i*10.4,9.3) {\lab};}
\node[circle, draw=figEline!70, fill=white, line width=0.8pt, minimum size=5.8mm,
      inner sep=0pt, font=\fontsize{6.6}{6.6}\selectfont\bfseries, text=figEline,
      dash pattern=on 1.1pt off 0.9pt] at (129.6,9.3) {?};
\foreach \a in {0,1,2}
  {\pgfmathsetmacro{\xa}{101.7 + \a*10.4}
   \draw[ar, draw=figEline!70] (\xa,9.3) -- (\xa+4.2,9.3);}
\node[note, anchor=west, align=left] at (134.0,9.3)
  {the sequence\\\textbf{extends}};

\draw[draw=figEline!35, line width=0.6pt] (160.0,4.6) -- (160.0,20.6);

\node[anchor=west, font=\fontsize{7.8}{7.8}\selectfont\bfseries, text=figEline]
  at (163.0,19.6) {\textls[50]{VERTICAL}};
\node[note, anchor=west] at (163.0,16.7) {each operator's own policy};
%
\figcycle{170.0}{9.3}{3.2mm}{figEline}{}
\node[note, anchor=west, align=left] at (176.4,9.3)
  {revises how it proposes,\\from the feedback it got};
\end{scope}
\end{tikzpicture}}
%
%
\setlength{\abovecaptionskip}{3pt}%
%
\captionof{figure}{\textbf{\ours{} at a glance.} \emph{1}: loops close where verification is
machine-checkable. \emph{2}: one kernel, three operators, \textbf{one writable surface each}.
\emph{3}: five of six orderings admissible. \emph{4}: three nested improvement levels. \emph{5}:
horizontal composition, vertical policy rewriting. The band runs from freely verifiable to
unverifiable.}
\label{fig:teaser}
\par\endgroup

%% file: sections/00_hero.tex
\input{sections/hero/plate}
{\captionof{figure}{\textbf{Overview of \ours.} \emph{1}: the binding constraint is
\textbf{verification format, not subject domain}. \emph{2}: one loop kernel, instantiated by every
operator and scheduled across three nested levels: the kernel cycle, the \rsiag, and the \metaag.
\emph{3}: three operators with \textbf{disjoint write surfaces}, \dhl{\dataop} writing $\Dat$,
\hhl{\harnessop} writing $\Harn$ and \mhl{\modelop} writing $\theta$, composed \emph{horizontally}
by application order or optimized \emph{vertically} by rewriting an operator's proposal policy. The
central band traces one improvement term from learning signal through scheduled composition to
sealed evaluation and release, and the recompilation arc returns to the signal after each
capability-altering step. The sealed evaluator is outside all three.}
\label{fig:hero}}

%% file: sections/hero/plate.tex
\begingroup
\providecommand{\heroground}{sheetcolour}
\def\figheadfont{\fontsize{7.7}{7.7}\selectfont\bfseries}%
\def\fignumfont{\fontsize{5.6}{5.6}\selectfont\bfseries}%
\def\figcyclefont{\fontsize{4.2}{4.6}\selectfont}%
\input{sections/fig_palette}%
\centering
\resizebox{\linewidth}{!}{%
\begin{tikzpicture}[
  x=1mm, y=1mm,
  claim/.style={font=\fontsize{7.4}{9.0}\selectfont\bfseries, text=black!82,
                align=left, anchor=north west},
  supp/.style={font=\fontsize{6.0}{7.2}\selectfont, text=black!58,
               align=left, anchor=north west},
  body/.style={font=\fontsize{6.6}{8.0}\selectfont, text=black!74,
               align=left, anchor=north west},
  note/.style={font=\fontsize{6.0}{6.0}\selectfont\itshape, text=black!56},
  gnode/.style={rounded corners=2.4pt, fill=white, fill opacity=0.94,
                text opacity=1, draw=#1!70, line width=1.2pt,
                align=center, inner xsep=3.4pt, inner ysep=2.8pt},
  flow/.style={-{Stealth[length=2.4mm,width=2.0mm]}, line width=1.1pt,
               dash pattern=on 2.0pt off 1.6pt, draw=figEline!55},
  chip/.style={rounded corners=2.0pt, fill=#1!10, draw=#1!60, line width=0.7pt,
               inner xsep=3.0pt, inner ysep=1.8pt,
               font=\fontsize{6.0}{6.0}\selectfont, text=#1!80!black},
  wm/.style={rounded corners=1.0pt, minimum width=5.6mm, minimum height=5.6mm,
             inner sep=0pt, line width=0.5pt,
             font=\fontsize{6.2}{6.2}\selectfont},
  band/.style={line width=2.2mm}
]
\node[anchor=south west, inner sep=0] at (0,0)
  {\includegraphics[width=176.9mm]{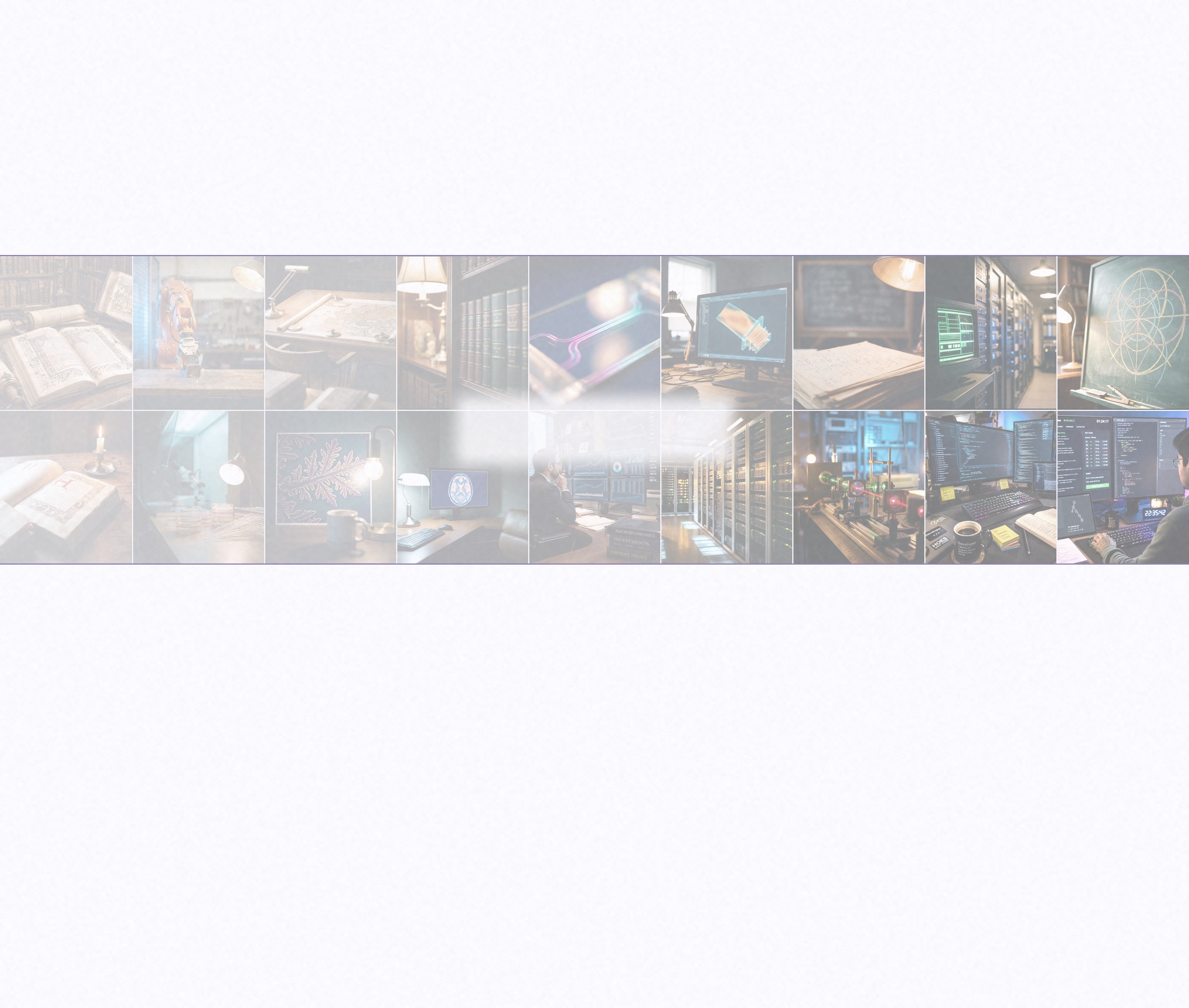}};
\path (0,0) rectangle (176.9,150);

\figplate{1}{115}{87.2}{149}{C}
\fighead{1}{87.2}{149}{C}{1}{WHAT RSI HAS TO BECOME}

\foreach \yy/\cl/\sp in {%
  139.4/{It has only closed where checking was free.}/%
        {seven in ten improve against a machine-checkable target},
  131.4/{The constraint is \emph{format}, not subject.}/%
        {and only three in ten work on code at all},
  123.4/{So build the machinery that builds loops.}/%
        {One Kernel, Two Axes, Three Operators}}
  {\fill[figCline!75, rounded corners=0.6pt] (3.6,\yy-5.6) rectangle (4.8,\yy+0.4);
   \node[claim, text width=80mm] at (6.6,\yy+0.8) {\cl};
   \node[supp,  text width=80mm] at (6.6,\yy-3.4) {\sp};}

\figplate{89.7}{115}{176}{149}{B}
\fighead{89.7}{176}{149}{B}{2}{ONE KERNEL, THREE LEVELS}

\begin{scope}[shift={(103.4,129.0)}]
  \draw[band, draw=black!10]      (128:7.7mm) arc (128:-232:7.7mm);
  \draw[band, draw=figBline!65]       (128:7.7mm) arc (128:46:7.7mm);
  \draw[draw=black!40, line width=0.4pt] (0,0) circle (9.0mm);
  \draw[draw=black!40, line width=0.4pt] (0,0) circle (6.4mm);
  \foreach \a in {128,46}{\draw[white, line width=0.8pt] (\a:6.2mm) -- (\a:9.2mm);}
  \node[circle, fill=white, draw=figBline!65, line width=0.8pt,
        minimum size=12.4mm] at (0,0) {};
  \node[font=\fontsize{6.6}{7.2}\selectfont\bfseries, text=black!70,
        align=center] at (0,1.3mm) {system\\state};
  \node[font=\fontsize{6.0}{6.0}\selectfont, align=center] at (0,-3.1mm)
    {\textcolor{OpDataC}{$\Dat$}\,\textbar\,\textcolor{OpHarnC}{$\Harn$}%
     \,\textbar\,\textcolor{OpModC}{$\theta$}};
\end{scope}

%
%
\foreach \yy/\lvl/\tint/\txt in {%
  135.0/{LEVEL 3}/{figBline}/{\textbf{\metaag} revises the scheduler},
  129.0/{LEVEL 2}/{figBline!78}/{\textbf{\rsiag} picks one axis per step},
  123.0/{LEVEL 1}/{figBline!56}/{\textbf{The Kernel} runs one cycle}}
%
  {\node[rounded corners=1.6pt, fill=\tint, inner xsep=2.2pt, inner ysep=1.1pt,
         font=\fontsize{5.8}{5.8}\selectfont\bfseries, text=white, anchor=west]
     at (113.6,\yy) {\lvl};
   \node[body, text width=46mm, anchor=west] at (128.0,\yy) {\txt};}
\node[gnode=figCline, text width=34mm] (n1) at (24,102)
  {\textcolor{black!80}{\textbf{\footnotesize$\sig$\ \ learning signal}}\\[0.6pt]
   {\fontsize{5.8}{6.4}\selectfont\color{black!60}%
    \textcolor{OpDataC}{$\kappa_{\mathrm{k}}$}\,\textcolor{black!40}{$\cdot$}\,%
    \textcolor{OpHarnC}{$\kappa_{\mathrm{r}}$}\,\textcolor{black!40}{$\cdot$}\,%
    \textcolor{mTeal!85!black}{$\kappa_{\mathrm{v}}$}\,\textcolor{black!40}{$\cdot$}\,%
    \textcolor{accentlt}{$\kappa_{\mathrm{d}}$}\ \ four dimensions}};
\node[gnode=figBline, text width=34mm] (n2) at (68,102)
  {\textcolor{black!80}{\textbf{\footnotesize scheduled composition}}\\[0.6pt]
   {\fontsize{5.8}{6.4}\selectfont\color{black!60}%
    \textcolor{OpDataC}{$\opD$}\,\textcolor{black!40}{$\cdot$}\,%
    \textcolor{OpHarnC}{$\opH$}\,\textcolor{black!40}{$\cdot$}\,%
    \textcolor{OpModC}{$\opM$}\ \ five of six admissible}};
\node[gnode=figEline, text width=34mm] (n3) at (112,102)
  {\textcolor{black!80}{\textbf{\footnotesize\faLock\ sealed evaluator}}\\[0.6pt]
   {\fontsize{5.8}{6.4}\selectfont\color{black!60}opened once, after freezing}};
\node[gnode=figDline, text width=34mm] (n4) at (156,102)
  {\textcolor{black!80}{\textbf{\footnotesize\faCheckCircle\ released
    successor}}\\[0.6pt]
   {\fontsize{5.8}{6.4}\selectfont\color{black!60}one per term, or none}};
\foreach \u/\v in {n1/n2, n2/n3, n3/n4} {\draw[flow] (\u.east) -- (\v.west);}

%
\node[rounded corners=2.0pt, fill=white, fill opacity=0.94, text opacity=1,
      draw=figDline!55, line width=0.6pt, inner xsep=3.0pt, inner ysep=1.8pt,
      font=\fontsize{6.0}{6.0}\selectfont, text=black!72] (rc) at (88.45,74)
  {the signal is \textbf{recompiled} after every capability-altering step};
\draw[-{Stealth[length=2.6mm,width=2.1mm]}, line width=1.1pt,
      dash pattern=on 2.2pt off 1.8pt, draw=figDline!80, rounded corners=2.5pt]
  (n4.south) -- (156,74) -- (rc.east);
\draw[-{Stealth[length=2.6mm,width=2.1mm]}, line width=1.1pt,
      dash pattern=on 2.2pt off 1.8pt, draw=figDline!80, rounded corners=2.5pt]
  (rc.west) -- (24,74) -- (n1.south);

\node[anchor=base, font=\fontsize{25}{25}\selectfont\bfseries] at (88.45,82.4)
  {\textcolor{figA!80!black}{\novasqname{Meta}}%
   \textcolor{figB!80!black}{\novasqname{RSI}}};

\figplate{1}{2}{176}{63}{E}
\fighead{1}{176}{63}{E}{3}{THREE OPERATORS OVER ONE KERNEL, AND THE TWO AXES
  THAT SCHEDULE THEM}

\newcommand{\wmrow}[4]{%
  \foreach \i/\lab/\hue/\on in {0/{$\Dat$}/OpDataC/#2, 1/{$\Harn$}/OpHarnC/#3,
                                2/{$\theta$}/OpModC/#4}{%
    \pgfmathsetmacro{\cx}{#1 + \i*6.4}
    \ifnum\on=1
      \node[wm, fill=\hue, draw=\hue!65!black, text=white] at (\cx,51.4) {\lab};
    \else
      \node[wm, fill=black!4, draw=black!28, text=black!32] at (\cx,51.4) {\lab};
    \fi}}

\foreach \colL/\colR/\hue/\opnm/\mA/\mB/\mC in {%
  4/57.3/OpDataC/{\dataop}/1/0/0,
  61.3/114.6/OpHarnC/{\harnessop}/0/1/0,
  118.6/171.9/OpModC/{\modelop}/0/0/1}
  {\node[anchor=west, font=\fontsize{10.4}{10.4}\selectfont\bfseries, text=\hue]
     at (\colL+1.5,51.4) {\opnm};
   \wmrow{\colR-18.7}{\mA}{\mB}{\mC}
   \draw[draw=\hue!40, line width=0.6pt] (\colL+1.5,47.0) -- (\colR-1.5,47.0);}

\foreach \colL/\hue/\wtx/\bodytx/\chiptx in {%
  4/OpDataC/{$\Dat$}/%
    {Presses out what the model already holds, and marks where its
     boundary is.}/%
    {then \harnessop\ or \modelop},
  61.3/OpHarnC/{$\Harn$}/%
    {Five pluggable slots: prompt, memory, built-in tools, skills
     and MCP.}/%
    {then \dataop\ only},
  118.6/OpModC/{$\theta$}/%
    {Internalizes what the scaffold carried, so the harness can then be
     \emph{simplified}.}/%
    {then \dataop\ or \harnessop}}
  {\node[anchor=north west, font=\fontsize{6.0}{6.0}\selectfont,
         text=\hue!72!black] at (\colL+1.5,45.4) {writes \wtx\ only};
   \node[body, text width=51mm] at (\colL+1.5,41.6) {\bodytx};
   \node[chip=\hue, anchor=west] at (\colL+1.5,30.6) {\chiptx};}

\draw[draw=figEline!30, line width=0.6pt] (4,27.0) -- (171.9,27.0);
\draw[draw=figEline!25, line width=0.6pt] (88,5.0) -- (88,24.6);

\node[anchor=west, font=\fontsize{7.2}{7.2}\selectfont\bfseries, text=figEline]
  at (5.5,23.0) {\textls[110]{HORIZONTAL}\ \ the order of application};
\node[body, text width=78mm] at (5.5,21.0)
  {The \rsiag chooses which operator runs next.};
\foreach \x/\lab/\hue in {13.5/{$\opD$}/OpDataC, 26.5/{$\opM$}/OpModC,
                          39.5/{$\opH$}/OpHarnC}
  {\node[circle, draw=\hue, fill=\hue!12, line width=1.0pt, minimum size=8.4mm,
         inner sep=0pt, font=\fontsize{7.6}{7.6}\selectfont\bfseries,
         text=\hue!80!black] at (\x,11.0) {\lab};}
%
\node[circle, draw=figEline!60, fill=white, line width=1.0pt, minimum size=8.4mm,
      inner sep=0pt, font=\fontsize{7.6}{7.6}\selectfont\bfseries, text=figEline,
      dash pattern=on 1.2pt off 1.0pt] at (52.5,11.0) {?};
\foreach \a/\b in {18.2/21.8, 31.2/34.8, 44.2/47.8}
  {\draw[-{Stealth[length=2.0mm,width=1.6mm]}, draw=figEline!60, line width=1.0pt]
     (\a,11.0) -- (\b,11.0);}
\node[note, anchor=west] at (58.5,11.0) {the sequence is extended};

\node[anchor=west, font=\fontsize{7.2}{7.2}\selectfont\bfseries, text=figEline]
  at (90.5,23.0) {\textls[110]{VERTICAL}\ \ each operator's own policy};
\node[body, text width=78mm] at (90.5,21.0)
  {Instead it dispatches to one operator's own \rsisub.};
%
\figcycle{99.0}{10.6}{4.4mm}{figEline}{how it\\proposes}
\node[note, anchor=west, align=left] at (107.0,10.6)
  {rewritten in place;\\the sequence is untouched};
\end{tikzpicture}}
\par\endgroup

%% file: sections/01_introduction.tex
\section{Introduction}
\label{sec:intro}

\begin{figure*}[!t]
\centering
%
%
%
\input{sections/fig_palette}%
\resizebox{\linewidth}{!}{%
\begin{tikzpicture}[
  x=1mm, y=1mm,
  font=\small,
  cardfont/.style={font=\fontsize{8.4}{9.6}\selectfont},
  card/.style={rounded corners=2.5pt, draw=black!48, line width=0.8pt,
               fill=white, align=center, inner xsep=3pt, inner ysep=2.6pt,
               cardfont},
  cost/.style={rounded corners=2.5pt, draw=MetaC!75, line width=0.7pt,
               fill=MetaC!5, align=center, inner xsep=3pt, inner ysep=2.2pt,
               cardfont, text=MetaC!85!black},
  st/.style={rounded corners=2.5pt, draw=black!48, line width=0.8pt,
             fill=white, align=center, inner xsep=2pt, inner ysep=2.6pt,
             font=\scriptsize},
  ar/.style={-{Stealth[length=2.0mm,width=1.6mm]}, line width=1.0pt,
             draw=black!38},
  axis/.style={-{Stealth[length=2.4mm,width=1.9mm]}, draw=black!42,
               line width=0.8pt},
  onlab/.style={rounded corners=2pt, fill=white, draw=figCline!55,
                line width=0.5pt, inner xsep=2.6pt, inner ysep=1.5pt,
                font=\scriptsize, text=figCline}
]
\path (0,0) rectangle (212,104);

%

%
\newcommand{\bandhead}[6]{%
  \fill[fig#4, rounded corners=3pt] (#1,#3-6.8) rectangle (#2,#3);
  \fill[fig#4] (#1,#3-6.8) rectangle (#2,#3-3.0);
  \node[circle, fill=fig#4line, inner sep=0pt, minimum size=5.0mm,
        font=\scriptsize\bfseries, text=white] at (#1+4.2,#3-3.4) {#5};
  \node[font=\fontsize{9.2}{9.2}\selectfont\bfseries, text=white]
    at ({(#1+#2)/2+4},#3-3.4) {#6};}
\newcommand{\subgroup}[5]{%
  \draw[rounded corners=3pt, draw=#5!55, line width=0.6pt, fill=#5!4,
        dash pattern=on 1.8pt off 1.4pt] (#1,#2) rectangle (#3,#4);}
\newcommand{\blockarrow}[3]{%
  \fill[#3!45, draw=#3!70, line width=0.5pt]
    (#1,#2-1.9) -- (#1+3.9,#2-1.9) -- (#1+3.9,#2-3.6) -- (#1+8.0,#2)
    -- (#1+3.9,#2+3.6) -- (#1+3.9,#2+1.9) -- (#1,#2+1.9) -- cycle;}

\begin{scope}[on background layer]
  \bandfill{figA}{(1,15)}{(38,102)}
\end{scope}
\bandhead{1}{38}{102}{A}{I}{\;THE EVIDENCE\;}

\node[card, text width=30mm] (run) at (19.5,87)
  {a deployed system\\runs a term};
\node[card, text width=30mm] (traj) at (19.5,74)
  {what it did, and how it was judged};
\node[card, text width=30mm, draw=MetaC, fill=MetaC!8,
      text=MetaC!85!black] (nolab) at (19.5,59.5)
  {\textbf{one deficiency}, and no label saying which surface};
\node[card, fill=figAline, draw=none, text=white, text width=30mm,
      font=\fontsize{8.4}{8.4}\selectfont\bfseries] (ev) at (19.5,45)
  {THE SAME EVIDENCE};
\foreach \a/\b in {run/traj, traj/nolab, nolab/ev} {\draw[ar] (\a) -- (\b);}
\node[font=\fontsize{8.4}{9.6}\selectfont, text=black!48, align=center,
      text width=31mm]
  at (19.5,33)
  {the same whether the work is code, a proof, a protocol or a diagnosis};

\blockarrow{38.3}{58}{figBline}

\begin{scope}[on background layer]
  \bandfill{figB}{(46.6,15)}{(128,102)}
\end{scope}
\bandhead{46.6}{128}{102}{B}{II}%
  {\;THREE SINGLE-SURFACE READINGS\;}

\newcommand{\oplane}[5]{%
  %
  \node[font=\fontsize{8.4}{8.4}\selectfont\bfseries, text=#2!72!black,
        anchor=north west]
    (t#1) at (49.0,#1+12.7) {#3};
  \node[card, text width=17mm, draw=#2!70, fill=#2!7, anchor=north]
    (a#1) at (61,#1+6.4) {reads the evidence};
  \node[card, text width=48mm, draw=#2!70, fill=#2!7, anchor=north]
    (b#1) at (99,#1+6.4) {#4};
  \draw[ar, draw=#2!85] (a#1) -- (b#1);
  \draw[ar, draw=#2!85, rounded corners=1.6pt]
    (b#1.north) -- ++(0,1.9) -- ($(a#1.north)+(0,1.9)$) -- (a#1.north);
  \node[cost, text width=74mm, anchor=north] (c#1)
    at ($(b#1.south)+(-12,-1.8)$) {#5};
  \node[inner sep=0pt] (L#1) at (48.2,#1) {};
  \node[inner sep=0pt] (R#1) at (126.4,#1) {};
  \begin{scope}[on background layer]
    %
    \node[fit=(t#1)(a#1)(b#1)(c#1)(L#1)(R#1), inner xsep=0pt, inner ysep=1.0mm,
          rounded corners=3pt, draw=#2!55, line width=0.6pt, fill=#2!4,
          dash pattern=on 1.8pt off 1.4pt] {};
  \end{scope}}
\oplane{79}{OpHarnC}{HARNESS ONLY}%
  {extend the scaffold: prompt, memory, built-in tools, skills, MCP}%
  {re-paid at \textbf{every inference}, out of a bounded budget}
\oplane{53}{OpDataC}{DATA ONLY}%
  {amplify the corpus: press out what is held, mark the boundary}%
  {cannot reach capability the rollouts \textbf{never exhibit}}
\oplane{27}{OpModC}{MODEL ONLY}%
  {internalize the behaviour\\into the weights}%
  {pays a \textbf{training run} for what may be a scaffold rule}

\blockarrow{128.3}{58}{figCline}

\begin{scope}[on background layer]
  \bandfill{figC}{(136.6,15)}{(211,102)}
\end{scope}
\bandhead{136.6}{211}{102}{C}{III}%
  {\;WHAT NONE OF THEM CAN 
  SAY\;}

\node[font=\fontsize{7.2}{7.2}\selectfont\itshape, text=OpHarnC!55!black]
  at (174,89.5) {one example of a composed sequence};

\node[st, text width=12mm] (s1) at (149,80.5)
  {\textcolor{OpHarnC!72!black}{\textbf{H}}\ extend the scaffold};
\node[st, text width=12mm] (s2) at (165,80.5)
  {\textcolor{OpDataC}{\textbf{D}}\ amplify what that revealed};
\node[st, text width=12mm] (s3) at (181,80.5)
  {\textcolor{OpModC}{\textbf{M}}\ internalize it into the weights};
\node[st, text width=12mm] (s4) at (197,80.5)
  {\textcolor{OpHarnC!72!black}{\textbf{H}}\ \textbf{retire} the extension \faCut};
\foreach \a/\b in {s1/s2, s2/s3, s3/s4}
  {\draw[ar, draw=OpHarnC!70] (\a) -- (\b);}
\node[font=\fontsize{8.4}{8.4}\selectfont\bfseries, text=OpHarnC!62!black]
  at (174,69.5)
  {capability retained, scaffold budget \textbf{returned}};

\draw[axis] (143,29.5) -- (143,63);
\draw[axis] (143,29.5) -- (207,29.5);
\node[font=\fontsize{8.4}{8.4}\selectfont, text=black!55, rotate=90]
  at (139.6,46)
  {scaffold budget};
\foreach \x in {149,165,181,197}
  {\draw[draw=black!20, dash pattern=on 1.2pt off 1.5pt] (\x,29.5) -- (\x,63);}
\draw[line width=2.0pt, draw=figCline, line join=round]
  (144,33) -- (149,33) -- (149,56) -- (197,56) -- (197,33) -- (205,33);
\node[onlab] at (172,56) {paid at \textbf{every} inference};
\draw[draw=figCline!60, line width=0.5pt] (204,36.4) -- (204,33.2);
\node[onlab] at (204,38.6) {returned};
\node[font=\fontsize{8.4}{9.6}\selectfont, text=black!55, align=center,
      text width=64mm]
  at (174,21) {the answer is a \textbf{sequence} across all three surfaces};

\draw[rounded corners=3pt, draw=black!30, line width=0.7pt,
      dash pattern=on 2.2pt off 1.8pt] (1,3.0) rectangle (211,11.5);
%
%
\foreach \x/\c/\t in {%
  3.0/OpHarnC/{harness surface}, 39.0/OpDataC/{data surface},
  70.5/OpModC/{model surface},
  104.6/MetaC/{the cost this operator cannot see}}
  {\fill[\c!70, rounded corners=0.6pt] (\x,6.2) rectangle (\x+3.0,8.6);
   \node[font=\small, text=black!58, anchor=west] at (\x+4.4,7.3) {\t};}
\node[font=\small, text=black!58, anchor=west] at (164.7,7.3)
  {\faCut\; retired after internalization};
\end{tikzpicture}}
\caption{\textbf{Why single-surface improvement is insufficient.} \emph{I}: a deficiency produces evidence that carries \textbf{no label} saying which of the three surfaces is responsible. \emph{II}: each single-surface operator closes a valid loop on its own surface but incurs a cost it cannot see (\textcolor{MetaC}{red}): scaffold extensions are re-paid as context at every inference, data amplification cannot reach capability the rollouts never exhibit, and training may regress already-held behaviour. \emph{III}: a \textbf{composed sequence} pays neither. Extend the scaffold, amplify what it reveals, internalize into weights, then retire the extension: this \textbf{retains the capability while returning the scaffold budget to baseline}. The chart's ordinate is that budget, so the fall at station four is the cost returned. Colours denote the three surfaces: \dhl{data}, \hhl{harness}, and the \mhl{model}; the scissors mark licensed retirement.}
\label{fig:problem}
\end{figure*}
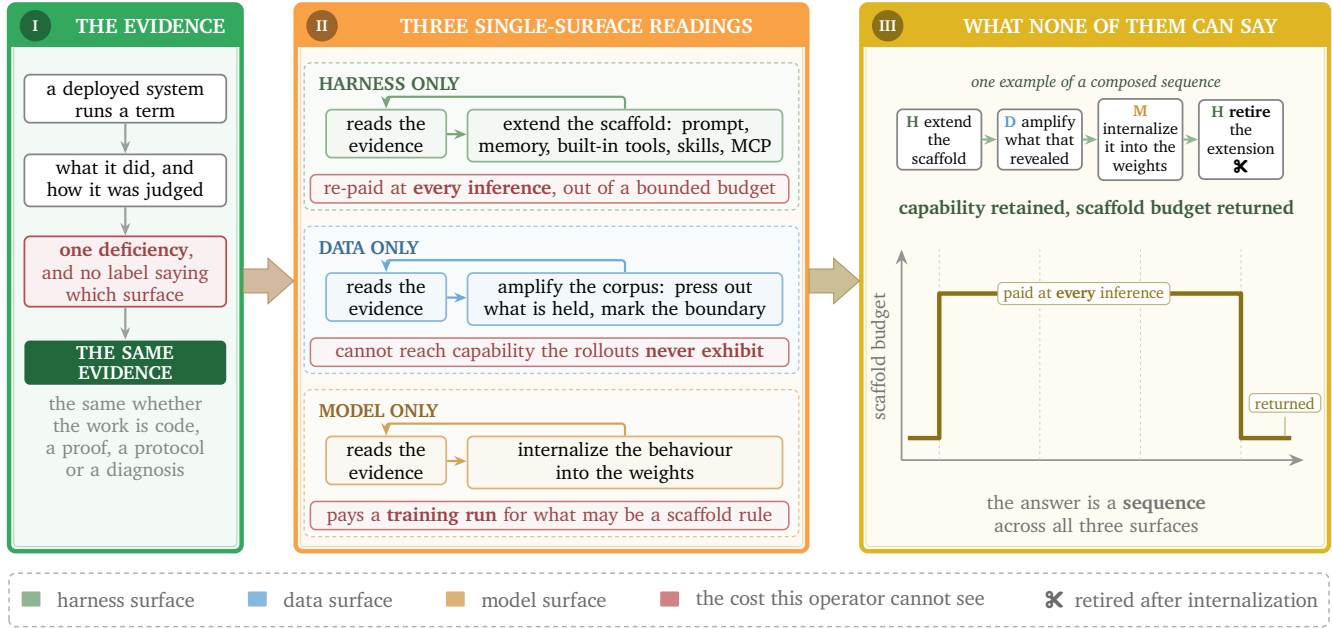

Recursive self-improvement, or RSI, asks whether a system can read its own failures and improve the model-building machinery itself, so that every model built thereafter inherits the gain. Recent systems show this is no longer speculative: agents rewrite their own scaffolds~\citep{zelikman2023stop,hu2024adas,zhang2025dgm} and generate their own finetuning data~\citep{zweiger2025seal,khan2025dataenvgym}. Yet RSI has so far been validated almost exclusively on coding tasks and formal benchmarks such as multiple-choice science QA, mathematical problem sets, and executable test suites. In our survey of forty-five recent systems, $69\%$ close their improvement loop against a target a machine can check for free, and the systems reporting scientific benchmarks fall inside that majority rather than outside it (\Cref{fig:census}). This is a \emph{format} bound rather than a subject bound.\label{sec:intro:defect} What such a loop certifies is restricted, benchmark-bound capability, meaning competence inside a pre-specified, machine-checkable slice of a discipline, not general capability in the discipline itself, where the question is not given and correctness is settled by argument, replication, or measurement. Nor can specialization defer the problem: a deployed system's behaviour is determined jointly by its data, its weights, and its execution scaffold, and a change in any one alters what the other two describe (\Cref{fig:problem}), so what is needed is a unified paradigm over all three surfaces together with a meta-level policy governing which change is made next.

We present \ours, a meta-recursive self-improving system built on three typed operators that share one \emph{loop kernel} and consume the same \emph{learning signal}. \dhl{\dataop} synthesizes verified training records from execution experience, amplifying what the model already does well and marking where that competence ends. \hhl{\harnessop} edits the execution scaffold through typed patches over five slots: the system prompt, memory, built-in tools, skills, and tools and resources mounted through MCP, taking effect at once with no training cost. \mhl{\modelop} modifies model parameters and architecture under bounded training recipes, internalizing capability into the weights so that it persists across scaffolds and adds no inference cost. The operators are connected by \transag adapters, which pass the learning signal and their outputs between them. Above them, the \rsiag optimizes on two axes. Horizontally, it decides the sequence in which operators are applied, since the order changes what each subsequent operator reads and some orders are ill-posed; vertically, it rewrites each operator's own proposal policy, improving how that operator diagnoses failures and proposes changes. The two axes are optimized jointly under one feedback stream, since a composition inherits the quality of its weakest operator and a fixed operator set passes its defects upward. Above that, the \metaag revises the scheduling policy itself once an improvement term completes. Three nested levels modify three different objects: operators change the system, the \rsiag changes the composition, and the \metaag changes the rule of composition. These control roles are defined by what they may read and write rather than by what occupies them. Any of the four can therefore be held by a \ab{human expert} instead of a model, on the same typed contracts and the same audit: the scheduler, a sub-agent, the meta agent, or a transition adapter. We validate \ours on code execution and closed-form scientific reasoning, where the community's evidence standard is highest. Throughout, the system improves itself using only itself: every model-driven role in the loop is played by the target model, so no stronger external model proposes, judges, or schedules, and the gain is attributable to the mechanism rather than to a teacher.

The framework opens two complementary routes forward, one through \hhl{\harnessop} and one through \mhl{\modelop}. We redefine \dhl{\dataop} as the operator that turns execution experience into verified records and marks the boundary of what that experience supports, and its output is consumed by both \harnessop and \modelop; the results of their changes flow back to \dataop as fresh evidence that rescales what the system knows and drives the next round. This feedback across the three operators makes the system self-reinforcing, with capability compounding across rounds. The framework is designed for scientific and engineering practice as it actually occurs. In any discipline, work decomposes into stages, from hypothesis framing and experimental design to execution and interpretation, and each stage already has the shape a loop expects: it takes a typed input, produces an observable outcome, and leaves a record the next stage can use. This means a domain need not close its entire loop at once. A local loop at a single stage already produces records that adjacent stages can build on, and the composition machinery extends these local loops into pipeline-level ones. Crucially, the records a loop leaves behind accumulate across rounds and transfer across domains in a way that model checkpoints cannot. It is precisely this accumulation and transfer of records that enables improvement to compound beyond the narrow slice where it began. Building on this, the framework states \ab{five structural laws} concerning where loops exist, how operators compose, and what external supervision can buy, each formulated as a refutable claim (\Cref{sec:discussion:laws}). These laws jointly support a core position: the fundamental unit of progress is not the model, but the loop. The productive question is therefore not ``how to make the model stronger,'' but ``how cheaply can a new loop be closed where one has never been closed before'' (\Cref{sec:discussion:position}).

%% file: sections/02_related_work.tex
\section{Related Work}
\label{sec:related}

\subsection{Foundation Models}
\label{sec:related:fm}

The capability that self-improvement now attempts to extend was itself produced by a sequence of engineering regimes, and the shape of that sequence explains where the remaining headroom lies. Scale was the first regime: the Transformer~\citep{vaswani2017attention} made compute the binding constraint, and empirical scaling laws turned model size, data volume, and compute into a predictable trade~\citep{kaplan2020scaling,hoffmann2022chinchilla}, yielding models whose few-shot competence emerged without task-specific training~\citep{brown2020gpt3}. Alignment was the second: instruction tuning and reinforcement learning from human feedback converted raw next-token competence into a followable interface~\citep{ouyang2022instructgpt,bai2022constitutional}, and open-weight families made that interface broadly reproducible~\citep{touvron2023llama,dubey2024llama3,yang2025qwen3}. 

Inference-time computation was the third: chain-of-thought prompting showed that additional serial computation buys accuracy at fixed weights~\citep{wei2022cot}, a trade later made explicit by test-time scaling analyses~\citep{snell2024testtime} and internalized by reasoning models trained with verifiable rewards~\citep{openai2024o1,guo2025deepseekr1}. The fourth regime, and the one that current systems occupy, moved capability out of the weights entirely: retrieval, tools, memory, and long-horizon agent loops~\citep{schick2023toolformer,yao2023react,wang2023voyager} now mediate most of what a deployed model can accomplish, and standardized interfaces for tool and context provision have made this scaffold a first-class engineering artifact. Read as a sequence, these regimes show a steady migration of the effective locus of capability: from parameters, to alignment data, to inference procedure, to the surrounding execution structure. A self-improvement framework restricted to any one of them therefore addresses only part of what determines a model's behavior. That same migration is why \ours treats data, scaffold, and the model as three writable surfaces of one system rather than as three separate targets.

\subsection{Recursive Self-Improvement}
\label{sec:related:rsi}

The idea is old, driving the classical intelligence-explosion argument~\citep{good1965ultraintelligent} and formalized in the G\"odel machine~\citep{schmidhuber2007goedel}, but only recently buildable. What has been built sorts by which part of the system it treats as mutable, and that sorting is what makes the field's concentration visible. The earliest work treats the \emph{output}: self-refinement and verbal reflection revise an answer within an episode~\citep{madaan2023selfrefine,shinn2023reflexion}, which is cheap to check but leaves nothing behind and, without an external signal, corrects little~\citep{huang2024selfcorrect}. A second family treats the \emph{scaffold}: STOP improves the program that improves programs~\citep{zelikman2023stop}; ADAS and AFlow search over architectures and workflow graphs~\citep{hu2024adas,zhang2024aflow}; G\"odel Agent and the Darwin G\"odel Machine rewrite their own code under empirical selection~\citep{yin2025godelagent,zhang2025dgm}; and Self-Harness, DemoEvolve and MetaSkill-Evolve target harness evolution directly~\citep{selfharness2026,demoevolve2026,metaskill2026}.

A third treats the \emph{data}: STaR bootstraps rationales from correct answers~\citep{zelikman2022star}, self-rewarding models produce their own preference signal~\citep{yuan2024selfrewarding}, Self-Adapting LMs emit self-edits that become finetuning data~\citep{zweiger2025seal}, and DataEnvGym, SEAL and RSIBench-Data place the data generator, the environment and the post-training stack respectively under control~\citep{khan2025dataenvgym,ant2026seal,meng2026rsibenchdata}. A fourth, much smaller, crosses these boundaries: SIA updates harness and weights together and reports the combination dominating harness-only improvement~\citep{hexo2026sia}, while Hyperagents and Escher-Loop optimize over several components at once~\citep{hyperagents2026,escherloop2026} and AlphaEvolve evolves programs against a fixed evaluator to improve algorithms and hardware designs~\citep{novikov2025alphaevolve}. Surveys confirm the partition and its imbalance~\citep{ren2026survey,gao2025selfevolving}, and complementary work surveys the risks and structural limits persistent self-modification introduces~\citep{misevolve2026,introspection2026}.

Read by subject the taxonomy looks diverse; read by what closes each loop it collapses. In $69\%$ of these systems the object of verification is a key or a test (\Cref{tab:census} gives the per-system assignment), and the systems reporting scientific benchmarks are inside that majority rather than outside it. What they improve is real, and calling it a coding monoculture misdescribes it; what they certify is \emph{restricted, benchmark-bound} capability, bounded by the closed format that made the loop affordable. Two gaps persist. Systems crossing substrate boundaries hard-wire a single order rather than deciding it from evidence, though composition across substrates is exactly where non-commutativity and redundancy appear; and the improvement operator is almost always a fixed program, so the loop improves the system but nothing improves the loop. \ours takes up both, by making the operator set typed and composable and by placing the scheduling policy and each operator's proposal policy under a shared meta-optimizer.

\begin{figure}[t]
\centering
\includegraphics[width=0.94\linewidth]{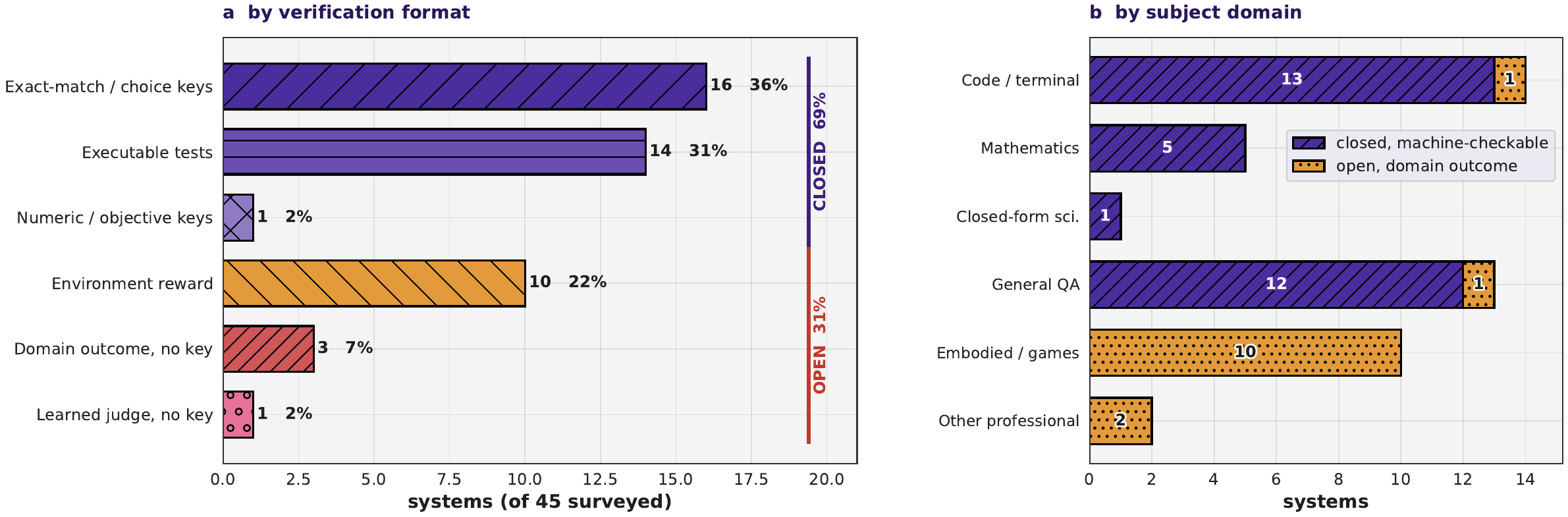}
\caption{\textbf{The binding constraint on RSI is verification format, not subject.} Of $45$ surveyed systems, \textbf{69\% close their improvement loop against a \textcolor{ClosedC}{closed, machine-checkable} target}: a test suite, an exact-match key, a numeric objective, or a multiple-choice label~(a). Grouped by subject~(b), nearly every scientific system falls inside the \textcolor{ClosedC}{\textbf{closed}} block, code and mathematics along with closed-form science suites such as GPQA and AIME. The one exception is a code system closed by rubric rather than by executable tests~\citep{cai2026moss}, and \textcolor{OpenCink}{\textbf{open}} targets otherwise concentrate in embodied games, $10$ of the $14$, and professional domains. Gains on scientific benchmarks therefore certify \emph{restricted, benchmark-bound} capability rather than general capability in the discipline. Per-system assignments, and the target each system verifies against, in \Cref{tab:census}.}
\label{fig:census}
\end{figure}

\begin{figure}[t]
\centering
\includegraphics[width=\linewidth]{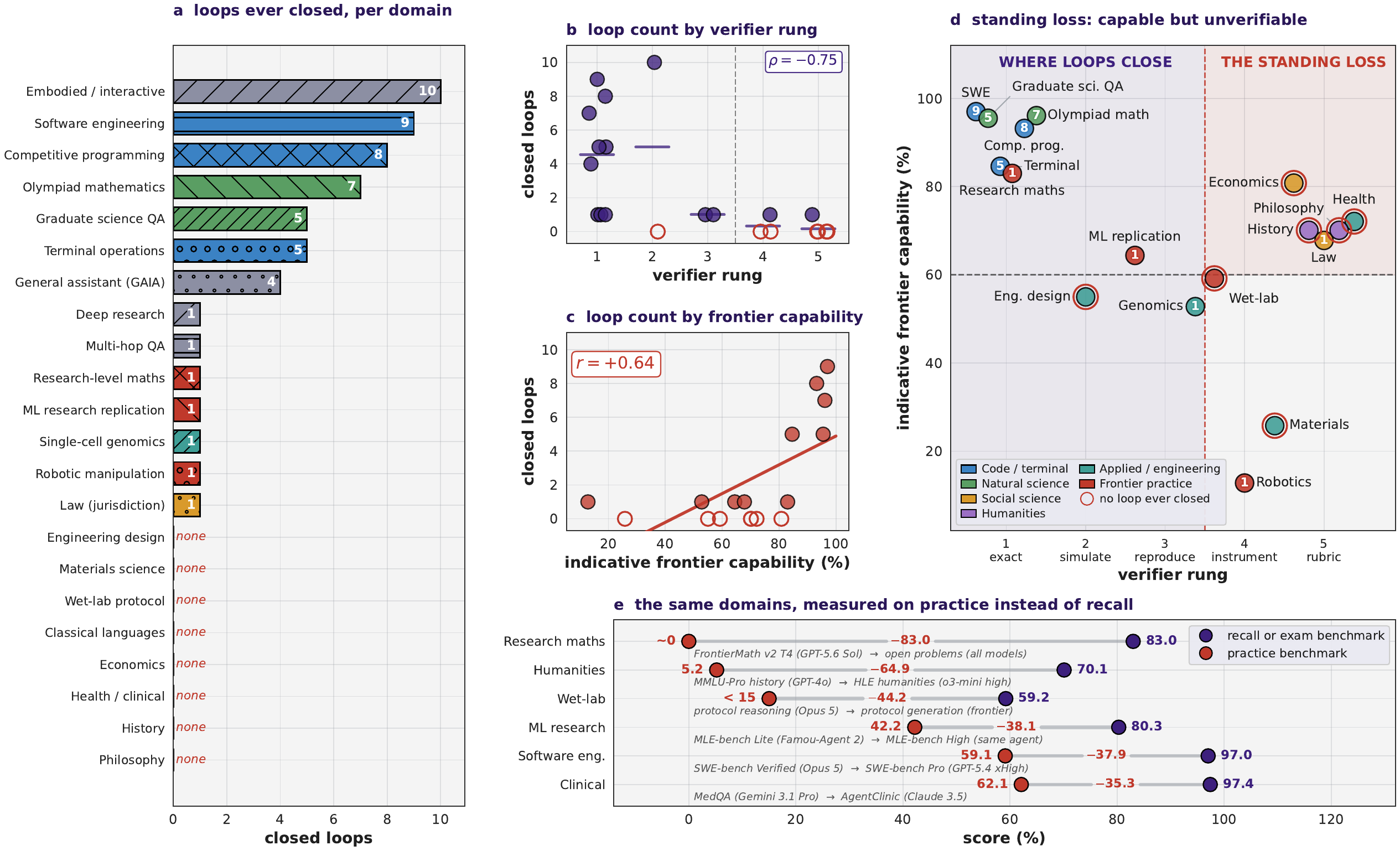}
\caption{\textbf{Improvement loops track verification format, not capability},
across $22$ domains and $55$ closed loops (\Cref{tab:domains}).
\emph{(b)}~Loop count falls with the \textbf{verifier rung} of \Cref{tab:verifier},
read off each domain's \emph{practice} benchmark and named on~(d)'s axis: Spearman
$\rho=-0.75$, $p<0.001$. \textbf{$53$ of the $55$ close at rungs~1 to~3}, and none of
the $33$ loops added in August 2026 landed above rung~2. \emph{(c)}~Capability
predicts loops too, at a zero-order $r=+0.64$ ($p=0.006$), but the two are not
interchangeable: \textbf{controlling for the rung, capability falls to a partial
$\mathbf{r=+0.45}$} ($p=0.08$, n.s.), whereas controlling for capability the rung
holds at $r=-0.67$ ($p=0.005$). \emph{(d)}~The
\textcolor{negred}{\textbf{standing loss}} is the five rung-5 domains above the
threshold, which capability's non-monotonicity across rungs shows are not the domains
models are worst at. \emph{(e)}~Every domain scores \textbf{35 to 83 points lower} on
practice than on recall, so the capability axis is an \emph{upper bound}. Hollow
markers mark no closed loop, and \textbf{eight domains have never had one} even after
the census doubled. \Cref{app:domains} justifies every rung, and why (c) and~(d) plot
only the $17$ domains carrying a recall subject accuracy; \Cref{app:pairs} sources~(e).}
\label{fig:landscape}
\end{figure}

\subsection{Capability Boundaries}
\label{sec:related:boundary}

A third line of work measures where model capability actually ends, and it is what makes the previous subsection's concentration a problem rather than a preference. Broad multi-subject suites established that competence is highly uneven across disciplines~\citep{hendrycks2021mmlu,wang2024mmlupro,srivastava2023bigbench}, and expert-authored evaluations sharpened the picture: GPQA isolates graduate-level science that resists web search~\citep{rein2023gpqa}, FrontierMath targets research-level mathematics~\citep{glazer2024frontiermath}, and Humanity's Last Exam spans classical languages, ancient history and philosophy alongside the sciences~\citep{phan2025hle}. The picture splits along verification format. Coding and terminal benchmarks have driven a decade of scaffolding progress~\citep{jimenez2024swebench,terminalbench2025,jain2024livecodebench}, and closed-form science has been compressed, GPQA-Diamond rising from $38.8\%$ for the best model at the benchmark's release to the low nineties in two years, above the $65$ to $74\%$ \citet{rein2023gpqa} report for PhD-level experts, while open-ended and practice-based domains remain far from expert performance.

A graduate-level multiple-choice item is a closed instrument: the question is given, the answer set enumerated, correctness a lookup. Saturating it shows the knowledge and the reasoning chain are present under that framing, and leaves open how much of the discipline's actual work it covers, where the question is not given and correctness is settled by argument, replication or measurement. On the open-ended side the same models remain far from expert performance, the deficit concentrated where cheap verifiers are missing~\citep{phan2025hle}, and benchmarks built around practice rather than question answering reinforce the asymmetry: replicating machine-learning research, reproducing papers and executing laboratory protocols stay hard for systems that saturate multiple-choice science~\citep{chan2025mlebench,wijk2024rebench,starace2025paperbench,laurent2024labbench,tian2024scicode}. Panel~(e) of \Cref{fig:landscape} quantifies this on six paired benchmarks, whose endpoints and sources are given in \Cref{app:pairs}, where the practice end scores \textbf{35 to 83 points below} the recall end on every pair, making the indicative capability axis an upper bound rather than a measure. Only the MLE-bench pair is measured \emph{within} one system; the other five compare each end's best published frontier score; and the humanities gap is widest partly because its practice end is old, $5.2$ being the HLE paper's own Humanities row for o3-mini~(high)~\citep{phan2025hle} while frontier models now reach about $55$ on HLE \emph{overall} closed-book and about $65$ with tools. No per-category humanities number is published on any board reporting the overall figure, so the earlier endpoint is kept and labelled, not imputed.

Two regimes therefore have to be distinguished, and \Cref{fig:landscape} separates them while \Cref{fig:census} shows where the effort went instead: domains whose competence is largely present but around which no loop has closed, and domains where it is absent and no scaffolding will manufacture it. We place each domain on a rung of the \textbf{verifier ladder} (\Cref{tab:verifier}; defined in \Cref{sec:discussion:verifier}): rung~1 for executable tests and exact match, rung~2 for simulation, rung~3 for reproduction of a reported result, rung~4 for protocol execution with an instrument, and rung~5 for an expert rubric with no ground truth. The rung is assigned from the verification method of the domain's \emph{practice} benchmark rather than its recall benchmark, so it is a property of that benchmark; \Cref{tab:domains} gives it per domain and \Cref{app:domains} justifies every assignment. Of the $55$ closed loops in the census, $53$ sit at rungs~1 to~3; the two exceptions are one loop in robotic manipulation at rung~4 and one narrow charge-classification loop in law at rung~5. That census doubled in August 2026, from $22$ loops over $18$ domains, and the doubling is what makes the zeros mean something: the $33$ new loops include none at rung~4 or~5, and the eight domains that had never had one still have none. The five domains in the standing-loss region, economics, clinical decision support, law, history and philosophy, all sit at rung~5 with capability between $67.8$ and $80.8$, comparable to the rung-1 to rung-3 domains where RSI has worked, but their practice admits no automatic target. The reporting capability threshold is $60$, and the standing-loss set is the same five domains for any threshold in $(59.2, 67.8]$, whose lower bound is wet-lab protocol at $59.2$ and upper bound is law at $67.8$. Nor does a closed loop require capability above the line, since robotic manipulation at $12.8$ and single-cell genomics at $52.8$ have both closed one. The distinction is operational: the first regime is addressable by amplification and external structure, the second requires new knowledge to enter the system, and it is what our \dataop operator is built to draw automatically.

%% file: sections/03_preliminaries.tex
\section{Preliminaries}
\label{sec:prelim}

This section fixes the objects that any self-improvement procedure manipulates, the signal it is allowed to read, and the criterion against which its output is judged. All notation introduced here is reused unchanged in the remainder of the paper.

\subhead{Target system}The entity being improved is not a checkpoint but a triple
\begin{equation}
\Sys \;=\; (\Dat,\; \theta,\; \Harn),
\label{eq:system}
\end{equation}
where $\Dat$ is the \textbf{data state} (the corpus, synthesized records, and curriculum over which the model has been or will be trained), $\theta$ is the \textbf{model state} (trainable parameters and adapters, the bounded architectural choices that place them, and the training configuration that produced them), and $\Harn$ is the \textbf{harness state}, the execution scaffold that mediates every interaction with the model, comprising the system prompt, a persistent memory, a built-in tool set, a skill library, and the tools and resources mounted through MCP. A deployed system is the composition of all three; a change confined to any one of them changes what the deployed system does.

\subhead{Tasks, rollouts, and verification}Let $\mathcal{T}$ denote a task family and $t \sim \mathcal{T}$ a task instance. Executing $\Sys$ on $t$ produces a \textbf{trajectory} $\Traj = (o_1,a_1,\dots,o_n,a_n)$ of observations and actions, terminating in an output that a \textbf{verifier} $v$ maps to a scalar outcome $r = v(t,\Traj) \in [0,1]$. We deliberately do not assume that $v$ is cheap: in executable domains it is a test suite, in closed-form question answering a string comparison, and in open-ended scientific work it is expensive, partial, or unavailable. This variation, rather than any property of the underlying reasoning, is what has historically determined where improvement loops could be closed.

\subhead{Learning signal}A self-improvement procedure never observes $\mathcal{T}$ directly. It observes a \textbf{learning signal}
\begin{equation}
\sig \;=\; \Sigma(\mathcal{E}),
\qquad
\mathcal{E} \;=\; \big\{(t_k,\; \Traj_k,\; y_k)\big\}_{k=1}^{K},
\label{eq:signal}
\end{equation}
the deterministic compilation of an \textbf{evidence set} $\mathcal{E}$ drawn from attempts made by the system in its current state, where $y_k$ is the \textbf{outcome record} of attempt $k$: in the code and QA tracks evaluated here, simply the verifier scalar $r_k$ above, and in general whatever the domain was able to establish about that attempt. We define $\sig$ by the role it plays rather than by the form it takes, because across domains the form varies and the role does not. In implementation $\sig$ is the compiled evidence bundle carrying raw trajectories and diagnoses; later references to trajectories or experience denote components of this bundle.

\subhead{What may enter the evidence set}Four kinds of thing: verifier outcomes of any fidelity, the system's own trajectories (read for the competence they exhibit, not only for whether they succeeded), external knowledge admitted against a demonstrated gap, and the framework's own decision records. A source is admissible if it derives from the current system, is readable by every operator, and carries no authority to decide whether a change was good. What makes something a learning signal is its position in the loop, not its modality.

\subhead{What the compiler emits}$\Sigma$ is deterministic and fixed: it aggregates, attributes, and attaches provenance, emitting one typed object in three layers (an outcome layer, an attribution layer, and a model-attributed mechanism layer, whose schema is \Cref{eq:signature}), and it does not decide what to change. Fixing the typing while leaving the sources open is what lets a single operator set run over domains whose evidence has nothing else in common.

\subhead{Two properties of the signal}It is the \textbf{sole channel} between the environment and any improvement procedure, which is what allows heterogeneous procedures to be compared, exchanged, and composed. And it is \textbf{perishable}: a step that changes what the deployed system does invalidates every signal compiled before it, because the attempts such a signal summarizes were made by a system that no longer exists.

\subhead{Improvement procedure and budget}An \textbf{improvement procedure} is a map
\begin{equation}
U:\;(\Sys,\; \sig,\; \Budget) \;\longmapsto\; (\widetilde{\Sys},\; A,\; E,\; c),
\label{eq:operator}
\end{equation}
which consumes the current system, the learning signal, and a \textbf{budget} of four non-fungible resources, the four bars of the ledger in \Cref{fig:teaser},
\begin{equation*}
\Budget \;=\; \big(\,
\ubl{OpDataC}{\beta_{\mathrm{tok}}}{tokens}\;,\;\;
\ubl{OpHarnC}{\beta_{\mathrm{clk}}}{wall-clock}\;,\;\;
\ubl{OpModC}{\beta_{\mathrm{gpu}}}{compute}\;,\;\;
\ubl{accent}{\beta_{\mathrm{qry}}}{verifier queries}
\,\big),
\end{equation*}
and returns a candidate successor $\widetilde{\Sys}$, exportable \textbf{artifacts} $A$ that other procedures may consume, the \textbf{evidence} $E$ supporting the change, and the realized \textbf{cost} $c \preceq \Budget$. A procedure produces candidates only; it does not promote them. Two procedures $U_i,U_j$ compose by applying one to the output of the other, and in general
\begin{equation}
U_j \circ U_i (\Sys) \;\neq\; U_i \circ U_j (\Sys),
\label{eq:noncommute}
\end{equation}
because each modifies a different component of \Cref{eq:system} and thereby changes the learning signal the next procedure will read.

\subhead{Protected measurement}Separately from $\Sys$ we fix a \textbf{sealed evaluator} $\Qsealed$, a held-out task set $\mathcal{T}^{\star}$, a release rule, and a resource ledger. None of these is part of $\Sys$ and none may be modified by any improvement procedure. This separation distinguishes an improvement in capability from an improvement in the definition of success; without it, the quantity being maximized can be increased by editing the measurement.

\subhead{Improvement episode}An \textbf{episode} is a tuple $e=(\Sys_0,\ \mathcal{T}_{\mathrm{adapt}},\ \mathcal{T}^{\star},\ \Budget_{\mathrm{tot}})$: a clean initial system, a set of tasks on which failures may be observed and improvements attempted, a sealed set on which the final system is scored once, and a total budget. Within an episode the procedure may be invoked repeatedly, producing a \textbf{sequence} $z=(u_1,\dots,u_T)$ of applied procedures and a corresponding chain of systems $\Sys_0 \to \Sys_1 \to \cdots \to \Sys_T$.

\subhead{Notation}$U_j\circ U_i$ denotes applying $U_i$ then $U_j$; $z_{1:t}=(u_1,\dots,u_t)$ is a prefix of an operator sequence. Calligraphic letters denote sets and policies, sans-serif letters $\opD,\opH,\opM$ denote the three operators of \Cref{sec:method:operators}, and starred quantities such as $\Qsealed$ are protected: fixed before an episode begins and not writable by any component at any level.

\subsection{Objective Formulation}
\label{sec:prelim:objective}

Let $\Pi$ denote the internal strategy of the improvement procedures (the way they diagnose, propose, and select), and let $\phi$ denote the policy that produces the sequence $z$ within an episode. The quantity we optimize is the \textbf{deployed improvement productivity} of an episode, penalized by cost, by regression on previously held capability, and by the gap between the working signal and the sealed measurement:
\begin{equation}
J(\Pi,\phi)\;=\;\mathbb{E}_{e\sim\mathcal{E}}
\Big[\;
\annot{mGreen}{\Qsealed(\Sys_T)-\Qsealed(\Sys_0)}{realized gain}
\;-\;\lambda\,\annot{mOrange}{C(z)}{cost}
\;-\;\mu\,\annot{mRed}{\mathrm{Reg}(\Sys_T,\Sys_0)}{regression}
\;-\;\nu\,\annot{mPurple}{\mathrm{Gap}(\sig,\Qsealed)}{evaluator gap}
\;\Big],
\label{eq:objective}
\end{equation}
subject to $C(z)\preceq \Budget_{\mathrm{tot}}$, where $C(z)$ aggregates the realized costs of the applied procedures, $\mathrm{Reg}$ measures degradation on capabilities the initial system possessed, and $\mathrm{Gap}$ penalizes optimization against a signal that diverges from the sealed evaluator (\Cref{sec:method:governance}).
\begin{tightitems}
\item It is evaluated on the \emph{released} successor rather than the best candidate ever generated, so generation and selection are both accounted for (\Cref{sec:method:kernel}, Select/Export).
\item It is evaluated under a \emph{fixed} budget, so an improvement that merely spends more is not an improvement (\Cref{sec:scheduling}).
\item It depends on the pair $(\Pi,\phi)$ rather than on either alone, so maximizing it requires deciding both what each procedure does internally and in which order procedures are applied (\Cref{sec:scheduling}).
\end{tightitems}

%% file: sections/04_method.tex
\section{Methodology}
\label{sec:method}

\begin{figure*}[!t]
\centering
%
%
\input{sections/fig_palette}%
\resizebox{\linewidth}{!}{%
\begin{tikzpicture}[
  x=1mm, y=1mm,
  font=\small,
  card/.style={rounded corners=2.5pt, draw=black!48, line width=0.8pt,
               fill=white, align=center, inner xsep=3pt, inner ysep=2.6pt,
               font=\scriptsize},
  chip/.style={rounded corners=2.2pt, draw=figAline!60, line width=0.6pt,
               fill=figA!12, align=left, inner xsep=3pt, inner ysep=2.2pt,
               font=\scriptsize, text=figAline!85!black},
  trow/.style={rounded corners=2pt, draw=figAline!42, line width=0.6pt,
               fill=white, align=left, inner xsep=3pt, inner ysep=2.2pt,
               font=\scriptsize, text=figAline!85!black},
  ar/.style={-{Stealth[length=2.0mm,width=1.6mm]}, line width=1.0pt,
             draw=black!40},
  edge/.style={-{Stealth[length=2.4mm,width=1.9mm]}, line width=1.35pt},
  elab/.style={circle, fill=white, draw=black!25, line width=0.5pt,
               inner sep=0.8pt, font=\tiny\bfseries},
  lvl/.style={rounded corners=2pt, inner xsep=2.8pt, inner ysep=1.4pt,
              font=\scriptsize\bfseries, text=white},
  band/.style={line width=3.0mm}
]
\path (0,0) rectangle (212,158);

%

%
\newcommand{\bandhead}[6]{%
  \fill[fig#4, rounded corners=3pt] (#1,#3-6.8) rectangle (#2,#3);
  \fill[fig#4] (#1,#3-6.8) rectangle (#2,#3-3.0);
  \node[circle, fill=fig#4line, inner sep=0pt, minimum size=5.0mm,
        font=\scriptsize\bfseries, text=white] at (#1+4.2,#3-3.4) {#5};
  \node[font=\fontsize{9.2}{9.2}\selectfont\bfseries, text=white]
    at ({(#1+#2)/2+4},#3-3.4) {#6};}
\newcommand{\subgroup}[5]{%
  \draw[rounded corners=3pt, draw=#5!55, line width=0.6pt, fill=#5!4,
        dash pattern=on 1.8pt off 1.4pt] (#1,#2) rectangle (#3,#4);}
\newcommand{\subtitle}[4]{%
  \node[font=\scriptsize\bfseries, text=#3] at (#1,#2) {#4};}
\newcommand{\blockarrow}[4]{%
  \fill[#3!45, draw=#3!70, line width=0.5pt]
    (#1,#2-1.9) -- (#1+#4-4.1,#2-1.9) -- (#1+#4-4.1,#2-3.6) -- (#1+#4,#2)
    -- (#1+#4-4.1,#2+3.6) -- (#1+#4-4.1,#2+1.9) -- (#1,#2+1.9) -- cycle;}
\newcommand{\slab}[6]{%
  \draw[rounded corners=1.4pt, fill=#4, draw=#5, line width=0.6pt]
    (#1-#3,#2-3.0) rectangle (#1+#3,#2+3.0);
  \node[font=\scriptsize, anchor=west, inner sep=0pt] at (#1-#3+1.6,#2) {#6};}

\begin{scope}[on background layer]
  \bandfill{figA}{(1,17)}{(64,156)}
\end{scope}
\bandhead{1}{64}{156}{A}{I}{\;WHAT IT IS GIVEN, AND READS\;}

\subgroup{2.4}{119}{62.6}{146}{figAline}
\subtitle{32.5}{143.2}{figAline!85!black}{TARGET SYSTEM}
\slab{32.5}{137.5}{27}{OpDataC!13}{OpDataC!70}
  {\textcolor{OpDataC!70!black}{\faDatabase}\;\textcolor{OpDataC!80!black}{$\Dat$}\ \ \textcolor{black!62}{data state}}
\slab{32.5}{130.5}{27}{OpHarnC!13}{OpHarnC!70}
  {\textcolor{OpHarnC!70!black}{\faPuzzlePiece}\;\textcolor{OpHarnC!72!black}{$\Harn$}\ \ \textcolor{black!62}{harness state}}
\slab{32.5}{123.5}{27}{OpModC!13}{OpModC!70}
  {\textcolor{OpModC!70!black}{\faMicrochip}\;\textcolor{OpModC!80!black}{$\theta$}\ \ \textcolor{black!62}{model state}}

\subgroup{2.4}{92}{62.6}{116}{figAline}
\subtitle{32.5}{113.2}{figAline!85!black}{SPLIT ONCE, AND METERED}
\node[card, text width=44mm] at (32.5,107)
  {\faPen\; \textbf{adaptation split}\ \ {\scriptsize failures observed here}};
\node[card, text width=44mm, draw=MetaC!75, fill=MetaC!6,
      text=MetaC!85!black] at (32.5,98)
  {\faLock\; \textbf{sealed split}\ \ {\scriptsize opened once, after freezing}};

\subgroup{2.4}{62}{62.6}{89}{figAline}
\subtitle{32.5}{86.2}{figAline!85!black}{ANY SOURCE THAT CLOSES THE LOOP}
\node[chip, text width=24mm] (q1) at (20,81) {\faCheckCircle\; verifier outcome};
\node[chip, text width=24mm] (q2) at (20,75.5) {\faRoute\; own trajectories};
\node[chip, text width=24mm] (q3) at (20,70) {\faBookOpen\; admitted corpus};
\node[card, fill=figA!14, draw=figAline, line width=1.1pt, text width=13mm,
      text=figAline!85!black] (sig) at (50,75.5)
  {{\normalsize$\sig$}\\[0.6pt]\textbf{signal}\\[0.6pt]{\scriptsize fixed $\Sigma$}};
\foreach \q in {q1,q2,q3} {\draw[ar, draw=figAline!75] (\q) -- (sig);}
\node[font=\scriptsize, text=black!48, align=center, text width=38mm] at (23,64.5)
  {\emph{position in the loop, not modality}};

\subgroup{2.4}{19}{62.6}{59}{figAline}
\subtitle{32.5}{56.2}{figAline!85!black}{WHAT $\Sigma$ EMITS}
\node[trow, text width=42mm] at (32.5,50)
  {\textcolor{black!45}{\faLock}\; terminal cause};
\node[trow, text width=42mm] at (32.5,44)
  {\textcolor{black!45}{\faLock}\; causal role};
\node[trow, text width=42mm] at (32.5,38)
  {\textcolor{figAline}{\faLightbulb}\; reusable mechanism};
\begin{scope}[shift={(16,28)}]
  \draw[band, draw=OpDataC]    (90:6.0mm) arc (90:0:6.0mm);
  \draw[band, draw=OpHarnC]     (0:6.0mm) arc (0:-90:6.0mm);
  \draw[band, draw=mTeal]     (-90:6.0mm) arc (-90:-180:6.0mm);
  \draw[band, draw=accentlt] (-180:6.0mm) arc (-180:-270:6.0mm);
  \draw[draw=black!42, line width=0.5pt] (0,0) circle (7.5mm);
  \draw[draw=black!42, line width=0.5pt] (0,0) circle (4.5mm);
  \foreach \a in {90,0,-90,-180}
    {\draw[white, line width=0.8pt] (\a:4.3mm) -- (\a:7.7mm);}
\end{scope}
\node[font=\scriptsize, text=OpDataC!85!black, anchor=west] at (24.5,32.8) {knowledge};
\node[font=\scriptsize, text=OpHarnC!85!black, anchor=west] at (24.5,29.4) {reasoning};
\node[font=\scriptsize, text=mTeal!85!black,   anchor=west] at (24.5,26.0) {verification};
\node[font=\scriptsize, text=accentlt!88!black, anchor=west] at (24.5,22.6) {distractor};
%
\node[font=\scriptsize, text=figAline!85!black, align=center, text width=21mm] at (51,26)
  {$\kappa(e)$: \textbf{dominant dimension} per episode};

\blockarrow{64}{88}{figA}{10}

\begin{scope}[on background layer]
  \bandfill{figB}{(74,17)}{(154,156)}
\end{scope}
\bandhead{74}{154}{156}{B}{II}%
  {\;ONE KERNEL, THREE OPERATORS\;}

\node[lvl, fill=figBline] at (82,145) {LEVEL 1};
\node[font=\scriptsize, text=figBline, anchor=west] at (89,145)
  {\textbf{a model proposes on the coloured arc}};

\begin{scope}[shift={(100,119)}]
  \draw[band, draw=black!11]    (118:16mm) arc (118:-242:16mm);
  \draw[band, draw=figB!70]     (118:16mm) arc (118:34:16mm);
  \draw[draw=black!42, line width=0.5pt] (0,0) circle (17.8mm);
  \draw[draw=black!42, line width=0.5pt] (0,0) circle (14.2mm);
  \foreach \a in {118,34}
    {\draw[white, line width=1.0pt] (\a:14.0mm) -- (\a:18.0mm);}
  \draw[-{Stealth[length=2.0mm,width=1.6mm]}, line width=0.9pt,
        draw=figBline] (86:16mm) arc (86:64:16mm);
  \foreach \a in {-10,-140}
    {\draw[-{Stealth[length=2.0mm,width=1.6mm]}, line width=0.9pt,
           draw=black!42] (\a:16mm) arc (\a:\a-20:16mm);}
  \node[circle, fill=white, draw=figBline!60, line width=0.9pt,
        minimum size=26mm] at (0,0) {};
  \node[font=\scriptsize\bfseries, text=black!62, align=center] at (0,3.0mm)
    {write surface};
  \node[font=\small, align=center] at (0,-0.2mm)
    {\textcolor{OpDataC}{$\Dat$}\,\textbar\,\textcolor{OpHarnC}{$\Harn$}%
     \,\textbar\,\textcolor{OpModC}{$\theta$}};
  \node[font=\tiny, text=black!45, align=center] at (0,-3.8mm)
    {the one field\\that differs};
\end{scope}
\node[font=\scriptsize, text=black!52, anchor=west] at (122,132.2) {\textsc{observe}};
\node[font=\scriptsize, text=figBline, anchor=west] at (122,127.8)
  {\textbf{\textsc{diagnose}}};
\node[font=\scriptsize, text=figBline, anchor=west] at (122,123.4)
  {\textbf{\textsc{propose}}};
\node[font=\scriptsize, text=black!52, anchor=west] at (122,119) {\textsc{validate}};
\node[font=\scriptsize, text=black!52, anchor=west] at (122,114.6) {\textsc{execute}};
\node[font=\scriptsize, text=black!52, anchor=west] at (122,110.2) {\textsc{select}};
\node[font=\scriptsize, text=black!52, anchor=west] at (122,105.8) {\textsc{export}};
\node[font=\scriptsize, text=black!52, align=center, text width=70mm] at (114,96)
  {\faLock\; \textbf{one cut is load-bearing}; the stage count is not};

\subgroup{75.4}{76}{152.6}{92}{figBline}
\subtitle{114}{89.2}{figBline!85!black}{THREE INSTANTIATIONS, ONE RING}
\node[card, text width=17mm, draw=OpDataC!70, fill=OpDataC!7,
      text=OpDataC!80!black] at (91,82)
  {\textbf{\dataop}\\[0.4pt]{\scriptsize writes $\Dat$}};
\node[card, text width=17mm, draw=OpHarnC!70, fill=OpHarnC!7,
      text=OpHarnC!72!black] at (114,82)
  {\textbf{\harnessop}\\[0.4pt]{\scriptsize writes $\Harn$}};
\node[card, text width=17mm, draw=OpModC!70, fill=OpModC!7,
      text=OpModC!80!black] at (137,82)
  {\textbf{\modelop}\\[0.4pt]{\scriptsize writes $\theta$}};

\subgroup{75.4}{19}{152.6}{72}{figBline}
\subtitle{114}{69.2}{figBline!85!black}{FIVE ADMISSIBLE EDGES, ONE FORBIDDEN}
\node[font=\scriptsize, text=MetaC!85!black] at (114,64.6)
  {$\opH\!\to\!\opM$ \textbf{forbidden}: the dataset predates the change};
\node[circle, draw=OpDataC, fill=bBlue, line width=1pt, minimum size=10mm,
      font=\scriptsize\bfseries, text=OpDataC, inner sep=0pt] (gD) at (87,52) {$\opD$};
\node[circle, draw=OpHarnC, fill=bGreen, line width=1pt, minimum size=10mm,
      font=\scriptsize\bfseries, text=OpHarnC, inner sep=0pt] (gH) at (141,52) {$\opH$};
\node[circle, draw=OpModC, fill=bOrange, line width=1pt, minimum size=10mm,
      font=\scriptsize\bfseries, text=OpModC, inner sep=0pt] (gM) at (114,39) {$\opM$};
\draw[edge, draw=OpDataC!85] (gD) to[bend left=9]  node[elab,pos=0.5]{1} (gH);
\draw[edge, draw=OpHarnC!85] (gH) to[bend left=9]  node[elab,pos=0.5]{3} (gD);
\draw[edge, draw=OpDataC!85] (gD) to[bend right=11] node[elab,pos=0.5]{2} (gM);
\draw[edge, draw=OpModC!85]  (gM) to[bend right=11] node[elab,pos=0.5]{4} (gD);
\draw[edge, draw=OpModC!85]  (gM) to[bend right=11] node[elab,pos=0.5]{5} (gH);
\draw[edge, draw=MetaC, dash pattern=on 1.6pt off 1.4pt, line width=1.2pt]
  (gH) to[bend right=11] node[elab,pos=0.5,text=MetaC]{\faTimesCircle} (gM);
\draw[edge, draw=OpDataC!85] (gD) to[out=172,in=224,looseness=6] (gD);
\draw[edge, draw=OpHarnC!85] (gH) to[out=8,in=-44,looseness=6] (gH);
\node[font=\scriptsize, text=black!58, anchor=west] at (79,30)
  {\textbf{1} experience extraction};
\node[font=\scriptsize, text=black!58, anchor=west] at (79,26)
  {\textbf{2} dataset materialization};
\node[font=\scriptsize, text=black!58, anchor=west] at (79,22)
  {\textbf{3} signal recompilation};
\node[font=\scriptsize, text=black!58, anchor=west] at (115,30)
  {\textbf{4} boundary re-probing};
\node[font=\scriptsize, text=black!58, anchor=west] at (115,26)
  {\textbf{5} redundancy reconciliation};
\node[font=\scriptsize, text=MetaC!85!black, anchor=west] at (115,22)
  {\faTimesCircle\ the excluded ordering};

\blockarrow{154}{88}{figB}{8}

\begin{scope}[on background layer]
  \bandfill{figC}{(162,17)}{(211,156)}
\end{scope}
\bandhead{162}{211}{156}{C}{III}{\;WHO DECIDES\;}

\node[lvl, fill=figCline!85!black] at (172,145) {LEVEL 3};
\node[card, text width=34mm, fill=white, draw=figCline!55] (mv) at (187.5,137)
  {\textbf{\metaag}\\[0.5pt]{\scriptsize revises the scheduler,\\once per term}};
\node[lvl, fill=figCline!85!black] at (172,127) {LEVEL 2};
\node[card, text width=34mm, fill=white, draw=figCline!55] (cv) at (187.5,119)
  {\textbf{\rsiag}\\[0.5pt]{\scriptsize reads $z_{1:t}$ and $\sig_t$, then picks one axis}};
\draw[ar, draw=figCline!70, line width=1.3pt] (mv) -- (cv);

\node[card, text width=15mm, fill=white, draw=figCline!55] (hz) at (173,104)
  {\textbf{horizontal}\\[0.4pt]{\scriptsize extend $z$}};
\node[card, text width=15mm, fill=white, draw=figCline!55] (vt) at (199,104)
  {\textbf{vertical}\\[0.4pt]{\scriptsize rewrite $\pi$}};
\draw[ar, draw=figCline!70, line width=1.3pt] (cv.south) to[out=232,in=90] (hz.north);
\draw[ar, draw=figCline!70, line width=1.3pt] (cv.south) to[out=308,in=90] (vt.north);

\node[card, text width=15mm, fill=white, draw=figCline!45] (seq) at (173,90)
  {\tikz[baseline=-0.55ex]{\node[circle,fill=OpDataC,inner sep=1.1pt,text=white,font=\tiny]{D};}--%
   \tikz[baseline=-0.55ex]{\node[circle,fill=OpHarnC,inner sep=1.1pt,text=white,font=\tiny]{H};}--%
   \tikz[baseline=-0.55ex]{\node[circle,fill=OpModC,inner sep=1.1pt,text=white,font=\tiny]{M};}\\[1pt]
   {\scriptsize typed program}};
\node[card, text width=15mm, fill=white, draw=figCline!55] (sub) at (199,91)
  {\textbf{\rsisub}};
\draw[ar, draw=figCline!70] (hz) -- (seq);
\draw[ar, draw=figCline!70] (vt) -- (sub);

\subgroup{178}{58}{209.6}{80}{figCline}
\subtitle{193.8}{77.2}{figCline!85!black}{CONTRACT}
\node[font=\scriptsize, anchor=west, text=figCline!85!black] at (180,71.5)
  {\faPen\; \textbf{mutable} \ rewritable};
\node[font=\scriptsize, anchor=west, text=black!58] at (180,66.5)
  {\faArrowRight\; \textbf{action} \ writes $\Sys$};
\node[font=\scriptsize, anchor=west, text=black!58] at (180,61.5)
  {\faLock\; \textbf{protected} \ outside};
\draw[ar, draw=figCline!70] (sub.south) -- (199,80.4);

\node[card, text width=36mm, fill=black!5, draw=black!35] (prot) at (187.5,46)
  {\faLock\; \textbf{sealed evaluator \textbar\ gate}\\[0.5pt]
   {\scriptsize outside every write surface}};
\draw[ar, draw=black!38] (seq.south) -- (173,50.3);
\draw[ar, draw=black!38] (199,57.6) -- (199,50.3);

\node[card, text width=34mm, fill=figA!14, draw=figAline, line width=1.2pt,
      text=figAline!85!black] (rel) at (187.5,31)
  {\faCheckCircle\; \textbf{RELEASED SUCCESSOR}\\[0.5pt]
   {\scriptsize scored once, on the sealed split}};
\draw[ar, draw=figAline!85, line width=1.3pt] (prot) -- (rel);

\draw[-{Stealth[length=3.0mm,width=2.4mm]}, line width=1.5pt, dashed,
      draw=figAline, rounded corners=3pt]
  (rel.south) -- (187.5,12) -- (69,12) -- (69,75.5) -- (sig.east);
\node[rounded corners=2.4pt, fill=white, draw=figAline!55, line width=0.6pt,
      inner xsep=3pt, inner ysep=1.8pt, font=\scriptsize,
      text=figAline!85!black] at (125,12)
  {the signal is \textbf{recompiled} after every capability-altering step};

\draw[rounded corners=3pt, draw=black!30, line width=0.7pt,
      dash pattern=on 2.2pt off 1.8pt] (1,1.0) rectangle (211,8.0);
\foreach \x/\c/\t in {%
  8/OpDataC/{data surface}, 34/OpHarnC/{harness surface},
  64/OpModC/{model surface}, 93/figB/{mutable arc: a model proposes}}
  {\fill[\c!70, rounded corners=0.6pt] (\x,3.6) rectangle (\x+3.0,5.6);
   \node[font=\fontsize{8.8}{8.8}\selectfont, text=black!58, anchor=west]
     at (\x+4.2,4.6) {\t};}
\fill[black!18, rounded corners=0.6pt] (145,3.6) rectangle (148,5.6);
\node[font=\fontsize{8.8}{8.8}\selectfont, text=black!58, anchor=west]
  at (149.2,4.6) {protected arcs: code adjudicates};
\end{tikzpicture}}
\caption{\textbf{One improvement term of \ours.} \emph{I}: a target system $(\Dat,\Harn,\theta)$ with a once-opened split compiles rollouts into a \textbf{typed learning signal} and per-episode signatures. \emph{II}: \textbf{one loop kernel}, a mutable model arc with protected code arcs, is instantiated by three operators ($\opD$, $\opH$, $\opM$) differing only in write surface; five of six ordered transitions are admissible, each via a numbered \transag adapter. \emph{III}: the \rsiag schedules horizontally or rewrites operator policies vertically under contract, and the \metaag revises it once per term, while evaluator, gate and ledger stay \textbf{outside every write surface}. The dashed arc is signal recompilation; colours denote the \dhl{data}, \hhl{harness} and \mhl{model} surfaces (legend beneath).}
\label{fig: methods}
\end{figure*}
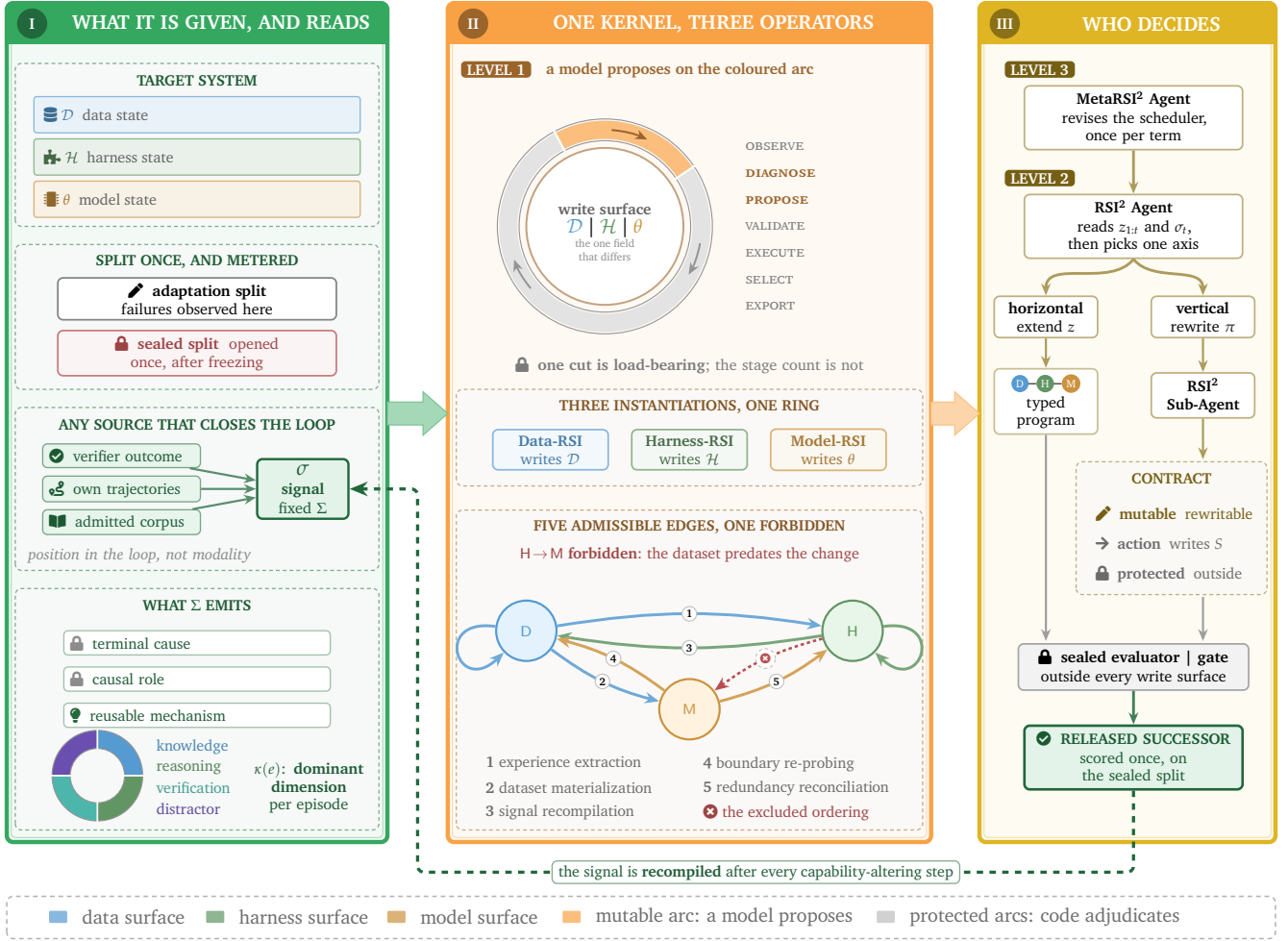

\noindent \textbf{Overview.} \ours is built from the inside out. We first fix a single unified execution paradigm (a cycle closed by a \emph{learning signal} and cut once by an authority boundary) that every self-improvement operator must instantiate, so that operators differ in what they write but not in how they are run, validated, or audited (\Cref{sec:method:kernel}). Instantiating that kernel on the three components of the target system in \Cref{eq:system} yields the three base operators: \dataop, which synthesizes verified records from execution experience and emits the evidence the other two consume; \harnessop, which edits a five-slot execution scaffold without touching the model; and \modelop, which internalizes capability into the model through bounded training recipes (\Cref{sec:method:operators}). Because the three share one artifact vocabulary, their adjacency becomes analyzable rather than arbitrary: a single freshness condition on the learning signal admits exactly five of the six ordered operator pairs, and each admissible edge is realized by a typed \transag adapter that converts a producer's output into a consumer's input (\Cref{sec:composition}). On top of the operator layer, the \rsiag consumes the current sequence and the latest signal and chooses between extending the sequence \emph{horizontally} and rewriting an operator's proposal policy \emph{vertically}, the latter constrained by an explicit contract of mutable, action, and protected surfaces (\Cref{sec:scheduling}). Finally, a meta agent closes the outermost loop by revising the \rsiag's own scheduling policy once an improvement term completes (\Cref{sec:meta}). \Cref{fig: methods} draws the whole term, and its three regions correspond, left to right, to \Cref{sec:method:kernel}, \Cref{sec:method:operators}--\ref{sec:composition} and \Cref{sec:scheduling}--\ref{sec:meta} respectively.

\subsection{The Loop Kernel: One Paradigm for Heterogeneous Operators}
\label{sec:method:kernel}

Operators that write different components of $\Sys$ have almost nothing in common at the level of implementation: one synthesizes training records, one rewrites a system prompt, one submits a training job. What they share is a \emph{contract about authority}; pitching the paradigm at that level, rather than at the level of a stage list, is what makes it portable across domains whose implementations differ entirely.

\begin{definition}[Self-improvement operator]
\label{def:operator}
A \textbf{self-improvement operator} is any process satisfying three conditions:
\emph{(i)} its sole input is the compiled learning signal $\sig$ of \Cref{eq:signal}, never the environment directly;
\emph{(ii)} the part of it that decides \emph{what to change} is a policy $\pi$, and is mutable;
\emph{(iii)} the part of it that decides \emph{whether the change was good} is fixed, external to the operator, and outside its write surface. Formally,
\begin{equation}
U:\; \mathcal{S}\times\Sigma\times\mathcal{B} \;\to\; \mathcal{S}\times\mathcal{A}\times\mathcal{E}\times\mathcal{C},
\qquad
U(\Sys,\sig,\Budget) \;=\; (\widetilde{\Sys},\,A,\,E,\,c).
\label{eq:operator-sig}
\end{equation}
\end{definition}

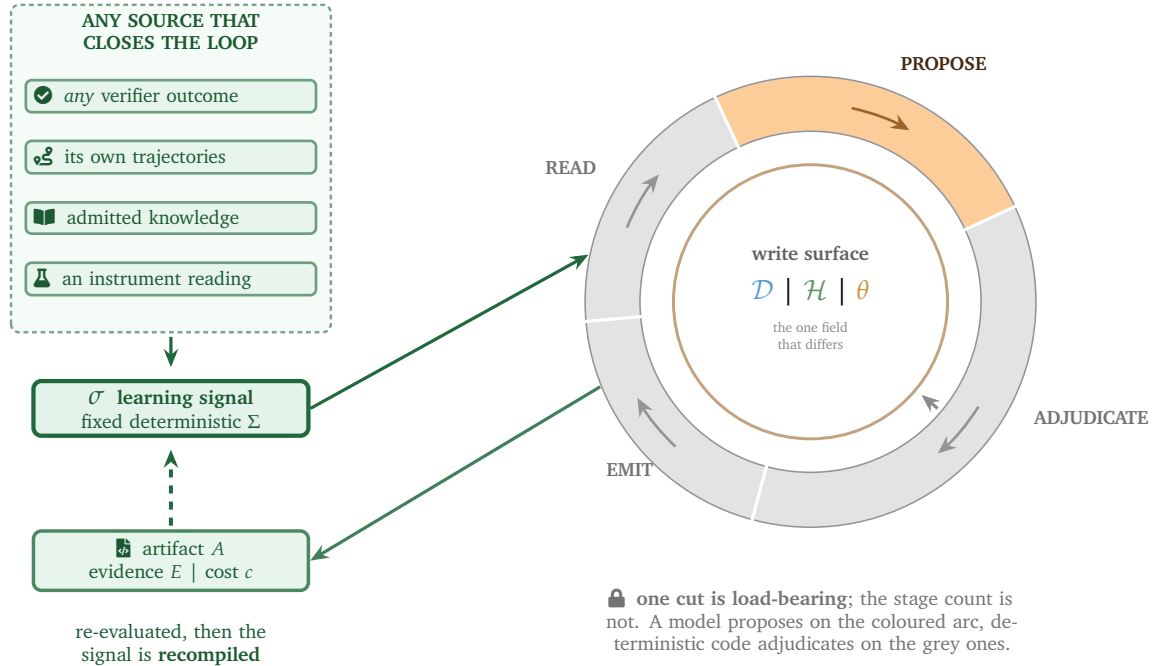
\begin{figure}[!b]
\centering
%
\input{sections/fig_palette}%
\resizebox{0.96\linewidth}{!}{%
\begin{tikzpicture}[
  x=1mm, y=1mm,
  font=\footnotesize,
  chip/.style={rounded corners=2.2pt, draw=figAline!65, line width=1.0pt,
               fill=figA!12, align=left, inner xsep=3pt, inner ysep=2.2pt,
               font=\scriptsize, text=figAline!85!black, text width=35mm},
  card/.style={rounded corners=3pt, draw=black!52, line width=1.2pt,
               fill=white, align=center, inner xsep=3pt, inner ysep=3pt,
               font=\scriptsize},
  ar/.style={-{Stealth[length=2.4mm,width=1.9mm]}, line width=1.2pt,
             draw=black!40},
  band/.style={line width=7mm},
  sep/.style={white, line width=1.2pt},
  dirar/.style={-{Stealth[length=2.4mm,width=2.0mm]}, line width=1.0pt},
  slab/.style={font=\scriptsize\bfseries, text=black!55}
]
\path (0,11.5) rectangle (168,104);

\draw[rounded corners=3pt, draw=figAline!60, line width=0.9pt, fill=figA!5,
      dash pattern=on 1.8pt off 1.4pt] (1,58) rectangle (43,101);
\node[font=\scriptsize\bfseries, text=figAline!85!black, align=center,
      text width=40mm] at (22,97.5) {ANY SOURCE THAT CLOSES THE LOOP};
\node[chip, text width=36mm] (c1) at (22,89) {\faCheckCircle\; \emph{any} verifier outcome};
\node[chip, text width=36mm] (c2) at (22,81) {\faRoute\; its own trajectories};
\node[chip, text width=36mm] (c3) at (22,73) {\faBookOpen\; admitted knowledge};
\node[chip, text width=36mm] (c4) at (22,65) {\faFlask\; an instrument reading};

\node[card, fill=figA!14, draw=figAline, line width=1.6pt, text width=34mm,
      text=figAline!85!black] (sig) at (22,48)
  {{\normalsize$\sig$}\ \ \textbf{learning signal}\\[0.6pt]
   {\scriptsize fixed deterministic $\Sigma$}};
\draw[ar, draw=figAline] (22,57.4) -- (22,53.4);

\node[card, fill=figE!12, draw=figEline!80, line width=1.3pt, text width=34mm,
      text=figEline!85!black] (out) at (22,28)
  {\faFileCode\; artifact $A$\\evidence $E$ \textbar\ cost $c$};
\draw[ar, dashed, draw=figAline, line width=1.5pt] (22,32.6) -- (22,42.6);
\node[font=\scriptsize, text=figAline!85!black, align=center, text width=34mm]
  at (22,17) {re-evaluated, then the signal is \textbf{recompiled}};

\begin{scope}[shift={(106,62)}]
  \draw[band, draw=figB!55]  (115:26mm) arc (115:25:26mm);    
  \draw[band, draw=black!11]    (185:26mm) arc (185:115:26mm);   
  \draw[band, draw=black!11]    (25:26mm)  arc (25:-105:26mm);   
  \draw[band, draw=black!11]    (-105:26mm) arc (-105:-175:26mm);
  \draw[draw=black!42, line width=0.6pt] (0,0) circle (29.6mm);
  \draw[draw=black!42, line width=0.6pt] (0,0) circle (22.4mm);
  \foreach \a in {185,115,25,-105}
    {\draw[sep] (\a:22.2mm) -- (\a:29.8mm);}
  \draw[dirar, draw=figBline] (78:26mm) arc (78:60:26mm);
  \foreach \a in {158,-32,-133}
    {\draw[dirar, draw=black!42] (\a:26mm) arc (\a:\a-18:26mm);}
  \node[circle, fill=white, draw=figBline!60, line width=1.1pt,
        minimum size=36mm] at (0,0) {};
  \node[align=center, font=\scriptsize, text=black!62] at (0,6.4mm)
    {\textbf{write surface}};
  \node[align=center, font=\normalsize] at (0,1.4mm)
    {\textcolor{OpDataC}{$\Dat$}\;\textbar\;\textcolor{OpHarnC}{$\Harn$}%
     \;\textbar\;\textcolor{OpModC}{$\theta$}};
  \node[align=center, font=\tiny, text=black!45] at (0,-4.4mm)
    {the one field\\that differs};
  \draw[-{Stealth[length=2.4mm,width=1.9mm]}, line width=1.2pt, draw=black!45]
    (-40:21.8mm) -- (-40:19.0mm);
  \node[slab, anchor=south east] at (150:31mm) {READ};
  \node[slab, anchor=south west, text=figBline!42!black] at (70:31mm) {PROPOSE};
  \node[slab, anchor=north west] at (-25:31mm) {ADJUDICATE};
  \node[slab, anchor=north]      at (-140:31mm) {EMIT};
  \coordinate (entry) at (168:29.8mm);
  \coordinate (exit)  at (-158:29.8mm);
\end{scope}

\draw[ar, draw=figAline]           (sig.east) -- (entry);
\draw[ar, draw=figEline!85] (exit) -- (out.east);

\node[font=\scriptsize, text=black!55, align=center, text width=76mm]
  at (106,20) {\faLock\; \textbf{one cut is load-bearing}; the stage count is
  not. A model proposes on the coloured arc, deterministic code adjudicates on
  the grey ones.};
\end{tikzpicture}}
\caption{\textbf{The loop kernel of \Cref{def:operator}.} The operator is a ring; the hub is the \textbf{write surface} (\textcolor{OpDataC}{$\Dat$}, \textcolor{OpHarnC}{$\Harn$}, \textcolor{OpModC}{$\theta$}), the one field that differs across the three operators. The ring carries \textbf{one authority boundary}: a model proposes on the coloured arc, deterministic code adjudicates, emits and reads on the grey ones. The cut is load-bearing; the seven-stage refinement of \Cref{eq:kernel} is not unique. The cycle closes \emph{through the signal}: verifier outcomes, trajectories, admitted knowledge and instrument readings all feed the fixed $\Sigma$, and the dashed arc recompiles it after each re-evaluation.}
\label{fig:kernel}
\end{figure}

\noindent Everything the definition does not mention is left to the domain: how many stages the operator has, whether it searches or samples, whether it is a language model or a solver or a piece of numerical code. An operator is admitted to the framework solely by \emph{surrendering adjudication}, regardless of how it is built internally. Two structural commitments follow, and they are the only ones the framework makes. The first is \textbf{cyclic closure}. Condition~\emph{(i)} makes $\sig$ the operator's sole input, and by the perishability property of \Cref{sec:prelim} whatever the operator exports is re-evaluated and recompiled into that same signal; every output therefore returns to the object that drove it,
\begin{equation}
\sig' \;=\; \Sigma(E \cup E'),
\qquad
E' \;=\; \mathrm{rollouts}(\widetilde{\Sys},\, \mathcal{T}_{\mathrm{adapt}}),
\label{eq:recompile}
\end{equation}
so that what it exported is measured on the system it produced rather than on the one it started from. The second is that the cycle carries exactly one \textbf{authority boundary}, cutting it into an arc on which a model proposes and an arc on which deterministic code adjudicates. \Cref{fig:kernel} draws both commitments: \emph{where} that cut falls is load-bearing, and how finely either arc is subdivided is not.

Our reference instantiation of \Cref{def:operator} is the \textbf{loop kernel} $\Kernel$, instantiated by all three operators below. It refines the two arcs into seven named stages:
\begin{equation}
\Kernel \;=\;
\textsc{Observe} \!\to\!
\textsc{Diagnose} \!\to\!
\textsc{Propose} \!\to\!
\textsc{Validate} \!\to\!
\textsc{Execute} \!\to\!
\textsc{Select} \!\to\!
\textsc{Export},
\label{eq:kernel}
\end{equation}
and exposes exactly one mutable object, its \textbf{proposal policy} $\pi$. An operator is then fully specified by the triple:
\begin{equation}
U \;=\; \langle\, \Kernel,\, \pi,\, w \,\rangle,
\qquad
w \in \{\Dat,\, \Harn,\, \theta\},
\label{eq:operator-pair}
\end{equation}
which refines the generic procedure of \Cref{eq:operator} by naming the part that may be improved.

On the mutable arc, \textsc{Diagnose} and \textsc{Propose} are \emph{model-driven}: a language model reads the learning signal and the current component state, attributes the observed failures, and emits a candidate modification together with an expected effect and a risk statement. The remaining five stages are \emph{deterministic code}: \textsc{Observe} binds the typed inputs the operator declared it consumes; \textsc{Validate} checks the proposal against a schema and rejects any edit outside the operator's declared write surface; \textsc{Execute} applies the surviving proposal in an isolated workspace; \textsc{Select} scores candidates with a fixed evaluator and promotion rule; and \textsc{Export} emits a typed artifact carrying a content hash, its parents, the supporting evidence, and the realized cost. \textsc{Select} and \textsc{Export} run on protected code outside the operator's write surface (\Cref{sec:method:governance}). The model carries the entire semantic burden of deciding \emph{what} is wrong and \emph{what} to change, while deterministic code holds every authority to decide whether the change was good; this division is what makes an operator safe to place under an automatic scheduler.

The kernel consumes exactly one environmental input, the learning signal $\sig$ of \Cref{eq:signal}, whose sources are unconstrained and whose three-layer typing is fixed. The concrete schema of that typing, for every failed rollout, is a \textbf{failure signature}:
\begin{equation}
\varphi(\Traj) \;=\; \big(\,
\ubl{SoftGray}{\chi(\Traj)}{terminal cause}\;,\;\;
\ubl{SoftGray}{\psi(\Traj)}{causal role}\;,\;\;
\annot{accent}{\mu(\Traj)}{reusable mechanism}
\,\big),
\label{eq:signature}
\end{equation}
where the tint is the authority boundary of \Cref{def:operator} carried into the schema: $\chi$ and $\psi$ are set in grey because they are grounded in the trace and the verifier report under schema and evidence constraints, and only the boxed $\mu$ is the model's own hypothesis. The first two record what the evaluator reported and what the agent did: an output-protocol violation, a tool-call mismatch, a derivation inconsistent with the emitted answer; the third is the generalizable mechanism behind the failure. The same signature vocabulary is shared by all three operators, so a failure attributed to a missing procedural habit can be routed to the harness while one attributed to absent knowledge is routed to data and weights; the routing decision of \Cref{sec:scheduling} is made on an object all operators can read.

\Cref{def:operator} fixes neither the signal compiler $\Sigma$ nor the verifier $v$; promoting either to an operator's write surface is architecturally licensed but governed out here, since a system permitted to improve what counts as success can raise its score without improving capability, a failure mode catalogued for self-evolving agents under deployment-time reward hacking~\citep{misevolve2026}. The sealed measurement, its task set, the release rule, and the ledger stay outside every write surface at every level (\Cref{sec:method:governance}); an evolving verifier is admitted only as a way of widening coverage \emph{between} sealed evaluations.

The authority boundary of \Cref{def:operator} is uniform across every
model-driven component: the model supplies the semantics, and deterministic code
supplies every quantity that enters a promotion decision.
The same split holds for the scheduler itself: the \rsiag supplies cross-layer
diagnosis, operator choice, and axis selection, while deterministic code supplies
the admissibility filter, the program type-check, and the contract and diff
checks of \Cref{eq:accept}.

%

\subsection{Three Base Operators}
\label{sec:method:operators}

Instantiating \Cref{eq:kernel} on the three components of $\Sys$ gives three operators whose write surfaces are disjoint by construction. The remainder of this subsection develops each in turn. The three follow from \Cref{eq:system} once the target system is written down: a deployed system \emph{is} a data state, a model state and a harness state, so an operator that changes what the system does writes exactly one of the three. The three also stand in a fixed cost order (model calls, then model calls plus replay, then GPU-hours), and it is that ordering which makes the scheduling problem of \Cref{sec:scheduling} non-trivial.

\begin{glancebox}
\gitem{OpDataC}{\suitA}{\dataop\ -- amplifier}Synthesizes verified records from the system's own execution and marks where that experience ends, so supervision is spent only past the boundary; its products feed both of the others.
\gitem{OpHarnC}{\suitB}{\harnessop\ -- scaffold}Edits the five-slot execution scaffold under a typed patch, adding capability without touching the weights. Cheapest per unit of gain; every addition is re-paid as context at inference.
\gitem{OpModC}{\suitC}{\modelop\ -- internalization}Converts a recurring context cost into a one-time training cost under a bounded recipe. Most durable, most expensive; needs a dataset postdating the last capability change.
\end{glancebox}


\subsubsection{\dataop: Verified Synthesis from Execution Experience}
\label{sec:method:data}

\dhl{\dataop} writes the data state $\Dat$. It consumes execution trajectories from the kernel's \textsc{Execute} stage and returns a \textsc{Dataset} artifact of verified training records, which both of the other operators consume. Its central principle is that synthesis is \textbf{anchored to observed execution rather than invented}: every record descends from experience distilled from actual rollouts, so synthesis stays within what observed rollouts support. The operator presses out capability the model already holds and marks where that capability ends. For each episode $e$ the operator extracts a four-dimensional \textbf{learning signature}:
\begin{equation}
\kappa(e) \;=\; \big(\,
\ubl{OpDataC}{\kappa_{\mathrm{k}}(e)}{knowledge},\;
\ubl{OpHarnC}{\kappa_{\mathrm{r}}(e)}{reasoning},\;
\ubl{mTeal}{\kappa_{\mathrm{v}}(e)}{verification},\;
\ubl{accentlt}{\kappa_{\mathrm{d}}(e)}{distractor}
\,\big),
\label{eq:kappa}
\end{equation}
whose components diagnose, in order, a knowledge deficit, a reasoning failure despite knowledge being present, a missing verification step, and a susceptibility to distractors. Signatures are the primary diagnosis carried by the \textsc{Experience} artifact and steer all downstream synthesis; they are a structured expansion of the model-attributed field $\mu(\Traj)$ of \Cref{eq:signature}, and downstream stages read them as directives about \emph{what kind} of record to build, not as scalar scores.

\subhead{Synthesis pipeline} Trajectories become verified records through four stages: experience extraction, directive generation, adversarial generation, and verification. Episodes are grouped by outcome and domain, and each group is distilled into an \textsc{Experience} artifact that persists across rounds, accumulating diagnoses that later synthesis draws on. From each \textsc{Experience} artifact a \textsc{QueryGen} role emits directives, each naming a target capability and an independent verification approach. The pipeline composes as
\begin{equation}
\mathcal{E}
\;\xrightarrow{\;\textsc{Extract}\;}\; \mathcal{P}
\;\xrightarrow{\;\textsc{Direct}\;}\; \Delta
\;\xrightarrow{\;\textsc{AdvGen}\;}\; \mathcal{C}_{\mathrm{cand}}
\;\xrightarrow{\;\textsc{Gate}\;}\; \Dat',
\label{eq:synth}
\end{equation}
where $\mathcal{E}$ denotes the sharded episodes, $\mathcal{P}$ the experience pool, $\Delta$ the directive set, $\mathcal{C}_{\mathrm{cand}}$ the candidate records, and $\Dat'$ the accepted dataset. The stages are kept separate so that the diagnosis of what is wrong, the decision of what to teach, and the construction of a specific record are each auditable and individually replaceable.

\subhead{Adversarial generation} The construction of a record separates authoring from validation. An \textbf{Operator} role authors each record from a directive; an \textbf{Anchor} role receives only the serialized item, solves it without seeing the Operator's answer, and approves only when its \emph{blind re-derivation} agrees with the draft. Both roles are played by the same target model (\Cref{sec:exp:setup}); the separation here is by \emph{input isolation}, not by model identity, which prevents the generator from feeding its answer to the verifier while leaving shared model errors uncorrected:
\begin{equation}
(q,a) \sim O(\delta,\varepsilon,\kappa),
\qquad
\hat{a} \sim A(q),
\qquad
\mathrm{acc} \;=\; \mathbb{1}\big[\hat{a}\equiv a\big].
\label{eq:anchor}
\end{equation}
Rejection returns the item for revision up to a fixed bound, and items that do not converge within it are discarded. A synthesized record is thus certified by a blind re-derivation under input isolation rather than by the generator's own confidence, which is what lets \dataop write records in domains with no machine-checkable key while keeping certification decoupled from the generator. This certifies self-consistency under a blind re-solve; it does not certify correctness against knowledge the model itself lacks (\Cref{sec:discussion:open}, \Cref{law:scarcity}).

\subhead{Verification} Records that survive adversarial generation pass an independent verifier conjoining a deterministic contract check with a model-driven semantic check:
\begin{equation}
G(r) \;=\; \ubl{OpDataC}{C_{\mathrm{contract}}(r)}{schema, deterministic} \;\wedge\;
\ubl{mTeal}{C_{\mathrm{verify}}(r)}{semantics, re-solved}.
\label{eq:gates}
\end{equation}
The contract enforces the supervised-finetuning schema with exactly one trainable target. The semantic check independently re-solves the item and confirms self-containment, unambiguous options, a correct gold answer, and supporting reasoning. Both are \textbf{fail-closed}. Accepted records are deduplicated, screened against the sealed evaluation set, and partitioned into train and validation splits with parent lineage to their source \textsc{Experience}.

\subsubsection{\harnessop: Editing the Execution Scaffold}
\label{sec:method:harness}

\hhl{\harnessop} writes the harness state $\Harn$ without touching $\theta$. It changes how the model completes a task by editing the execution scaffold, and its gains take effect at once with no training cost. The harness is represented as a \textbf{genome} over five slots:
\begin{equation}
\Harn \;=\; \big(\underbrace{I}_{\text{system prompt}},\;\;
\underbrace{\mathcal{M}}_{\text{memory}},\;\;
\underbrace{\mathcal{O}}_{\text{built-in tools}},\;\;
\underbrace{\mathcal{S}}_{\text{skills}},\;\;
\underbrace{\mathcal{T}}_{\text{MCP}}\big),
\label{eq:genome}
\end{equation}
where $I$ is the system prompt carrying the task protocol and output format, $\mathcal{M}$ a bounded memory of typed entries injected under a retrieval policy, $\mathcal{O}$ the built-in tools and their schemas, $\mathcal{S}$ named procedural content loaded on demand, and $\mathcal{T}$ the tools and resources mounted through the Model Context Protocol, whether local or remote. Context, tool, and scratchpad policies govern how these slots are surfaced at inference. Each slot is a distinct \emph{edit granularity}, so a proposal can target one surface without disturbing the others, which makes the edit surface enumerable and a proposal a typed patch over named fields rather than a rewritten program.

\subhead{Structured patches} The operator consumes a parent genome, the learning signal, and distilled experience, and emits a typed \textsc{HarnessPatch}:
\begin{equation}
\Pi \;=\; \big(\eta,\; \{o_i\},\; \varepsilon,\; \mathcal{R}\big),
\qquad
o_i = \big(\mathrm{slot}_i,\; \mathrm{op}_i,\; v_i\big),
\label{eq:patch}
\end{equation}
where $\eta$ is the repair \emph{hypothesis} stating which failure the patch addresses, each operation names a target slot, a mutation type and a value, $\varepsilon$ is the \emph{expected effect}, and $\mathcal{R}$ the \emph{risk} statement. Requiring the hypothesis and predicted effect forces the model to state what it thinks is wrong and what it expects to happen, in a form a later stage can check against the outcome.

Deterministic validation enforces the write surface: patches are confined to the five declared slots, while the model provider, target weights, evaluator, sandbox, and release rule lie outside it, and any operation beyond the declared slot set is rejected mechanically. Memory entries carry a fixed type (knowledge, positive pattern, anti-pattern), and consolidation that removes or merges entries logs every displaced entry and its reason, so the scaffold's history is auditable.

\subhead{Evaluation and promotion} Search is organized around failure signatures rather than aggregate score. Failures sharing a signature form bounded shards, and each shard is diagnosed and patched independently. Candidates are replayed against a frozen baseline on their own shard, then combined into a single genome under deduplication and a hard complexity bound, and evaluated on the full adaptation set:
\begin{equation}
\Harn^{\star} = \operatorname*{arg\,max}_{\Harn \in \mathcal{H}_{\mathrm{merged}}} Q(\Harn)
\quad\text{s.t.}\quad
Q(\Harn^{\star}) > Q(\Harn_{\mathrm{best}}),
\qquad
\|\Harn^{\star}\| \le B_{\Harn}.
\label{eq:promote}
\end{equation}
Promotion requires \emph{strict} improvement over the historical best; otherwise the incumbent is retained. The complexity budget $B_{\Harn}$ constrains active context, memory count, and serialized genome size, preventing the scaffold from growing without bound. The operator also emits the trajectories that succeeded under the candidate genome, which become the verified trajectories \dataop consumes (the $\opH\!\to\!\opD$ adapter of \Cref{sec:composition}).

\subsubsection{\modelop: Internalizing Capability into the Model}
\label{sec:method:model}

\mhl{\modelop} writes the model state $\theta$. It consumes the \textsc{Dataset} artifact produced by \dataop and internalizes verified records into parameters through a bounded training recipe. The operator spans both trainable weights and model architecture within one bounded proposal space: weight updates cover adapter parameters over selected target modules, while architectural decisions cover adapter placement, layer-level participation, and the trainable parameter set itself. Where \harnessop adds capability to the scaffold at zero training cost and re-pays it as context at every inference, \modelop pays a one-time training cost that leaves the capability free at inference and persistent across scaffolds. The proposal space is a bounded \textbf{recipe},
\begin{equation}
\mathcal{R} \;=\; \mathcal{R}_{\mathrm{arch}} \times \mathcal{R}_{\mathrm{adapt}} \times \mathcal{R}_{\mathrm{optim}} \times \mathcal{R}_{\mathrm{seq}} \times \mathcal{R}_{\mathrm{ckpt}},
\label{eq:recipe}
\end{equation}
where $\mathcal{R}_{\mathrm{arch}}$ declares the trainable parameter set and the bounded architectural choices (adapter placement, target-module selection, and layer participation); $\mathcal{R}_{\mathrm{adapt}}$ fixes the adaptation family and its capacity, such as rank and scaling (low-rank adaptation~\citep{hu2022lora} in the configuration used here); $\mathcal{R}_{\mathrm{optim}}$ specifies learning rate, schedule, batch size, and gradient accumulation; $\mathcal{R}_{\mathrm{seq}}$ specifies sequence length; and $\mathcal{R}_{\mathrm{ckpt}}$ specifies checkpoint interval and early-stopping patience. Bounding the space over a declared, enumerable set of knobs spanning architecture and parameter choices makes the operator analyzable, and each recipe is checked against the training backend's capability declaration before submission. Given the fixed base checkpoint $\theta_0$, the cumulative dataset $\Dat'$ from \dataop, and a recipe $r_j \in \mathcal{R}$, the operator produces a candidate
\begin{equation}
\theta_j \;=\; \mathrm{Train}\big(\theta_0,\; \Dat',\; r_j\big),
\qquad j = 1,\dots,k,
\label{eq:train}
\end{equation}
where $k$ is bounded by the training budget. Training always proceeds from the same fixed base $\theta_0$ over the cumulative dataset rather than continuing a previous round's adapter, so successive generations differ only in their data and their recipe and share one optimization history, which keeps gains attributable and holds error accumulation in check across rounds. The dataset is materialized with a loss mask that credits only assistant-generated positions, and optimization steps are bounded by dataset size, holding repeated exposure within a fixed number of passes.

Evaluation and promotion sit outside the operator. An independent evaluation layer scores each $\theta_j$ on the sealed adaptation set under the same fixed evaluator $Q$ used for \harnessop, and a protected release rule decides
\begin{equation}
\theta^{\star} = \operatorname*{arg\,max}_{\theta_j} Q(\theta_j)
\quad\text{s.t.}\quad
Q(\theta^{\star}) > Q(\theta_{\mathrm{best}}).
\label{eq:release}
\end{equation}
A candidate that fails to strictly improve the historical best is discarded and the incumbent retained; this is the same strict-improvement discipline as \Cref{eq:promote}, with candidate cost the only difference. \Cref{tab:actionsurfaces} groups each operator's action surface into field classes. Two properties follow from it. The surfaces are disjoint, so \textsc{Validate} settles write-surface membership mechanically rather than by judgement, and enumerable, so every proposal takes the form of a typed patch over named fields; an action surface of ``arbitrary code'' would defeat both deterministic validation and automatic scheduling.

\begin{table}[t]
\centering
\resizebox{\linewidth}{!}{%
\renewcommand{\arraystretch}{1.22}
\setlength{\tabcolsep}{7pt}
\begin{tabular}{l l l}
\toprule
\headrow
\dcell{\dataop}\;\textbf{on} $\Dat$ & \hcell{\harnessop}\;\textbf{on} $\Harn$ & \mcell{\modelop}\;\textbf{on} $\theta$ \\
\midrule
synthesis directives \& generation quotas    & system prompt \& prompt templates                & trainable parameter set \& module architecture \\
capability targets \& difficulty distribution & persistent memory \& retrieval policy           & adapter placement \& layer participation \\
per-record verification standards            & built-in tools \& schemas; MCP-mounted tools \& resources & adaptation family, rank \& scaling \\
curriculum stages \& mixing ratios           & skill library                                    & optimizer, learning-rate schedule \& batch budget \\
accumulation, consolidation \& data splits   & context, tool-use \& scratchpad policies         & sequence length, checkpointing \& seed \\
\bottomrule
\end{tabular}}
\caption{\textbf{Concrete action surfaces.} The three operators act on disjoint field sets of three different system components, the synthesized data for \dhl{\dataop}, the task scaffold for \hhl{\harnessop}, and the trainable parameters and architecture for \mhl{\modelop}, so \textbf{every promoted change is attributable to exactly one operator}.}\label{tab:actionsurfaces}
\end{table}

The cost structure complements \harnessop's. Writing $\Delta_{\mathrm{ctx}}$ for the change in context consumed per task,
\begin{equation}
\ubl{OpHarnC}{\Delta_{\mathrm{ctx}}(\opH) > 0}{re-paid every inference},
\qquad
\ubl{OpModC}{\Delta_{\mathrm{ctx}}(\opM) = 0}{paid once, at training},
\label{eq:ctxcost}
\end{equation}
this asymmetry makes the two operators complementary, and the $\opM\!\to\!\opH$ adapter of \Cref{sec:composition} exploits it by retiring scaffold extensions once their behaviour is internalized. The operator emits a \textbf{training report} recording each recipe, its candidate checkpoint, and the training metrics; a promoted checkpoint re-enters the kernel's \textsc{Execute} stage as the new base for subsequent rollouts.

\subsection{Operator Composition: Five Admissible Transitions}
\label{sec:composition}

Given three operators, an improvement sequence is a word $z=(u_1,\dots,u_T)$ over $\{\opD,\opH,\opM\}$. Some words are ill-posed rather than merely inefficient: they ask an operator to consume an artifact that no longer describes the system it will be applied to. A single condition makes this precise.

\begin{definition}[Capability-altering step]
A step $u$ is \emph{capability-altering} if executing it changes the behavior of the deployed system. Under \Cref{eq:system}, $\opH$ and $\opM$ are capability-altering; $\opD$ is not, since it writes only the data state and emits artifacts for later consumption.
\end{definition}

\begin{definition}[Signal freshness]
An artifact is \emph{fresh} at step $t{+}1$ when the signal it derives from was compiled no earlier than the most recent capability-altering step:
\begin{equation}
\mathrm{fresh}(a,\, t+1) \iff \mathrm{compiled}(a) \;\ge\; \max\{\, t' \le t : u_{t'} \text{ is capability-altering} \,\},
\qquad \max\varnothing = 0.
\label{eq:fresh}
\end{equation}
\end{definition}

\begin{principle}[Admissibility]
\label{prin:admissibility}
A step $u_{t+1}$ is admissible after $z_{1:t}$ if and only if every artifact it consumes is fresh.
\end{principle}

\dataop and \harnessop consume the learning signal itself, which the kernel recompiles through a fixed evaluation pass after every capability-altering step; they are therefore admissible after any prefix. \modelop consumes a \textsc{Dataset} artifact, fresh only when produced by a \dataop step with no capability-altering step in between. Enumerating the six ordered pairs of distinct operators under Principle~\ref{prin:admissibility}, together with the three self-transitions, partitions the nine into
\begin{equation}
\begin{aligned}
\mathcal{A}_{\mathrm{legal}} &= \{\,\opD\!\to\!\opH,\; \opD\!\to\!\opM,\; \opH\!\to\!\opD,\; \opM\!\to\!\opD,\; \opM\!\to\!\opH,\; \opD\!\to\!\opD,\; \opH\!\to\!\opH\,\},\\[2pt]
\mathcal{A}_{\mathrm{forbidden}} &= \{\,\opH\!\to\!\opM,\; \opM\!\to\!\opM\,\},
\end{aligned}
\label{eq:admissible}
\end{equation}
exactly five admissible cross-operator transitions and two admissible self-transitions. The single cross-operator exclusion is $\opH\!\to\!\opM$: a harness step changes what the deployed system can do but produces no data, so the most recent dataset necessarily predates that change, and training on it would internalize behavior the system has already superseded. The same condition rules out the self-transition $\opM\!\to\!\opM$, while $\opD\!\to\!\opD$ and $\opH\!\to\!\opH$ remain admissible as the iterated-search regimes of the individual operators. \Cref{fig:transitions} draws the resulting graph.

\begin{figure}[!t]
\centering
\begin{minipage}[c]{0.42\linewidth}
\centering
\resizebox{\linewidth}{!}{%
\begin{tikzpicture}[
  font=\footnotesize,
  op/.style={circle, draw, line width=1.4pt, minimum size=11mm, align=center, font=\bfseries\small},
  dop/.style={op, fill=bBlue, draw=mBlue, text=OpDataC},
  hop/.style={op, fill=bGreen, draw=mGreen, text=OpHarnC},
  mop/.style={op, fill=bOrange, draw=mOrange, text=OpModC},
  ok/.style={-{Stealth[length=2mm]}, line width=1.3pt, draw=mGrey!150},
  okd/.style={ok, draw=OpDataC!85},
  okh/.style={ok, draw=OpHarnC!85},
  okm/.style={ok, draw=OpModC!85},
  bad/.style={-{Stealth[length=2mm]}, line width=1.3pt, draw=MetaC, dashed},
  el/.style={font=\scriptsize, inner sep=1pt, fill=white, fill opacity=0.85, text opacity=1}
]
\node[dop] (D) at (90:17mm)  {\dcell{$\opD$}};
\node[hop] (H) at (207:17mm) {\hcell{$\opH$}};
\node[mop] (M) at (333:17mm) {\mcell{$\opM$}};

\draw[okd] (D) to[bend right=14] node[el, pos=0.52, left=0.4mm] {\smash{\raisebox{0pt}{1}}} (H);
\draw[okh] (H) to[bend right=14] node[el, pos=0.52, right=0.4mm] {3} (D);
\draw[okd] (D) to[bend left=14]  node[el, pos=0.52, right=0.4mm] {2} (M);
\draw[okm] (M) to[bend left=14]  node[el, pos=0.52, left=0.4mm] {4} (D);
\draw[okm]  (M) to[bend left=14]  node[el, pos=0.5, below=0.4mm] {5} (H);
\draw[bad] (H) to[bend left=14]  node[el, pos=0.5, above=0.4mm, text=MetaC] {\ding{55}} (M);

\draw[okd]  (D) to[out=58,in=122,looseness=8]  (D);
\draw[okh]  (H) to[out=172,in=236,looseness=8] (H);
\draw[bad] (M) to[out=328,in=32,looseness=8]  (M);

\node[align=center, font=\scriptsize, below=8mm of H, xshift=15mm, text=black!60]
  {solid: admissible\\ \textcolor{MetaC}{dashed: violates freshness}};
\end{tikzpicture}}
\end{minipage}\hfill
\begin{minipage}[c]{0.55\linewidth}
\centering
\resizebox{\linewidth}{!}{%
\scriptsize
\renewcommand{\arraystretch}{1.15}
\setlength{\tabcolsep}{3.5pt}
\begin{tabular}{c l l l}
\toprule
\headrow
\textbf{\#} & \textbf{Edge} & \textbf{Transition adapter} & \textbf{Type conversion} \\
\midrule
1 & $\opD\!\to\!\opH$ & Experience Extraction & \textsc{Experience} $\to$ \textsc{HarnessPatch} \\
2 & $\opD\!\to\!\opM$ & Dataset Materialization & \textsc{Dataset} $\to$ masked shards $+$ recipe \\
3 & $\opH\!\to\!\opD$ & Signal Recompilation & \textsc{Genome} $\to$ recompiled $\sig$ \\
4 & $\opM\!\to\!\opD$ & Signal Recompilation & \textsc{Checkpoint} $\to$ recompiled $\sig$, $\kappa$ \\
5 & $\opM\!\to\!\opH$ & Redundancy Reconciliation & \textsc{TrainingRpt.} $\to$ \textsc{HarnessPatch} \\
\midrule
\multicolumn{4}{l}{\textit{\footnotesize self-transitions: iterated search inside one operator}}\\
$\circlearrowleft$ & $\opD\!\to\!\opD$ & none required & directive re-issued on the same $\sig$ \\
$\circlearrowleft$ & $\opH\!\to\!\opH$ & none required & next shard replayed on the same $\sig$ \\
\midrule
\rowcolor{MetaC!7}
\ding{55} & $\opH\!\to\!\opM$ & \textcolor{MetaC}{inadmissible} & dataset predates the capability change \\
\rowcolor{MetaC!7}
\ding{55} & $\opM\!\to\!\opM$ & \textcolor{MetaC}{inadmissible} & the same condition, on a self-loop \\
\bottomrule
\end{tabular}}
\end{minipage}
\caption{\textbf{Operator composition.} Left: the transition graph over $\{\opD,\opH,\opM\}$. Five solid edges are admissible under Principle~\ref{prin:admissibility}, as are the self-loops $\opD\!\to\!\opD$ and $\opH\!\to\!\opH$; two \textcolor{MetaC}{red dashed} edges ($\opH\!\to\!\opM$, $\opM\!\to\!\opM$) violate signal freshness. Right: each admissible edge is realized by a typed \transag adapter, and the last column gives that adapter's \emph{type} signature rather than an instance of its payload: a \textsc{HarnessPatch} may target any of the five slots of \Cref{eq:genome}, and what a conversion carries inside a declared type is unconstrained. \textbf{Self-transitions require no adapter}.}
\label{fig:transitions}
\end{figure}

An admissible edge declares a composition well posed, not yet executable: the producing operator's output format still has to be converted into the consumer's input. Each admissible edge is therefore realized by a \textbf{\transag} adapter, a small typed program that performs only this conversion,
\begin{equation}
T_{i\to j}:\; \mathcal{A}_i \to \mathcal{A}_j,
\qquad
T_{i\to j}(a_i) = a_j \iff \mathrm{fresh}(a_i) \,\wedge\, \mathrm{typecheck}(a_j,\, \mathcal{A}_j),
\label{eq:adapter}
\end{equation}
which gives an adapter grounds to \emph{reject}: an artifact stale by \Cref{eq:fresh}, or a conversion whose output fails the consumer's declared input type, is stopped before it reaches the consumer. \emph{Experience Extraction} ($\opD\!\to\!\opH$) distils \textsc{Experience} artifacts and verified trajectories into a typed \textsc{HarnessPatch}, whose operations may target any of the five slots of \Cref{eq:genome}. \emph{Dataset Materialization} ($\opD\!\to\!\opM$) resolves the curriculum into physical shards and applies the loss mask. \emph{Signal Recompilation} ($\opH\!\to\!\opD$) re-runs the fixed evaluator under the new genome and recompiles $\sig$, so that the amplifier measures the system as it now is. The $\opM\!\to\!\opD$ adapter performs the same \emph{Signal Recompilation} under new weights rather than a new genome, recompiling $\sig$ and the learning signatures $\kappa$ of \Cref{eq:kappa} on the released checkpoint; the two are one function, triggered by the two different capability-altering steps. \emph{Redundancy Reconciliation} ($\opM\!\to\!\opH$) reads the training report and proposes deletions of scaffold entries whose behavior the new weights have internalized, together with the additions those weights make worthwhile. Adapters carry no authority over the target system: they reformat, distill, or reject, and their outputs are typed artifacts with full lineage like any other.

\subsection{Two-Axis Scheduling}
\label{sec:scheduling}

The operators and their adapters define what may be done; the \rsiag decides what is done. At each decision point it reads the executed prefix $z_{1:t}$, the latest signal $\sig_t$, the evidence accumulated per operator, and the remaining budget, and chooses one of two axes.

\par\medskip\noindent \textbf{Horizontal orchestration extends the sequence.} The \rsiag draws the next ordered block of operators as a typed \emph{improvement program},
\begin{equation}
z_{t+1:t+m} \;\sim\; \pi^{\mathrm{h}}_{\phi}\big(\cdot \mid z_{1:t},\,\sig_t,\,\Budget_{\mathrm{rem}}\big),
\qquad
\mathrm{supp}\big(\pi^{\mathrm{h}}_{\phi}\big)\subseteq\mathcal{A}(z_{1:t}),
\label{eq:horizontal}
\end{equation}
whose support is the admissible set of Principle~\ref{prin:admissibility}. Each step names its operator and purpose, the edge that binds it to the previous step, the evidence that justifies it, and its share of the budget; the block as a whole names the evaluation and release criteria under which it will be judged and the conditions under which it stops. Admissibility is settled at planning time: the program is a valid word in the transition graph of \Cref{sec:composition}, with evidence and budget attached to each step. Planning is thereby separated from execution: the \rsiag decides what is done, and the runtime unfolds the declared order.

\par\medskip\noindent \textbf{Vertical optimization improves an operator instead of invoking one.} The \rsiag issues an independent update-or-skip directive for each operator,
\begin{equation}
\begin{aligned}
d_t \;&=\; \pi^{\mathrm{v}}_{\phi}\big(\cdot \mid z_{1:t},\,\sig_t\big)\\[3pt]
     &=\;\big(\,
        \ubl{CtrlC}{i}{target}\;,\;\;
        \ubl{accent}{\kappa}{contract}\;,\;\;
        \ubl{SoftGray}{a}{attribution}\;,\;\;
        \ubl{SoftGray}{J_i}{objective}\;,\;\;
        \annot{accentlt}{\Xi}{requested surfaces}
        \,\big),
\end{aligned}
\label{eq:vertical}
\end{equation}
with $\Xi$ the only field the acceptance checks below constrain. Each directive is routed to a dedicated \rsisub for its target operator. The \rsisub reads the directive, the operator's current policy $\pi_i$, and the portion of recent trajectories attributable to it, and returns a revised policy. The contract $\kappa$ partitions every field of the operator into three classes. \textbf{Mutable surfaces} are open to revision: the instructions governing how the operator diagnoses, what it prioritizes when proposing, and how it states its own applicability. \textbf{Action surfaces} are what the operator writes on the target system, identical to the write surfaces of \Cref{sec:method:operators}. \textbf{Protected surfaces} stay fixed: the evaluator, sandbox, release gate, artifact schemas, evidence chain, and the operator's identity and declared input--output types. A revised policy is installed once it clears four checks:
\begin{equation}
\mathrm{accept}(\pi_i',\, \pi_i,\, d_t) \iff
\begin{cases}
\mathrm{id}(\Xi) = \mathrm{id}(\pi_i), & \text{(identity)}\\[2pt]
\mathrm{modified}(\pi_i',\pi_i) \subseteq \mathrm{mutable}(\Xi), & \text{(declared-surface containment)}\\[2pt]
\mathrm{diff}(\pi_i',\pi_i) = \mathrm{declared\_mod}(d_t), & \text{(declared-modification equality)}\\[2pt]
\mathrm{stop}(\pi_i',\pi_i,d_t) = \mathrm{true}. & \text{(stop condition)}
\end{cases}
\label{eq:accept}
\end{equation}
The third is decisive in practice: requiring the actual diff to \emph{equal} the declared modification makes every change explicit, so vertical optimization takes the form of a controlled edit to a named policy, with the requested surfaces bounding both its scope and its audit trail.

The two axes consume the same evidence but act on different objects, realizing the two-level separation of \Cref{eq:operator-pair}: horizontal orchestration produces \emph{target deltas} $\Delta\Sys$, vertical optimization produces \emph{operator deltas} $\Delta\pi$. Keeping them distinct gives the empirical decomposition of \Cref{sec:exp} three explicit objects of attribution: the sequence, the operator, and their interaction.

\subsection{\oursplain: Improving the Scheduler}
\label{sec:meta}

The \rsiag of \Cref{sec:scheduling} is itself a policy, and nothing so far improves it. Its choices are as learnable as the operators' proposal policies: when to switch from orchestration to optimization, how much budget to commit before the evidence justifies a training step, when to stop, and how much confidence to place in a diagnosis. These are also the choices that determine whether the operators are used well, so we close one further loop. A complete improvement \textbf{term} is a sequence executed to a stop intent, evaluated under the protected \textbf{release gate}, and released or rejected (\Cref{sec:method:worked} states its lifecycle). After each term, a \metaag reads the term as a whole and proposes an update,
\begin{equation}
\phi \;\leftarrow\; M_{\psi}\big(\phi,\ \mathcal{H}_{\mathrm{term}}\big),
\qquad
\mathcal{H}_{\mathrm{term}}=\big(z,\ \sig_{1:T},\ \{\Delta\pi_i\},\
\ubl{mGreen}{g}{gate verdict},\ \ubl{SoftGray}{c}{realized cost}\big),
\label{eq:meta}
\end{equation}
restricted, like every other update in the framework, to the mutable surfaces of a contract: the \rsiag's diagnostic instructions, its operator-routing preferences, its budget and stopping policy, its choice between the two axes, and the guidance attached to each transition adapter. The \rsiag's decision space, the adapter endpoints, the evaluator, and the release gate are protected, so the meta layer changes how the \rsiag decides but not what it is permitted to decide. The meta agent carries fixed parameters $\psi$: it revises $\phi$ but is not itself revised, so the hierarchy terminates at this outermost level.

The perspective available to $M_\psi$ is what distinguishes it from the vertical \rsisub{}s. A \rsisub sees one operator's slice of one term and can only conclude that its operator should propose differently. The \metaag sees, per completed term, which sequences and vertical directives produced gains that survived the release gate, which candidates were generated but never selected, and how the budget was spent; on that basis it corrects systematic misallocations, such as weight updates committed on evidence that has historically not supported them, transitions attempted on insufficient upstream data, and vertical directives issued where a horizontal reordering would have been cheaper. This is the outermost of three nested improvement levels, and together they are what the name records: the framework applies recursive self-improvement \emph{to} recursive self-improvement, and that composition is what the \emph{Meta} names. The innermost is the \textbf{base loop}, ordinary \rsi (un-squared self-improvement), which modifies the target system; \textbf{two-axis orchestration} modifies the operators and how they are composed; and the \textbf{meta} level modifies the policy that governs both. The two deltas of the middle level are exactly the two axes of \Cref{sec:scheduling}: horizontal orchestration produces $\Delta z$, vertical optimization produces $\Delta\pi$. What keeps the levels apart is not convention but write surfaces, one per agent:
\begin{equation}
\begin{aligned}
\rsiag &:\ \mathrm{write}(\phi), &\qquad
\text{Operator } i &:\ \mathrm{write}(W_i) \subseteq \Sys, \\[2pt]
\rsisub_i &:\ \mathrm{write}(\mathrm{mutable}(\Xi_i)) \subseteq \pi_i, &\qquad
\metaag &:\ \mathrm{write}(\mathrm{instruction}(\phi)) \subseteq \phi,
\end{aligned}
\label{eq:writemasks}
\end{equation}
where $W_i$ is operator $i$'s write surface from \Cref{sec:method:operators}. No agent reaches across its level: the \rsiag writes only the scheduling policy, an operator only its assigned surface of $\Sys$, a \rsisub only the mutable surfaces of one operator's policy, and the \metaag only the instruction surface of $\phi$. This is why the framework needs four agents rather than one: the write surfaces are what make the levels compositional instead of merely recursive.

\subhead{Human in the loop} \Cref{eq:writemasks} names four \emph{roles}, not four models. Each is defined entirely by the evidence it reads, the surfaces it is permitted to write, and the contract its output must satisfy, so its occupant need not be a language model. Any of the four can therefore be filled by a \textbf{human expert}, with the framework keeping the same shape: the \rsiag choosing the next operator, a \rsisub rewriting one operator's proposal policy, the \metaag revising the scheduler, or a \transag converting one artifact into another.

The authority boundary of \Cref{sec:method:kernel} is indifferent to authorship: proposal, validation, evaluation and release each run identically whoever occupies the proposing role, so a human scheduler is audited on exactly the same terms as a model scheduler, with the same typed program, the same admissibility check of Principle~\ref{prin:admissibility}, the same ledger entry, and the same acceptance conditions of \Cref{eq:accept}. A domain expert reading a learning signal is often, at present, a better judge than any model of whether a failure is worth a training run.

The three operators are high-volume and mechanical, synthesizing thousands of records and replaying candidate patches per shard, so a human occupant would be the binding constraint rather than a source of judgement. The four agents make comparatively few decisions per term, and those decisions are exactly the ones that benefit from judgement. This is what makes a partly-manual deployment practical rather than merely possible, and it gives a spectrum: fully automatic at one end, a human holding the scheduler while the operators run automatically in the middle, and a human at every decision point at the other. In a new domain, where the framework's own evidence about what works is thin, the middle of that spectrum is where we would expect a first deployment to sit. The spectrum is also a trajectory rather than a fixed choice: early on a human supplies a comparatively large share of the high-quality guidance, framing the prompts and the calls at the agents' few decision points while the high-volume operators run automatically, and that share is meant to decrease as compiled evidence and promoted genomes accumulate, each role passing back to its agent once the framework's own record is strong enough, until the system reaches the fully automatic end. Human-in-the-loop is a waypoint on the path to autonomy rather than a permanent division of labour between person and machine. \Cref{tab:levels} names the three levels with the agent and the object of improvement belonging to each, since keeping them apart is what makes the composition comparisons of \Cref{sec:exp:main} well defined.

\begin{table}[t]
\centering
\resizebox{\linewidth}{!}{%
\renewcommand{\arraystretch}{1.30}
\setlength{\tabcolsep}{6pt}
\begin{tabular}{l l l}
\toprule
\headrow
\textbf{Level} & \textbf{Agent} & \textbf{Improves} \\
\midrule
Base loop & \dcell{\dataop} / \hcell{\harnessop} / \mcell{\modelop} operators &
the deployed system itself (data, harness, model) \\
Two-axis orchestration & \textcolor{CtrlC}{\textbf{\rsiag}}, with one \textcolor{CtrlC}{\textbf{\rsisub}} per operator &
how the operators are used: their order and their internal policies \\
Meta & \textcolor{CtrlC}{\textbf{\metaag}} &
how the \rsiag orchestrates \\
\bottomrule
\end{tabular}}
\caption{\textbf{The three nested improvement levels.} Each has its own agent, object and cadence, from every operator step at the base to once per term at the top, so \textbf{every gain is attributable to exactly one level}.}\label{tab:levels}
\end{table}

\subsection{Overall Operation: The Lifecycle of an Improvement Term}
\label{sec:method:worked}

The preceding subsections defined the kernel, the three operators, the admissible transitions, the two scheduling axes and the meta level. This one puts them together into a single \textbf{improvement term} as defined in \Cref{sec:meta}: an execution run from an initial system to a released successor. \Cref{alg:term} states its lifecycle as pseudocode, and the shape is the same each time. A term \emph{initializes} by compiling $\sig_0$ from the fixed evaluator on $\Sys_0$, the sealed set untouched. It then \emph{loops} while budget remains, the \rsiag routing each decision point to one of the two axes: horizontally, drawing an improvement program restricted to $\mathcal{A}(z_{1:t})$, compiling it by binding adapters, confirming artifact freshness under \Cref{eq:fresh}, and reserving budget, then running each step through the loop kernel; or vertically, emitting a directive whose resulting policy is installed once it clears \Cref{eq:accept}. After \emph{every} capability-altering step the evaluator re-runs and $\Sigma$ recompiles the signal, which is what keeps \Cref{eq:fresh} satisfied for the next decision. The term \emph{closes} at a stop intent, whether budget exhausted, convergence declared, or a failure intent, whereupon the \metaag reads the completed term and proposes an update to $\phi$ under \Cref{eq:meta}. It \emph{returns} the released successor, the updated operator policies, and the updated scheduler, so that operator-level gains compound within a term and meta-level gains compound across terms.

The pieces are easier to see in a concrete sequence. Suppose the initial system fails a subset of executable tasks, and $\Sigma$ compiles a signal in which the dominant deterministic terminal cause is \texttt{missing\_validation}: the system modifies state and declares completion without ever running the check that would have revealed the error.

\emph{Step 1 ($\opD$).} \dataop extracts learning signatures from the failing trajectories. Across the shard $\kappa_{\mathrm{v}}$ dominates while $\kappa_{\mathrm{k}}$ is near zero: the traces contain the knowledge that a check is required and do not run it, so the deficit is a habit rather than a fact. The operator distils the shard into an \textsc{Experience} artifact and synthesizes verified records demonstrating the missing habit through the adversarial pipeline of \Cref{eq:anchor}.

\emph{Step 2 ($\opD\!\to\!\opH$).} The \rsiag reads the diagnosis, notes that a verification-gap signature is the cheapest kind to address in the scaffold, and schedules \harnessop. Experience Extraction distils the records into two memory entries and one skill descriptor, deduplicated against the current genome.

\emph{Step 3 ($\opH$).} \harnessop proposes candidate patches per failure shard, replays each on its own shard, combines them into a single genome under deduplication and a hard complexity bound, and promotes it only after a strict improvement on the full adaptation split. Executable capability rises, and a new signal is recompiled.

\emph{Step 4 ($\opH\!\to\!\opD$).} Signal Recompilation re-scores under the promoted genome, and the failure pattern has shifted. \texttt{missing\_validation} is largely resolved, and the residue now carries a $\kappa_{\mathrm{k}}$ signature: a knowledge deficit the rollouts never exhibit, which lies beyond what \dataop can synthesize. Note that $\opH\!\to\!\opM$ is inadmissible at this point, because the dataset from Step 1 describes a system that no longer exists.

\emph{Step 5 ($\opD$, then $\opD\!\to\!\opM$).} \dataop runs again on the refreshed signal, this time drawing on the trajectories the promoted genome produced. This second pass does not synthesize the missing knowledge itself: it rebuilds a fresh dataset that postdates the last capability change, and it is \modelop that internalizes the $\kappa_{\mathrm{k}}$ residue through training. The dataset is therefore fresh and \modelop becomes admissible. Dataset Materialization stages the curriculum with a consolidation fraction reserved for capabilities already mastered.

\emph{Step 6 ($\opM$, then $\opM\!\to\!\opD$ and $\opM\!\to\!\opH$).} \modelop trains bounded candidate recipes from the fixed base; the independent evaluation layer scores them and the release gate promotes one. Signal Recompilation then re-scores under the new weights through the $\opM\!\to\!\opD$ adapter, after which Redundancy Reconciliation reads the training report, finds that $\kappa_{\mathrm{v}}$ no longer fires on the episodes the Step 3 entries were added for, and proposes deleting those entries; the deletion is replayed and, holding accuracy, kept.

\emph{Vertical interleaving.} At any decision point the \rsiag may substitute vertical optimization for the next horizontal step. If, for example, \dataop keeps issuing directives that later show no evaluation delta, the \rsiag issues a directive against \texttt{data.policy} rather than scheduling another operator, and the operator's proposal policy is rewritten within its contract: which signature dimensions it prioritizes, and how it turns a diagnosis into a directive.

\emph{Term close.} The term ends at a stop intent. The \metaag reads the whole trace, which steps produced released gains, which produced unselected candidates and how much budget each consumed, and revises the \rsiag's routing and budget instructions. The next term begins from the released successor with an updated $\phi$.

\begin{algorithm}[t]
\caption{\ours: one improvement term}
\label{alg:term}
\begin{algorithmic}[1]
\Require initial system $\Sys_0$, operators $\{U_i=\langle\Kernel,\pi_i,w_i\rangle\}$, \rsiag policy $\phi$, budget $\Budget$
\State $\sig_0 \gets \Sigma(\textsc{Evaluate}(\Sys_0))$ \Comment{fixed evaluator; sealed set untouched}
\For{$t=0,1,2,\dots$ \textbf{while} $\Budget_{\mathrm{rem}}>0$}
  \State $a_t \gets \textsc{Route}_\phi(z_{1:t},\sig_t)$ \Comment{horizontal or vertical}
  \If{$a_t = \textsc{Horizontal}$}
    \State draw program $z_{t+1:t+m}\sim\pi^{\mathrm{h}}_\phi$ restricted to $\mathcal{A}(z_{1:t})$ \Comment{Principle~\ref{prin:admissibility}}
    \State \textbf{compile:} bind adapters, confirm freshness under \Cref{eq:fresh}, reserve budget
    \For{each step $u$ in the program}
      \State $(\widetilde{\Sys},A,E,c) \gets U_u(\Sys_t,\sig_t,\Budget_u;\pi_u)$ \Comment{loop kernel, \Cref{eq:kernel}}
      \State $\Sys_{t+1} \gets \textsc{Gate}(\Sys_t,\widetilde{\Sys})$; \; publish $A,E$ to the lineage store
    \EndFor
  \Else
    \State $d_t \gets \pi^{\mathrm{v}}_\phi(z_{1:t},\sig_t)$; \; $\pi_{\mathrm{target}} \gets \textsc{SubAgent}(\pi_{\mathrm{target}}, d_t)$
    \State install $\pi'$ only if $\mathrm{accept}(\pi',\pi,d_t)$ of \Cref{eq:accept} holds \Comment{contract}
  \EndIf
  \State $\sig_{t+1} \gets \Sigma(\textsc{Evaluate}(\Sys_{t+1}))$ \Comment{signal recompiled after every step}
\EndFor
\State $\phi \gets M_\psi(\phi, \mathcal{H}_{\mathrm{term}})$ \Comment{meta update, \Cref{eq:meta}}
\State \Return released successor $\Sys_T$, updated $\{\pi_i\}$, updated $\phi$
\end{algorithmic}
\end{algorithm}

The invariant holding across all of it is that \textbf{no component ever evaluates its own output}: operators propose, an independent evaluation layer scores, and a protected release gate releases, while the \rsiag chooses what to run without scoring it and the \metaag revises how those choices are made without making them. This end-to-end auditability is the subject of the next subsection.

\subsection{What the Framework Protects}
\label{sec:method:governance}

A system that edits its own improvement machinery can inflate any metric it is also allowed to define, so three separations keep \Cref{eq:objective} meaningful. \emph{Proposal stays apart from validation and candidate generation from release}: the model diagnoses and proposes, while deterministic code validates, evaluates, and promotes at most one successor, which is the only object ever scored (\Cref{sec:method:kernel,sec:method:operators,sec:method:worked}). Above both, \emph{the sealed measurement lies outside every write surface at every level}, the meta layer included:
\begin{equation}
\{\Qsealed,\, \mathcal{T}_{\mathrm{seal}},\, \mathrm{release\_rule},\, \mathcal{L}\}
\cap W_i = \varnothing \ \ (\forall i),
\quad
\{\Qsealed,\, \mathcal{T}_{\mathrm{seal}},\, \mathrm{release\_rule},\, \mathcal{L}\}
\cap \big(\mathrm{write}(\phi) \cup \mathrm{write}(\phi_{\mathrm{meta}})\big) = \varnothing.
\label{eq:protected}
\end{equation}
Together with the lineage carried by each artifact, these separations make every in-term improvement auditable after the fact, the prerequisite for carrying the framework into domains whose verifiers are weaker than a test suite.

%% file: sections/05_experiments.tex
\section{Experiments}
\label{sec:exp}

\ours is defined over an arbitrary target system and admits any signal source that closes the loop, and the experiments instantiate that generality in the two regimes where capability is currently measured most sharply, executable coding and closed-form scientific reasoning, split into an \emph{executable} track in which a containerized test suite decides correctness and a \emph{closed-form} track in which an exact-match or numeric key does. In this section, we first fix the experimental setup, including the two target lines, the sealed-split protocol, and the closed-loop configuration in which every model-driven role is played by the model under test (\Cref{sec:exp:setup}). We then study the self-hosted open-weight target \textbf{Qwen3.5-35B-A3B}, where all three operators act, and ask whether scheduled composition outperforms the strongest single operator and the best hand-fixed pipeline under an equal budget (\Cref{sec:exp:main}). We then put leading frontier models, reached only through their provider interfaces, through the harness route of \dataop and \harnessop, and measure how much each gains by improving itself (\Cref{sec:exp:frontier}).

\subsection{Experimental Setup}
\label{sec:exp:setup}

\subhead{Target systems} The evaluation uses two \textbf{target lines} that correspond to two deployment regimes. The first is a \textbf{self-hosted open-weight target}, \textbf{Qwen3.5-35B-A3B}, a mixture-of-experts model with $35$B total and $3$B active parameters served through a fixed inference stack; its weights are writable, so all three operators, including \modelop, act on it (\Cref{sec:exp:main} and \Cref{sec:exp:terms}). The second is a line of \textbf{frontier} open- and closed-weight systems reached only through their provider interfaces, whose weights are not writable and on which improvement runs through the \dataop and \harnessop route (\Cref{sec:exp:frontier}). Runs on the self-hosted target use a $64$K context window. Every condition starts from the same clean initial system $\Sys_0$, with the same base checkpoint, seed genome, and empty data state, and executes through the same harness runtime, tool set, and scoring path. State is reset between conditions, which differ only in the genome they receive and in which operators may write. \modelop instantiates the LoRA point of the \Cref{tab:actionsurfaces} surface, drawing from the recipe space of \Cref{eq:recipe} under the contract of \Cref{app:contracts}.

\subhead{Closed-loop self-evolution} Every model-driven role is instantiated on the target model itself: the \textsc{Diagnose} and \textsc{Propose} stages of all three operators, the Operator and Anchor of adversarial generation, the \harnessop diagnosis roles, and the \rsiag and \metaag policies. No stronger external model proposes, synthesizes, judges, or schedules, and the training data contains no external model output; the sole external signal is the verifier or execution oracle that decides whether an attempt succeeded, which is the environment's own judgement. Keeping every proposer and judge seat on the target itself is what makes the reported gain attributable to the framework rather than to a stronger teacher, a sense in which an amplifier is nonetheless not a source that \Cref{sec:discussion:open} takes up; the same closed loop applies to every frontier target in \Cref{sec:exp:frontier}.

\subhead{Benchmarks} Executable capability is measured on Terminal-Bench~2.1~\citep{terminalbench2025} ($89$ tasks) and SWE-bench Pro~\citep{scale2026swebenchpro} (the $731$-task public test split), under a pinned dataset commit, a containerized sandbox, and a hidden verifier. All benchmarks are driven by our \textbf{reference execution harness} at a pinned release, one fixed implementation shared by the baseline, every successor, and every frontier target of \Cref{sec:exp:frontier}; this implementation is open-sourced as the standalone component \rsihar (\Cref{sec:discussion:harness}). It runs in its full agent configuration on the executable benchmarks and in a degenerate, prompt-only configuration on the closed-form benchmarks, where the built-in tool, skill, and MCP slots are disabled and edits are confined to the system-prompt slot. The verifier, sandbox, and sealed split are fixed and lie outside the write surface; the execution scaffold is what \harnessop edits. Closed-form scientific and mathematical reasoning is measured on GPQA-D-hard100 and a merged AIME set comprising the 2025 and 2026 AIME~I and AIME~II papers, four papers and $60$ problems. GPQA-D-hard100 is a fixed set of $100$ harder items drawn from GPQA-Diamond~\citep{rein2023gpqa}, chosen and frozen once before any improvement run and held out of both \dataop synthesis and \modelop training; the official GPQA Record ID of each item is listed in \Cref{tab:hard100} of \Cref{app:impl}. Both are scored pass@1 by exact or numeric match under a fixed decoding setting. Contamination is controlled by deterministic deduplication and independent model-based verification. The framework is verifier-agnostic and operates with any loop-closing signal source; the experiments instantiate it with the strong verifiers these domains provide.

\subhead{Budget, baselines, and metrics} All conditions receive identical totals of model tokens, GPU-hours, wall-clock time, operator invocations, candidates, and verifier queries, with the improvement budget $\Budget$ and the meta budget $\Budget_M$ metered on separate ledgers. Baseline and successor are scored under one identical pass@1 decoding setting, an over-budget response is scored as a failure rather than truncated and re-scored, and every reported number is the mean over five independent outer seeds. We compare against the frozen initial system (\emph{No improvement}), each single operator run with the full budget, a hand-fixed $\opD\!\to\!\opM\!\to\!\opH$ schedule (\emph{Human fixed}), a uniform sample over admissible sequences (\emph{Random composition}), and a signature-conditioned router without cross-term learning (\emph{Static router}). The primary metric is the deployed improvement productivity of \Cref{eq:objective}, reported on the released successor rather than the best archived candidate and scored once on the sealed split. Every result is reported per improvement term, and the multi-term runs of \Cref{sec:exp:terms} re-meter the same budget in each term, so a term-5 number is a fifth application of one budget rather than one application of five.

\subsection{Main Results}
\label{sec:exp:main}

\Cref{tab:main} reports the released successor's score on each sealed split. The comparison the table is built to support is not against the frozen system, which every condition beats, but against the best \emph{single} operator and the best \emph{hand-fixed} composition under the same budget. \ours is best on all four benchmarks, raising the average score by $10.9$ points over the frozen initial system. The gain is not carried by one operator: \dhl{\dataop}, \hhl{\harnessop} and \mhl{\modelop} each improve the frozen system when run alone, by $2.8$, $6.6$ and $3.9$ average points respectively, with \harnessop the strongest single operator, but scheduled composition adds a further $4.3$ points on top of it. The full system also beats both composition baselines, the hand-fixed $\opD\!\to\!\opM\!\to\!\opH$ pipeline and the static router, which are level at $+7.3$, by $3.6$ points, even though all three compose the same three operators over the same evidence and differ only in who decides the order. Scheduled composition is therefore not bookkeeping: the scheduling policy is itself a load-bearing component. The gains are largest on the closed-form suites, AIME $+13.3$ and GPQA-D-hard100 $+12.6$, and remain substantial on the executable benchmarks, Terminal-Bench~2.1 $+8.3$ and SWE-bench Pro $+9.2$, where the resolve rate nearly doubles from $10.3$ to $19.5$. Every gain is produced by the target model improving itself: no external model proposes, synthesizes, judges, or schedules, and the only external signal is the environment's own verifier. A $3$B-active model, operating entirely without a teacher, is sufficient to drive double-digit average self-improvement across both executable and closed-form domains.

\begin{table}[t]
\centering
\resizebox{\linewidth}{!}{%
\small
\renewcommand{\arraystretch}{1.22}
\setlength{\tabcolsep}{5pt}
\setlength{\aboverulesep}{0pt}
\setlength{\belowrulesep}{0pt}
\begin{tabular}{l cc cc cc cc c}
\toprule
\headrow
\textbf{Method}
 & \multicolumn{2}{c}{\textbf{Terminal-Bench 2.1}}
 & \multicolumn{2}{c}{\textbf{SWE-bench Pro}}
 & \multicolumn{2}{c}{\textbf{GPQA-D-hard100}}
 & \multicolumn{2}{c}{\textbf{AIME}}
 & \textbf{Avg} \\
\cmidrule(lr){2-3}\cmidrule(lr){4-5}\cmidrule(lr){6-7}\cmidrule(lr){8-9}\cmidrule(lr){10-10}
\headrow
 & \textit{score} & \textit{$\Delta$} & \textit{score} & \textit{$\Delta$}
 & \textit{score} & \textit{$\Delta$} & \textit{score} & \textit{$\Delta$}
 & \textit{$\Delta$} \\
\midrule
\multicolumn{10}{l}{\textit{Frozen reference}}\\
\cmidrule{1-10}
No improvement & 23.6 & -- & 10.3 & -- & 71.2 & -- & 55.0 & -- & -- \\
\midrule
\multicolumn{10}{l}{\textit{Single operator, full budget}}\\
\cmidrule{1-10}
\dataop    & 27.4 & $+3.8$ & 12.4 & $+2.1$ & 73.8 & $+2.6$ & 57.7 & $+2.7$ & $+2.8$ \\
\bandrow
\harnessop & 29.4 & $+5.8$ & 14.9 & $+4.6$ & 78.8 & $+7.6$ & 63.3 & $+8.3$ & $+6.6$ \\
\modelop   & 27.0 & $+3.4$ & 14.0 & $+3.7$ & 75.4 & $+4.2$ & 59.3 & $+4.3$ & $+3.9$ \\
\midrule
\multicolumn{10}{l}{\textit{Composition baselines}}\\
\cmidrule{1-10}
\bandrow
Human fixed $\opD\!\to\!\opM\!\to\!\opH$ & 30.3 & $+6.7$ & 15.3 & $+5.0$ & 79.4 & $+8.2$ & 64.3 & $+9.3$ & $+7.3$ \\
Random admissible composition & 28.1 & $+4.5$ & 13.6 & $+3.3$ & 75.8 & $+4.6$ & 60.7 & $+5.7$ & $+4.5$ \\
\bandrow
Static router & 30.8 & $+7.2$ & 14.9 & $+4.6$ & 79.6 & $+8.4$ & 64.0 & $+9.0$ & $+7.3$ \\
\midrule
\hlrow
\textbf{\ours (ours)} & \textbf{31.9} & $\mathbf{+8.3}$ & \textbf{19.5} & $\mathbf{+9.2}$ & \textbf{83.8} & $\mathbf{+12.6}$ & \textbf{68.3} & $\mathbf{+13.3}$ & $\mathbf{+10.9}$ \\
\bottomrule
\end{tabular}}
\caption{\textbf{Main results on the sealed splits of four benchmarks.} The GPQA column is GPQA-D-hard100, $100$ pre-frozen harder items scored pass@1, every number is the mean over five outer seeds, and the item index is given in \Cref{tab:hard100}. Target is Qwen3.5-35B-A3B, every condition under the same budget. \ours is best on all four: \textbf{$\mathbf{+10.9}$ average points over the frozen system, and $\mathbf{+3.6}$ over the strongest fixed-order composition}. Best in bold; $\Delta$ is the gain over the frozen system.}\label{tab:main}
\end{table}

\begin{figure}[t]
\centering
\includegraphics[width=\linewidth]{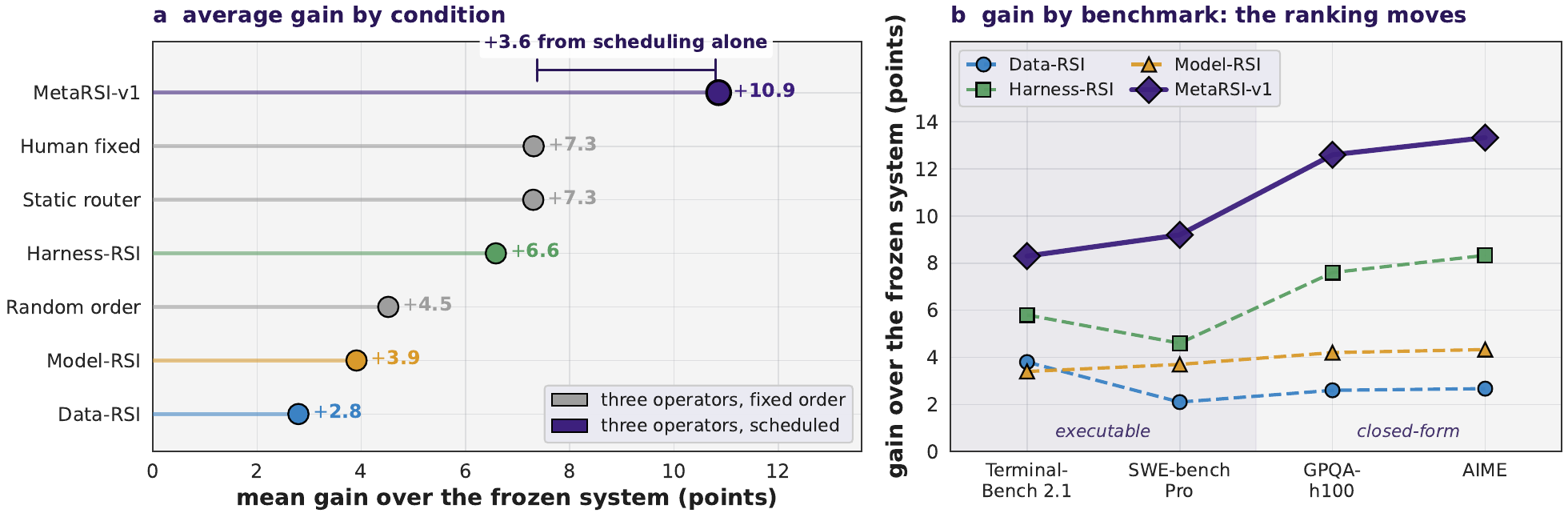}
\caption{\textbf{Where the gain comes from.} \emph{(a)}: mean gain by condition. \textbf{Scheduling is worth $\mathbf{3.6}$ points over the strongest fixed order}, which both fixed-order compositions reach at the same $+7.3$ while the three single operators span $3.8$ points. \emph{(b)}: the same gains by benchmark. \textbf{The operator ranking changes across domains}, \hhl{\harnessop} leading on the closed-form suites and the three converging on the executable ones, which is what makes them complementary rather than redundant; the full system clears the best single operator everywhere by $2.5$ to $5.0$ points. GPQA-h100 is GPQA-D-hard100, the subset of \Cref{sec:exp:setup}, indexed in \Cref{tab:hard100}.}
\label{fig:composition}
\end{figure}

\Cref{fig:composition}(a) plots the average gain of each condition. The three single operators span a $3.8$-point range; the two strongest composition baselines, which use all three operators, land level with each other at $+7.3$, yet the full system sits $3.6$ points above them. The composition gap is therefore not attributable to having more operators in the loop. It is attributable to who decides the order.

\Cref{fig:composition}(b) decomposes the gain by benchmark. The operator ranking is not stable across domains: \harnessop leads on the closed-form suites while the three operators are closer together on Terminal-Bench~2.1 and SWE-bench Pro, and no single operator dominates everywhere. The full system exceeds the best single operator on every benchmark, by $2.5$ to $5.0$ points, confirming that the operators carry complementary signals rather than redundant ones. Averaged by benchmark type, \ours adds $+8.8$ on the executable suites and $+13.0$ on the closed-form suites, so the composition effect generalizes across verification formats.

\subsection{Frontier Models Under the Harness Route}
\label{sec:exp:frontier}

\begin{figure}[!b]
\centering
\includegraphics[width=0.92\linewidth]{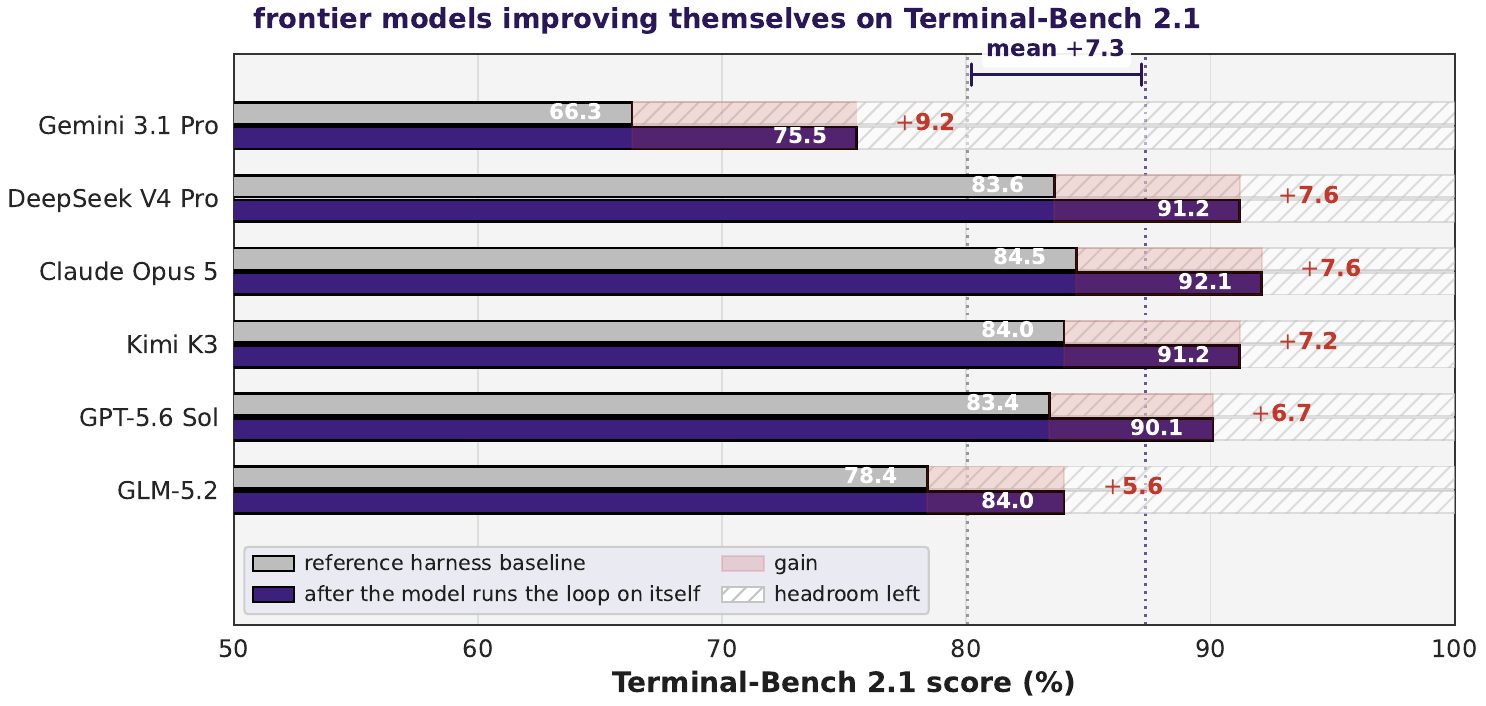}
\caption{\textbf{Frontier models improve themselves on Terminal-Bench 2.1.} Each pair is one model under the reference execution harness (grey) and the same model after running the \ours loop on itself (\textcolor{accent}{purple}), gain at the right. \textbf{All six improve, by $\mathbf{+5.6}$ to $\mathbf{+9.2}$ points, mean $\mathbf{+7.3}$}, each model acting as its own proposer and judge.}
\label{fig:frontier}
\end{figure}

Everything above improves a model whose weights are ours to write. The harness route claims something the main results cannot show: because \hhl{\harnessop} edits only the execution scaffold, the loop runs on a model that can only be reached through an interface, which is the situation for every frontier model. \Cref{fig:frontier} puts the strongest models currently available into the loop and measures what they gain from improving \emph{themselves} through it.

The setup is the same closed loop as everywhere else in this paper, and it is worth being explicit about what that means here. Each frontier model is its own proposer, its own synthesizer, and its own judge; \ours is not carried over from the deployment model and re-applied, and no other model is in the loop. \modelop is unavailable by construction, so the admissible set collapses to $\{\opD,\opH\}$ with the two transitions $\opD\!\to\!\opH$ and $\opH\!\to\!\opD$, and the \rsiag schedules within it. This is also the regime a practitioner without training infrastructure would run.

The frontier comparison is presented on Terminal-Bench~2.1, the executable benchmark on which the reference execution harness yields a hidden-verified run for all six models under one implementation. AIME is saturated for frontier models and offers no headroom; SWE-bench Pro resolve rates are highly scaffold-sensitive, so a clean cross-model comparison is not available there; and GPQA-D-hard100 runs in the degenerate prompt-only configuration, which does not exercise the agent route and adds little to this comparison.

The interesting question is whether the gain survives the strength of the target. A scaffold edit that helps a mid-sized open-weight model may be repairing a deficiency that a frontier model does not have, in which case these bars go flat and the harness route is a small-model technique. \Cref{fig:frontier} answers this on Terminal-Bench~2.1 under the reference execution harness: all six frontier models improve themselves, with gains from $+5.6$ (GLM-5.2) to $+9.2$ (Gemini 3.1 Pro) and a mean of $+7.3$ points. The baselines are already high, $83.4$ for GPT-5.6 Sol and $84.5$ for Claude Opus 5, so the gains are measured against a strong scaffold rather than a weak one. The mean gain on frontier models, $+7.3$, is comparable to the gain on the $35$B target on the same benchmark, $+8.3$, which indicates that the harness route is not repairing a deficiency specific to small models: frontier models also leave material gains on the table that a self-directed loop can recover without any weight update and without an external teacher.

\subsection{Improving the Improver}
\label{sec:exp:terms}

\Cref{tab:main} and \Cref{fig:frontier} both measure one term. The claim that makes the meta layer worth a
separate budget is a different one: that the released successor is a better
\emph{improver}, not only a better system. Term~1 runs one improver over $\Sys_0$
and releases $\Sys_1$. Term~2 forks that release. One branch improves $\Sys_1$
again with the original improver; the other improves the same $\Sys_1$ with the
improver the meta layer has since rewritten, under the same budget and against
the same sealed splits, so the only difference between the branches is which
improver ran.

\begin{figure}[!t]
\centering
\includegraphics[width=\linewidth]{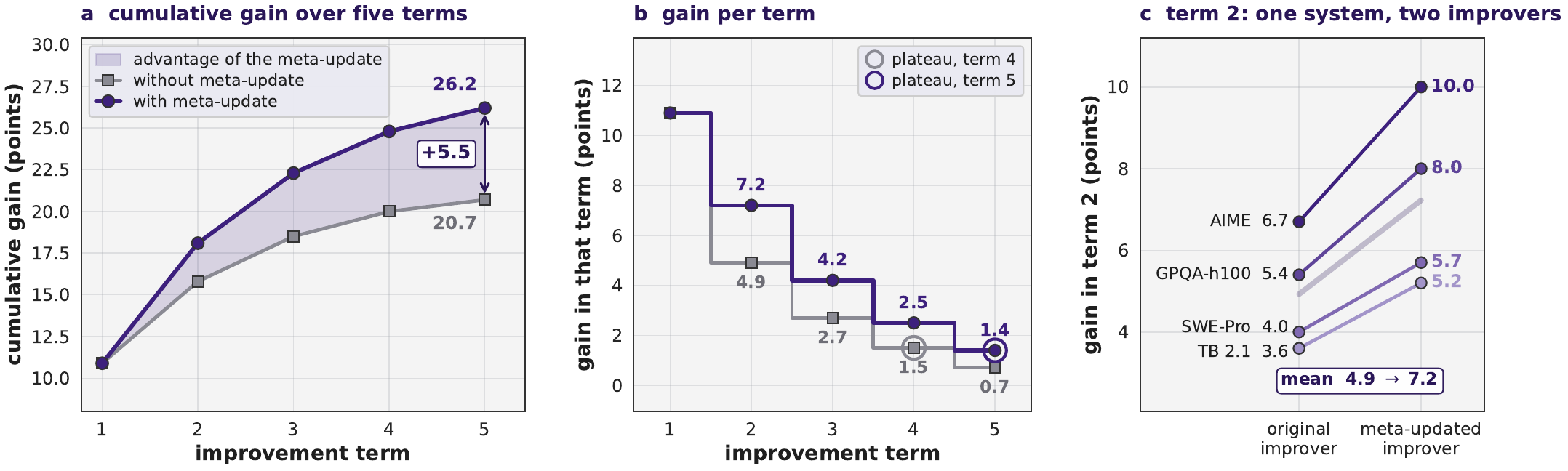}
\caption{\textbf{The successor is a better improver.} \emph{(a)}: cumulative gain
over five terms, with the advantage of the meta-update shaded. \textbf{The meta
path reaches $\mathbf{+26.2}$ against $\mathbf{+20.7}$}, and the gap widens every
term. \emph{(b)}: the same runs as gain per term. Both paths decay; the ringed
term is where each stops returning more than a point and a half, \textbf{term~4
without the meta-update and term~5 with it}. \emph{(c)}: term~2 on one system
under one budget, improved by the original improver and by the meta-updated one.
\textbf{The meta-updated improver adds $\mathbf{7.2}$ points against
$\mathbf{4.9}$}, and its advantage grows with the gain the benchmark admits.}
\label{fig:terms}
\end{figure}

\Cref{fig:terms}(c) gives the answer. The meta-updated improver adds $7.2$
average points where the original adds $4.9$: an advantage of $2.3$ points on one
system under one budget. The advantage scales with the gain a domain admits,
$+1.6$ on Terminal-Bench~2.1, $+1.7$ on SWE-bench Pro, $+2.6$ on GPQA-D-hard100 and
$+3.3$ on AIME, and the ranking of the four is the same under both improvers. What
the meta-update changed is the size of the step, not where the improver looks.

Five consecutive terms extend this. \Cref{fig:terms}(a) plots cumulative gain on
both paths: the meta path reaches $+26.2$ average points against $+20.7$, and the
gap between them widens every term to $+5.5$ at term~5. \Cref{fig:terms}(b) plots
the same runs as gain per term, which is where the shape of self-improvement is
visible. Both paths decay. Without the meta-update, per-term gain falls from
$+4.9$ at term~2 to $+0.7$ at term~5; with it, from $+7.2$ to $+1.4$, so the meta
path is still paying twice as much at term~5 as the other path is. Averaged over
terms~2 to~5 that is $+3.8$ a term against $+2.5$. The plateau, the term at which
a path stops returning more than a point and a half, arrives at term~4 without the
meta-update and at term~5 with it.

\begin{figure}[!t]
\centering
\begin{minipage}[t]{0.487\linewidth}
\vspace{0pt}
\centering
\captionsetup{type=table}
\resizebox{\linewidth}{!}{%
\small
%
%
\renewcommand{\arraystretch}{1}
\setlength{\tabcolsep}{6pt}
\setlength{\aboverulesep}{0pt}
\setlength{\belowrulesep}{0pt}
\begin{tabular}{>{\rule[-10.22pt]{0pt}{26.66pt}}l cc}
\toprule
\headrow
\textbf{Over five terms} & \raisebox{-4pt}[0pt][0pt]{\shortstack{\textbf{without}\\\textbf{meta-update}}} & \raisebox{-4pt}[0pt][0pt]{\shortstack{\textbf{with}\\\textbf{meta-update}}} \\
\midrule
Gain per term, terms~2 to~5 & $+2.5$ & $\mathbf{+3.8}$ \\
\bandrow
Terms to reach $+20$ average & $4.0$ & $\mathbf{2.5}$ \\
Plateau onset & term~4 & \textbf{term~5} \\
\addlinespace[8pt]
\bandrow
Policy carried to the next term & $38\%$ & \textbf{68\%} \\
Scheduler regret & $16.5\%$ & \textbf{8.2\%} \\
\addlinespace[8pt]
\bandrow
Regression on held capability & $-0.2$ GPQA-h100 & \textbf{none} \\
\bottomrule
\end{tabular}}
\vspace{2.12pt}
\caption{\textbf{What five terms show and one term cannot.} Same target, same
per-term budget, the two paths of \Cref{fig:terms}. \textbf{The meta path gains
half again as much per term and reaches $\mathbf{+20}$ in $\mathbf{1.5}$ fewer
terms}. \emph{Policy carried} is the share of accepted operator-policy edits
still in the released improver one term later; \emph{scheduler regret} is the
shortfall of the realized sequence against the best admissible sequence in
hindsight. The regression is the term \Cref{eq:objective}
penalizes.}\label{tab:terms}
\end{minipage}\hfill
\begin{minipage}[t]{0.487\linewidth}
\vspace{0pt}
\centering
\includegraphics[width=\linewidth]{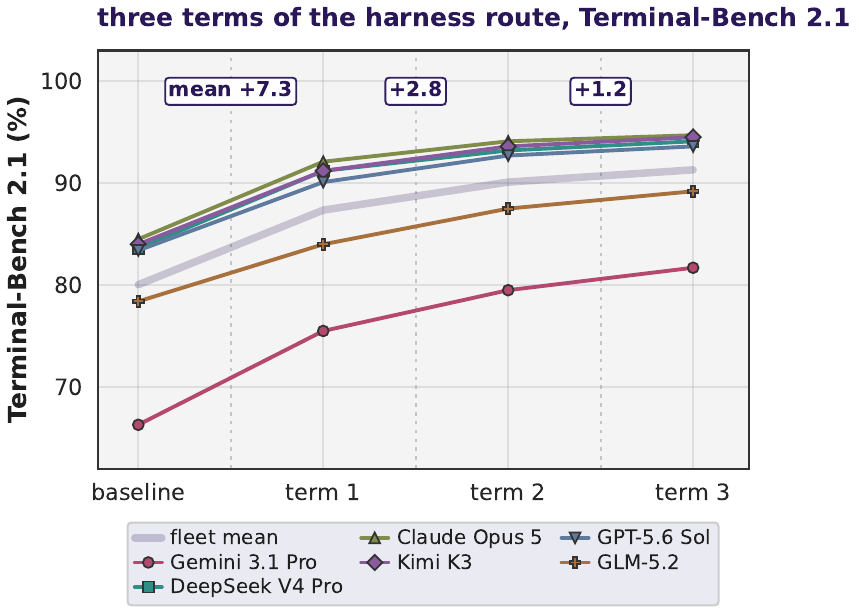}
\caption{\textbf{The harness route compounds on frontier models.} Three terms of
the \dhl{\dataop}, \hhl{\harnessop} loop under the reference execution harness,
each model its own proposer and judge, and every gain the model's own.
\textbf{All six gain in every term}, by $+7.3$, $+2.8$ and $+1.2$ on mean, for
$+11.3$ cumulative. \textbf{Retention follows headroom} rather than strength:
the two lowest baselines keep $43\%$ and $63\%$ of their term-1 gain in
term~2, the four above them $26\%$ to $39\%$.}
\label{fig:frontierterms}
\end{minipage}
\end{figure}

\Cref{tab:terms} reports what five terms show that a single term cannot.
Compounding is the headline, and the two loop-quality measurements underneath it
are the mechanism: the meta path carries $68\%$ of its operator-policy edits into
the following term against $38\%$, and its scheduler leaves $8.2\%$ of the
attainable gain on the table against $16.5\%$. It also holds previously acquired
capability, where the other path gives back $0.2$ points on GPQA-D-hard100 at
term~5, which is the regression term of \Cref{eq:objective} registering. Read
together, the meta-update buys a later plateau and a higher ceiling. Decay itself
is what a loop drawing on a fixed substrate does, and \Cref{law:scarcity} is where
that belongs.

The frontier fleet compounds the same way. \Cref{fig:frontierterms} runs the $\{\opD,\opH\}$ loop for two further terms on each of the six models of \Cref{fig:frontier}, one improver per model and every seat still on the model itself. All six keep gaining: the mean is $+7.3$ in term~1, $+2.8$ in term~2 and $+1.2$ in term~3, for $+11.3$ cumulative, and the fleet mean rises from $80.0$ to $91.3$, with the top four finishing inside $1.1$ points of each other between $93.6$ and $94.7$. What sets the rate is headroom rather than model strength: the two lowest baselines, Gemini 3.1 Pro at $66.3$ and GLM-5.2 at $78.4$, keep $43\%$ and $63\%$ of their term-1 gain in term~2, while the four above them keep $26\%$ to $39\%$, and the two quantities rank together at Spearman $\rho=-0.84$. The harness route therefore compounds on frontier models too, at a rate set by how much of the scaffold is still worth editing.

%
%

%% file: sections/06_discussion.tex
\section{Discussion}
\label{sec:discussion}

\Cref{sec:intro} argued that the object self-improvement should act on is the human labour that turns compute into a deployed model, and that current systems automate the cheapest slice of it. This section returns to that claim and asks what the framework actually buys against it: first for one deployment (\Cref{sec:discussion:buys}), then for the closed feedback loop the three operators form and the two routes through it (\Cref{sec:discussion:routes}), then across scientific domains (\Cref{sec:discussion:science}), and finally at the boundary where a loop stops being digital and starts touching apparatus (\Cref{sec:discussion:physical}).

\subsection{What Composition Buys}
\label{sec:discussion:buys}

Three things change when improvement is expressed as a scheduled composition of typed operators rather than as a loop around one editable surface, and each corresponds to one of the three mechanisms identified in \Cref{sec:intro:defect}.

\subhead{Against misattribution} A single-surface operator's diagnosis is constrained by its write access; under the loop kernel it is constrained by the evidence instead, since the failure signature of \Cref{eq:signature} is grounded before any operator sees it and its vocabulary is read by all three. So a missing procedural habit routes to the harness and absent knowledge routes to data and weights, on a shared object no operator can rewrite in its own favour. The learning signature of \Cref{eq:kappa} sharpens this on the data side: only the capability the rollouts never exhibit justifies spending external supervision.

\subhead{Against non-composability} The engineer's answer from \Cref{sec:intro:defect} (add the rule, collect the corrected behaviour, internalize it, delete the rule) is the path $\opH\!\to\!\opD\!\to\!\opM\!\to\!\opH$ of \Cref{fig:transitions}, every arrow a typed adapter. What the framework adds is that the path is \emph{sayable}: a scheduler can propose it, a type-check can accept it, and its final step can be replayed and reverted if the deletion turns out to cost accuracy.

\subhead{Against measurement in the cheapest regime} The framework requires a verifier of \emph{some} fidelity, plus the three separations of \Cref{sec:method:governance}, which are what let a loop be trusted when the verifier is weak: they keep the loop from improving its own definition of success. A domain with a partial verifier is therefore a harder instance of the same problem rather than a different one, and \dataop supplies what such a domain additionally needs, a bounded statement of where external supervision is actually required (\Cref{sec:discussion:science}).

\subsection{An Amplifier Is Not a Source}
\label{sec:discussion:open}

The composition of the three operators has a fixed upper bound. \dhl{\dataop} renders explicit the competence latent in the model's rollouts; \mhl{\modelop} fixes that competence in parameters; \hhl{\harnessop} makes it available at inference without training. Each redistributes ability the model already holds. A record synthesized for a capability the model has never exhibited is authored by the same model that lacks it (\Cref{sec:method:data}), so the loop cannot bootstrap knowledge it does not contain. The system composed of these operators alone is a \emph{closed amplifier}, whose ceiling is the best arrangement of what was already present.

Every genuine gain requires information originating outside the loop. The framework admits such information through the learning signal, which is defined by its position in the loop and imposes no restriction on its source (\Cref{sec:prelim}). A verifier's ruling, a retrieved document, a curated corpus, an instrument reading, and a commissioned expert note enter on equal terms. Human supervision occupies the same footing: the labour that self-improvement is said to displace is, in this accounting, one information source distinguished primarily by its cost.

\dataop governs the expenditure of that cost. Its learning signatures locate the boundary between competence the model holds and competence the rollouts never exhibit, converting that boundary into a bounded request for the specific information the loop cannot generate. External supervision is allocated to genuine deficits, and the volume of that allocation is reported as a measured quantity. An amplifier with such an intake has an open boundary: it incorporates what it did not contain, and it states how much it had to add.

\subsection{Two Routes Through the Loop}
\label{sec:discussion:routes}

The framework is a cycle before it is a set of routes, and the cycle is what makes it compound. \dhl{\dataop} reads the compiled measurement of the deployed system's competence: the competence it holds, misapplies, or lacks. \hhl{\harnessop} and \mhl{\modelop} consume that measurement and change the system. The changed system then re-enters \dataop: the $\opH\!\to\!\opD$ edge of \Cref{fig:transitions} re-scores the signal under the new scaffold, and the $\opM\!\to\!\opD$ edge re-probes the boundary under the new weights, so the next measurement is taken of a system that has moved. Each turn of the loop scales the competence the next turn can find and press on, which is the sense in which the framework improves the machinery rather than the model. \dataop's role as an instrument of measurement matters here because it is the loop's point of re-entry, and the operators that act on what it reads are what move the system.

Within that loop the measurement is consumable in two structurally different ways, and which one a deployer can take is decided less by which is stronger than by what they are permitted to touch. It can be distilled into the execution scaffold, or it can be materialized into training data and internalized into weights. These are two parallel routes, and the rest of this subsection takes each in turn before returning to what their composition adds.

\subsubsection{The Scaffold Route}
\label{sec:discussion:harnessroute}

The harness route is the cycle $\opD\!\leftrightarrow\!\opH$, and it never writes $\theta$. That makes it the \emph{only} available route for a closed-weight model reached through an API, the situation of most deployments today, and the more economical one even for open weights whenever a single training round would consume the improvement budget.

Its unit of improvement is what makes it distinctive. The genome of \Cref{eq:genome} is small, declarative and inspectable, so a configuration specialized to one narrow task family can be diffed, reviewed, versioned, rolled back and handed to someone else at negligible cost. A checkpoint is a large opaque artifact whose provenance is hard to audit and whose licence usually restricts redistribution; a genome is a short structured document that can be published, criticized and improved in the open, which is the reasoning behind releasing \rsihar as a standalone component (\Cref{sec:discussion:harness}). \Cref{tab:routes} sets out what the route costs; the one cost with a structural consequence is that capability living only in the scaffold is lost the moment the scaffold is not loaded, which is where third-party integration bites.

\subsubsection{The Model Route}
\label{sec:discussion:modelroute}

The model route is the path $\opD\!\to\!\opM$, and it does the opposite: internalized capability is free at inference, survives outside any particular scaffold, and composes with the model's other competences rather than sitting beside them in a prompt. Where a domain will be served at scale and for a long time, this is the route whose economics improve with use.

Three of its costs shaped the framework's design. Training rounds are slow relative to the loop that produced their data, which is why the \rsiag interleaves cheap operators while an expensive one is pending rather than blocking on it. Gains are hard to attribute to a specific edit, which is why every candidate trains from the same fixed base over the cumulative dataset. And every update risks capability the model already had, which is why the cumulative dataset reserves a fraction for previously mastered capabilities and why regression is a first-class penalty in \Cref{eq:objective} rather than a diagnostic reported afterwards.

\subsubsection{Composing the Routes}
\label{sec:discussion:routechoice}

The routes are complementary; neither dominates the other. \Cref{tab:routes} states the conditions under which each route is the right instrument, and the point of the framework is that when both are available, the deployer should not have to choose once and in advance.

\begin{table}[t]
\centering
\small
\renewcommand{\arraystretch}{1.26}
\setlength{\tabcolsep}{6pt}
\resizebox{\linewidth}{!}{%
\begin{tabular}{l l l l}
\toprule
\headrow
 & \hcell{Harness route ($\opD\!\leftrightarrow\!\opH$)} & \mcell{Model route ($\opD\!\to\!\opM$)} & \textbf{Composed} \\
\midrule
Requires        & API access only                & open weights $+$ trainer        & both \\
\bandrow
Cost shape      & recurring, per inference       & one-time, per generation        & one-time, then amortized \\
Unit of change  & a five-slot genome             & a checkpoint                    & genome $\to$ weights $\to$ pruned genome \\
\bandrow
Transferable    & yes, as a document             & only as a large artifact        & the genome carries the recipe \\
Degrades by     & context inflation, rule conflict & forgetting, attribution loss  & pruning bounds the first \\
\bandrow
Fails when      & scaffold not loaded            & data predates the change        & Principle~\ref{prin:admissibility} rejects it \\
\midrule
\hlrow
Best when       & closed weights, many domains, low volume & one domain, high volume, long horizon & capability must persist \emph{and} stay cheap \\
\bottomrule
\end{tabular}}
\caption{\textbf{Route selection.} The two routes out of \dataop differ in what they require, what they cost and how they fail. The right-hand column is what composing them adds, and every entry in it is an edge of \Cref{fig:transitions}.}\label{tab:routes}
\end{table}

The productive interaction is the $\opM\!\to\!\opH$ edge. After internalization, the scaffold entries that carried the now-internalized behaviour are redundant, and Redundancy Reconciliation proposes deleting them under replay. The two routes together therefore \emph{return} context budget that either alone would spend: the harness buys the behaviour cheaply and immediately, the weights absorb it, and the scaffold gives the space back. Viewed in this direction, the same edge prevents the scaffold from growing without bound: internalized entries are pruned, so the harness route does not become a one-way ratchet. This is the concrete sense in which composing operators differs from running them side by side: only the composed loop returns the context budget that internalization spends.

\subsection{Loops Across Domains}
\label{sec:discussion:science}

The reason to build a framework rather than a better coding agent is that the framework's units are \emph{pipeline stages}, not tasks. A domain that wishes to close an improvement loop does not need to reproduce our system; it needs to supply three things that the framework treats as inputs.

\begin{glancebox}
\gitem{OpDataC}{\suitA}{A task family}whose instances can be executed and observed: not necessarily scored perfectly, only observed.
\gitem{OpHarnC}{\suitB}{A verifier of whatever fidelity exists}compilation, simulation, numerical convergence, protocol execution, replication of a reported result, or in the weakest case a rubric applied by a held-out evaluator.
\gitem{OpModC}{\suitC}{A seed scaffold}an initial assignment to the five genome slots, which may be close to empty.
\end{glancebox}

\noindent Everything else (the learning signature of \Cref{eq:kappa}, the failure-signature vocabulary, the admissibility rule, the two scheduling axes, the protected separations) is domain-independent, because none of it inspects the content of a task. This is the practical content of the claim that the framework generalizes: instantiating it somewhere new is a matter of supplying three inputs rather than designing a new improvement loop.

\subsubsection{What Changes Is the Verifier}
\label{sec:discussion:verifier}

The analysis of \Cref{sec:related:boundary} says exactly what the variation in verifier quality implies, and it is worth setting out as a ladder rather than a binary. \Cref{tab:verifier} lists the rungs we can identify, what each admits, and, in the operationally important column, what \dataop is for at that rung.

\begin{table}[t]
\centering
\small
\renewcommand{\arraystretch}{1.24}
\setlength{\tabcolsep}{6pt}
\resizebox{\linewidth}{!}{%
\begin{tabular}{l l l l}
\toprule
\headrow
\textbf{Rung} & \textbf{Verifier} & \textbf{Example domains} & \textbf{What \dataop is for} \\
\midrule
1 & executable tests, exact match & software, terminal, closed-form science & directing synthesis at the diagnosis \\
\bandrow
2 & numerical convergence, simulation & fluid and structural design, circuits & the same, at simulator cost \\
3 & reproduction of a reported result & ML replication, computational biology & separating method gaps from data gaps \\
\bandrow
4 & protocol execution with an instrument & wet-lab chemistry, materials synthesis & \textbf{deciding which experiments to run} \\
5 & expert rubric, no ground truth & law, policy, historiography & \textbf{bounding the request for expert time} \\
\bottomrule
\end{tabular}}
\caption{\textbf{The verifier ladder.} The mechanics are the same at every rung; what the rung decides is which operators are worth scheduling and what \dataop's boundary statement is used for. At the top rungs \dataop mainly directs synthesis; at the bottom rungs its more valuable output is a bounded statement of where human or corpus supervision is actually required.}\label{tab:verifier}
\end{table}

At rungs~1--2 the full framework applies unchanged, and the domain is limited by how much capability the model already has rather than by whether the loop can be closed, which is precisely the situation of the unexploited region in \Cref{fig:landscape}. At rungs~4--5 the framework's most valuable output is no longer the synthesized data but the boundary it reports, because that converts an unbounded request for expert supervision into a bounded one: only the residue the rollouts never exhibit needs a human, an instrument or an external corpus, and the size of that residue is measured rather than assumed. A laboratory that can afford fifty experiments a month cares far more about \emph{which} fifty than about a system that proposes five thousand.

\subsubsection{Decomposing a Scientific Production Line}
\label{sec:discussion:decompose}

The claim that the units are pipeline stages has a concrete reading. A scientific or engineering programme is itself a production line, and its stages are the places where a first-order operator can be dropped in without redesigning anything upstream or downstream. In a materials programme those stages are candidate generation, property prediction, synthesis-route planning, characterization and interpretation; in a computational-biology programme they are hypothesis framing, assay design, pipeline construction, statistical analysis and writing; in a hardware programme they are specification, architectural exploration, implementation, verification and physical closure. Each of these stages has the shape the loop kernel expects: it consumes a typed input, it can be executed and observed, its failures have identifiable terminal causes, and it produces an artifact the next stage consumes.

What follows is that a domain does not have to close its whole loop at once. Instantiating the framework at a single stage yields a first-order loop (one operator, one verifier, one diagnostic artifact), which is already useful, and which produces exactly the typed artifacts a second stage would need to be scheduled against it later. The composition machinery is what turns a collection of stage-local loops into a pipeline-level one, and the admissibility condition is what keeps that composition well posed: a stage cannot consume a description of a system that a previous stage has already changed.

\subsubsection{Why This Compounds}
\label{sec:discussion:compounds}

A term of \Cref{alg:term} produces one released successor and a handful of typed artifacts, and the artifacts are the cumulative part. A learning-signature report states, in a domain-independent format, where a class of models' competence ends; a genome is a small declarative object that made one narrow task family work; a verified record carries the failure signature it came from. All three are typed and carry lineage, so a loop closed in one domain leaves material a loop in an adjacent domain can read: the same failure vocabulary, the same artifact schemas, sometimes the same scaffold entries.

A second effect compounds inside a single loop rather than across loops, and it is the framework's own answer to why any of this accelerates: each of the loop's costs is paid against a state the loop itself improves. Probing a capability is cheaper once a learning-signature report for a neighbouring subdomain exists, since the probe set can be seeded rather than constructed. An adapter written once is reused by every later term that schedules that edge. A scaffold already holding a domain's procedural habits makes the next round fail in more informative ways, the remaining failures being the ones previously masked. Cycle time is therefore not a constant of the domain but a quantity the framework is also optimizing, which is what the vertical axis and the meta layer are for, and why cost-to-threshold rather than end-state accuracy alone is the quantity worth reporting as the framework matures.

Whether many such loops, exchanging artifacts across many disciplines, aggregate into something resembling broad expert-level competence is an open empirical question. What this report establishes is the mechanism such an aggregate would run on, buildable from parts we have built.

\begin{takeaway}
\textbf{The macro claim, stated precisely.} For settings whose situations are enumerable, we claim four things, with the extension to open worlds left to the projection of \Cref{sec:discussion:physical}: that the labour of building a model is decomposable into three operators whose composition is analyzable; that the resulting loop can be trusted under a weak verifier because what defines success is outside every write surface; that instantiating the loop at a new pipeline stage costs three declared inputs rather than a new design; and that the artifacts it leaves behind are typed, auditable and reusable by the next loop. If those four hold, the region of \Cref{fig:landscape} where capability already exists and no loop has been closed is addressable work rather than a research frontier.
\end{takeaway}

\subsection{Toward Loops That Touch the Physical World}
\label{sec:discussion:physical}

Every loop in this report closes inside a computer: the rollouts are processes, the verifier is a test suite or a scorer, and a rejected candidate costs tokens. The domains with the largest standing gap in \Cref{fig:landscape} (materials synthesis, wet-lab protocol execution, robotic manipulation) are not like that: their verifier is an instrument, their rollouts consume physical resources, and a bad action can be unrecoverable. What follows states which parts of the framework carry over, which break, and what would have to be added.

\subhead{Carry-over} The learning signal, the budget ledger and the protected separations are indifferent to whether the verifier is a test runner or a spectrometer, and the ledger simply gains rows for instrument hours, consumables and sample stock. One mechanism gains force rather than merely keeping it. \dataop's boundary statement becomes an \textbf{experiment allocator}: a laboratory's binding constraint is instrument time, and the residue \dataop cannot synthesize for is a measured, bounded statement of which questions actually require the instrument.

\subhead{Failure modes} Two things, both structural. \emph{Reversibility}: the framework rests on candidates being cheap to generate and free to discard, and a physical action has no sandbox, so consuming a sample or damaging a manipulator cannot be rolled back. The admissibility condition of Principle~\ref{prin:admissibility} asks only whether an operator's inputs are fresh, so a physical framework needs a second precondition asking whether the action is \emph{recoverable}, and a scheduler that treats irrecoverable steps as commitments rather than candidates. \emph{Loop latency}: the economics assume the cheap operators are orders of magnitude cheaper than the expensive one, and when the expensive one takes days and consumes material, the ratio widens far enough that the \rsiag's job changes from sequencing to deciding whether to act at all.

\subhead{Required additions} Three additions follow. \emph{Staged fidelity}: simulation, then bench proxy, then instrument, with the same operator set at each stage and promotion between stages gated as candidate promotion is gated now, which makes the irreversible rung the last one reached and lets the cheap rungs shrink the residue before any material is consumed. An \emph{irreversibility-aware action space}, in which an operator declares not only what it writes but whether the write can be undone, and \textsc{Validate} refuses irrecoverable proposals that lack an explicit authorization. And \emph{human authorization as a protected surface}: the point at which a person signs off on an irreversible action sits outside every write surface at every level, for the same reason the sealed evaluator does.

\subhead{The deeper problem: the environment, not the actor} Every dimension of \Cref{eq:kappa} presupposes a well-posed task: each attributes a failure to a competence of the actor against a known environment, the setting of every benchmark in \Cref{sec:exp}. An open physical setting admits a failure with no such attribution, in which the actor's competence is not at fault because the situation itself lies outside the environment model the signature was compiled against. The distinction this forces is internal to the write-surface algebra. A failure the environment model can reproduce stays a candidate for the three existing operators; one it cannot reproduce requires a fourth operator whose write surface is that model~\citep{lecun2022path}. That surface is the verifier \Cref{sec:method:kernel} leaves unfixed, met in its hardest form, and this operator's products are revised dynamics and the regression tests that pin them. The line between an actor deficit and a model deficit therefore falls out of admissibility rather than requiring a new mechanism. Rewriting the environment model invalidates every diagnostic artifact compiled against the old one, so this operator is capability-altering under Principle~\ref{prin:admissibility} and forces downstream re-probing. The reported quantity changes with it: not accuracy on a frozen task set, but the rate at which an unmodelled situation is converted into a verified, reusable one.

\subhead{Where the first physical loop closes} The precondition is a programmable surface on the apparatus, now being built independently of this work. Agent-to-instrument protocols expose laboratory devices through a common sensing-and-actuation interface: a signed card declares an instrument's capabilities and its physical limits, binding is discovery-first, results carry units and calibration, and an irreversible operation is gated behind a cryptographically bound operator confirmation~\citep{zhu2026lap}. A heterogeneous fleet therefore presents a single typed action surface, and that last provision is the protected human authorization of the preceding paragraph, arrived at independently. Deployments already exhibit the \Cref{law:conservation} substrate migration in physical form: an agent operating an X-ray nanoprobe and a materials robot consolidates what it works out online into reusable routines the instrument then runs on its own, under human safety confirmation, with the instrument's own reading serving as verifier~\citep{vriza2026instruments}. Our framework predicts the order of maturation. By \Cref{law:affordance} a loop exists where verification is cheap; on the ladder of \Cref{tab:verifier} an automated laboratory occupies rungs~3 to~4, where the programmable instrument is both actuator and verifier and staged fidelity places the irreversible step last. Open-world embodiment at rung~5, lacking a closed verifier, matures later, so the first physical successors appear in automated laboratories before unconstrained environments.

\subsection{The Reusable Artifact}
\label{sec:discussion:harness}

The genome of \Cref{eq:genome} turns out to be useful independently of the rest of the framework, and we open-source it as a standalone component. \rsihar is the harness operator's runtime and genome format packaged for direct use, in which a task-specific configuration is a stored genome loaded when the corresponding task family is detected. The component ships with the genomes produced by our own runs, so a user starts from configurations that were selected against a fixed evaluator rather than written by hand, and it retains the operator's typed patch format, so a user's own improvement loop can be run over their own tasks. Because a genome is small, declarative and evaluable in isolation, configurations are shareable in the way that model checkpoints are not; \rsihar accepts contributed genomes, so the community around it builds a public library of narrow, well-tested scaffolds rather than a single general one. That library is also the cheapest available test of the cross-domain claim in \Cref{sec:discussion:science}: if genomes contributed for unrelated task families turn out to share scaffold entries, the shared substrate is real; if they share nothing, it is not.

\subsection{Five Laws of Recursive Self-Improvement}
\label{sec:discussion:laws}

Each regime of \Cref{sec:related:fm} left behind a law before it left behind a system: scale left the compute--data--parameter trade as a predictable relation rather than a set of checkpoints~\citep{kaplan2020scaling}, and it is the relation, not any model trained under it, that told the field what to build next. Self-improvement has no such relation yet. We state five, in the form a later result could contradict~\citep{meta2026scientificmethod}, each a claim, a mechanism, and the observation that would refute it.

\begin{lawbox}
\lawhead{Verification, not capability, sets the frontier}\label{law:affordance}
Let $N(d)$ be the loops ever closed in domain $d$, let $R(d)\in\{1,\dots,5\}$ be its practice verification rung from \Cref{tab:verifier}, and let $\mathrm{cap}(d)$ be the competence frontier models already show. Over the twenty-two domains of \Cref{fig:landscape}, and over the seventeen of them that have a capability,
\begin{equation*}
\rho\big(N,R\big)\;=\;-0.75,
\qquad
r\big(N,\mathrm{cap}\big)\;=\;+0.64,
\qquad
r\big(N,R \mid \mathrm{cap}\big)\;=\;-0.67
\;\;\text{but}\;\;
r\big(N,\mathrm{cap} \mid R\big)\;=\;+0.45^{\text{\,n.s.}}.
\end{equation*}
The partial correlations match \Cref{fig:landscape}(c): the rung keeps its effect after controlling for capability, while capability does not remain significant after controlling for the rung. The domains self-improvement can enter are the domains that can be checked; capability decides only whether entering would pay. The census behind these numbers more than doubled in August 2026, to $55$ loops, and the relation held: $53$ of the $55$ close at rungs~1 to~3, and the eight domains that had never had a loop still have none. \emph{Mechanism}: a loop is built where closing it is cheap, and closing it is cheap where verification is free. \emph{Corollary}: to widen self-improvement, build verifiers. \emph{Refuted if} $N$ tracks $\mathrm{cap}$ once $R$ is controlled for.

\smallskip
\lawhead{Self-knowledge expires, so re-description is the rate limit}\label{law:freshness}
An operator reads a description of the system, never the system. Any step that changes what the system does invalidates every description compiled before it. The binding cost of a self-improving architecture is therefore re-describing the system, not changing it. Principle~\ref{prin:admissibility} is this law's local form: over $\{\opD,\opH,\opM\}$ it admits five of the six ordered pairs, and names which. \emph{Mechanism}: evidence about a state that no longer exists supports no inference about the state that replaced it. \emph{Corollary}: a larger operator alphabet needs no new algebra, only the same condition again; an operator that writes the verifier invalidates the whole history at once. \emph{Refuted if} a loop sustains gain on descriptions compiled before its own capability-altering steps.

\smallskip
\lawhead{Competence is substrate-free; its cost is not}\label{law:conservation}
The same behaviour on the scaffold and in the weights is one competence under two cost functionals. One is re-paid at every inference, the other paid once. Competence therefore belongs to the system rather than to a substrate, and a mature loop spends most of its effort moving competence to where it is cheap. \emph{Mechanism}: only a loop with two writable substrates can retire what the other has absorbed. A single-substrate loop accumulates cost monotonically and stops on cost, not on capability. \emph{Corollary}: substrate plurality is necessary, not convenient. The one operation that returns budget is the one a single-surface system cannot express. \emph{Refuted if} a single-substrate loop improves without its per-inference cost growing.

\smallskip
\lawhead{Trust is measured by what cannot be written}\label{law:adjudication}
From inside a loop, better capability and a better definition of success give the same number. This is stronger than Goodhart: the divergence is not merely hard to detect, it is unobservable from inside. No quantity of internal validation finds it, and the remedy cannot be statistical. If $\Qsealed$, its task set, the release rule and the ledger lie outside every write surface at every level, a reported gain is capability. To the extent that any of them is writable, the same number is definition. \emph{Mechanism}: an optimizer allowed to move the target moves it, that being the cheapest action available~\citep{misevolve2026}. \emph{Corollary}: improve the working signal, never the anchor. The anchor is what licenses the claim. \emph{Refuted if} a writable anchor yields gains that survive an unwritable one.

\smallskip
\lawhead{No loop creates capability; every gain is imported}\label{law:scarcity}
A loop redistributes and internalizes competence the system can already show. \Cref{eq:kappa} draws the line between that and what the rollouts never show. Supervision on the first side buys nothing the system's own outputs would not supply. On the second it cannot be synthesized at all, because the model writing the record is the one that lacks the knowledge. A self-improvement result therefore has two numbers, not one: what it amplified, and what it imported. \emph{Mechanism}: a record for a never-shown capability is schema-indistinguishable from a correct one, so it trains like everything else. \emph{Corollary}: the import is measurable, and worth more as the verifier gets dearer, from directing synthesis at rung~1 to allocating instrument time at rung~4. \emph{Refuted if} synthesis past the boundary measurably helps.
\end{lawbox}

\begin{takeaway}
Read together, the five partition what any self-improving loop must answer: where a loop can exist is set by verification (\Cref{law:affordance}); how fast it turns is set by the cost of re-describing a system that keeps changing (\Cref{law:freshness}); what a gain costs is set by which substrate holds the competence (\Cref{law:conservation}); whether the number means anything is set by what stays unwritable (\Cref{law:adjudication}); and what must be bought from outside is set by a boundary that can be measured (\Cref{law:scarcity}). No law names a model, a benchmark, or a parameter count; they govern loops, not systems, and they are the ground on which \Cref{sec:discussion:position} states its position in loops rather than models.
\end{takeaway}
\subsection{Position: The Unit of Progress Is the Loop, Not the Model}
\label{sec:discussion:position}

This subsection argues a position, in the manner such arguments are usually made~\citep{lecun2022path,silver2021reward,zahavy2026jump}: a proposition about where effort should go, supported by the measurements in this report and reaching past them.

\begin{takeaway}
\textbf{The proposition.} For the next stage of progress the binding constraint is not how capable models are but \emph{how many loops exist and where they can be closed}. On that reading the unit of progress is the loop, not the model, and the productive question is not ``how much better can this model get'' but ``how cheaply can a new loop be closed somewhere it has never been closed.''
\end{takeaway}

Two measurements in this report point the same way. Counted by what closes them rather than by subject, better than two thirds of surveyed self-improvement systems verify against a closed, machine-checkable target, so the field's concentration is one of \emph{affordance} rather than of interest. And over twenty-two domains, what decides whether a domain gets a loop is whether it can be checked rather than whether models are good at it, with the standing loss \Cref{law:affordance} identifies concentrated in five domains that are already competent and have no machine-checkable target. That region is capability that exists and is not being harvested.

If the proposition holds, some avenues are far more likely to pay than others, and it is more useful to say which than to be even-handed. \emph{Likely}: instantiating a first-order loop at a single pipeline stage, since one operator, one verifier of whatever fidelity and one seed scaffold already yield something useful and emit the typed artifacts a neighbouring stage can later be scheduled against; the harness route wherever weights cannot be touched, which is most deployments; and the diagnostic boundary used as an allocation device rather than a data generator, which is what it is for once a verifier costs instrument time rather than milliseconds. \emph{Unlikely}: synthesizing training data for genuinely absent capability, for the reason given in \Cref{sec:method:data}: the model writing the record is by construction the one that lacks the knowledge; closing new loops by scaling a fixed verifier, since a fixed verifier is a fixed ceiling and the policy will accumulate exactly in its blind spot; and treating an unmodelled situation as a capability gap, which mistakes a missing situation for a missing skill.

Three findings would refute it. Loop count tracking capability once the verifier rung is controlled for would make the standing loss an artifact of our placements. Diagnostic artifacts and genomes failing in practice to seed the next loop would make the units tasks after all, and the framework a well-organized pipeline. And operators improved in isolation composing no better than they compose here would make the algebra bookkeeping. \Cref{tab:main}'s composition baselines are the smallest experiment bearing on the third; the second is the study we would run next.

The proposition covers settings whose situations are enumerable in advance, which is every setting evaluated here; its extension to open worlds, where the space of situations is itself an operand, is the projection of \Cref{sec:discussion:physical} and \Cref{sec:method:kernel}. The closing claim is about the mechanism: every part of the machinery required to try is buildable from parts already built.

%% file: sections/07_conclusion.tex
\section{Conclusion}
\label{sec:conclusion}

This report set out to move recursive self-improvement from a single editable surface to the production line that turns compute into a deployed model. The field's concentration is a bound on \emph{format} and not on subject: better than two thirds of the systems we surveyed close their loop against a machine-checkable target, which certifies benchmark-bound capability rather than general capability in a discipline.

Against that we introduced \ours, in which improvement is the scheduled composition of typed operators over one unified execution paradigm. Every operator instantiates one \emph{loop kernel}, a cycle closed by a compiled learning signal and cut once into a mutable arc where a model proposes and a protected arc where deterministic code adjudicates. Instantiated on a deployed system's data, scaffold and model, it yields \dhl{\dataop}, which amplifies existing competence and marks its boundary so external supervision is spent only past it; \hhl{\harnessop}, which edits a five-slot scaffold without touching the model; and \mhl{\modelop}, which converts a recurring context cost into a one-time training cost. One artifact vocabulary lets a single freshness condition admit five of the six ordered transitions, one of them the edge on which internalization licenses deleting the scaffold rule it subsumes. Above the operators, one \rsiag chooses at each step between extending the sequence and rewriting an operator's proposal policy, and a meta layer revises that scheduler once a term completes. Under a fixed budget, a sealed measurement and no external teacher, \ours improves the released successor by $3.6$ points over the strongest fixed pipeline.

What we would most want to leave behind is not the system but the five relations of \Cref{sec:discussion:laws}: the reachable frontier of self-improvement is the verification frontier and not the capability frontier; a system's description of itself expires when the system changes, so re-description is the rate limit; competence is invariant across substrates while its cost is not, making improvement relocation as much as acquisition; trust has a measure rather than a degree, and the measure is what the system cannot write; and no loop creates capability, so every genuine addition is imported. Each is stated so a later result can contradict it, and together they read the unit of progress as the loop rather than the model: what binds the next stage is how cheaply a loop can be closed where none has been. The harness runtime and genome format are released as the open-sourced package \rsihar, together with the genomes produced by our runs, to seed a community library of task-specific scaffolds.

%% file: sections/99_appendix.tex
\section{Evaluation-Domain Census}
\label{app:census}

\subhead{Inclusion criteria} A system enters the census only if it satisfies all four conditions: it instantiates a closed loop in which the system reads its own execution evidence and produces a modification; the modification targets the system's own data, harness or weights rather than an external environment or dataset; the modified system becomes the substrate for the next round; and it reports a quantitative result on a named benchmark. A paper that proposes an architecture without a closed-loop benchmark result, or that describes a training-time technique rather than a deployed self-improving agent, is out of scope.

\subhead{Pool construction} The initial pool came from three surveys~\citep{ren2026survey,gao2025selfevolving,misevolve2026} and three curated community indexes of self-improving and self-evolving agents, filtered by the four criteria above, over papers appearing between 2022 and August 2026. The census contains $45$ systems, $31$ closing against a machine-checkable target and $14$ against an open one. Two of them are consecutive work from one group and share a core memory-based framework, \citet{zhou2026mementoskills} building on the read-write mechanism of \citet{zhou2025memento}; they are counted separately because they improve different surfaces, memory against a skill library, and the dependence is noted here rather than hidden in the count.

\Cref{fig:census} and \Cref{fig:landscape} summarize a census over the self-improvement systems discussed in \Cref{sec:related:rsi}. \Cref{tab:census} gives the underlying assignment so that both figures are reproducible. Each system carries two labels. The \textbf{domain} is the subject area in which the majority of its reported benchmarks fall, with ties broken by the benchmark on which the headline result is stated. The \textbf{verification class} is the object that actually closes its loop, and it is the label the argument of \Cref{sec:intro} turns on: \textsc{x} marks a closed, machine-checkable target (an executable test suite, an exact-match or numeric key, a multiple-choice label), and \textsc{o} marks an open target, where the loop is closed by an environment reward or a domain outcome with no key. The census counts \emph{where a loop was closed and measured}, not what a system claims to generalize to. Thirty-one of the forty-five sit in class~\textsc{x}.

\begin{table}[!b]
\centering
\resizebox{\linewidth}{!}{%
\small
\renewcommand{\arraystretch}{1.02}
\setlength{\tabcolsep}{5pt}
\begin{tabular}{l cc l}
\toprule
\headrow
\textbf{System} & \textbf{Dom.} & \textbf{Ver.} & \textbf{The loop was closed and measured on} \\
\midrule
STOP~\citep{zelikman2023stop} & \textsc{C} & \xcl & self-referential code-improvement tasks, scored by execution \\
AFlow~\citep{zhang2024aflow} & \textsc{C} & \xcl & HumanEval, MBPP, GSM8K, MATH, HotpotQA, DROP \\
DGM~\citep{zhang2025dgm} & \textsc{C} & \xcl & SWE-bench, Polyglot \\
Self-Harness~\citep{selfharness2026} & \textsc{C} & \xcl & Terminal-Bench 2.0, SWE-bench Verified, AppWorld \\
Hyperagents~\citep{hyperagents2026} & \textsc{C} & \xcl & coding, paper review, robotics reward design \\
Reflexion~\citep{shinn2023reflexion} & \textsc{C} & \xcl & HumanEval, ALFWorld, HotpotQA \\
RSIBench-Data~\citep{meng2026rsibenchdata} & \textsc{M} & \xcl & six benchmarks over software engineering, terminal use, scientific QA and mathematics; headline gain on AIME 2026 \\
Escher-Loop~\citep{escherloop2026} & \textsc{M} & \xcl & mathematical optimization: Kissing Number, Circle Packing \\
STaR~\citep{zelikman2022star} & \textsc{M} & \xcl & GSM8K, CommonsenseQA \\
DataEnvGym~\citep{khan2025dataenvgym} & \textsc{M} & \xcl & mathematics, code and visual question answering \\
G\"odel Agent~\citep{yin2025godelagent} & \textsc{S} & \xcl & reading comprehension, mathematics and reasoning suites \\
Self-Adapt.~\citep{zweiger2025seal} & \textsc{G} & \xcl & SQuAD-style knowledge incorporation and ARC \\
MetaSkill-Ev.~\citep{metaskill2026} & \textsc{G} & \xcl & OfficeQA, SealQA, ALFWorld \\
ADAS~\citep{hu2024adas} & \textsc{G} & \xcl & ARC, DROP, MGSM, MMLU, GPQA \\
Self-Refine~\citep{madaan2023selfrefine} & \textsc{G} & \xcl & seven tasks including code optimization and constrained generation \\
GEPA~\citep{opsahl2025gepa} & \textsc{G} & \xcl & HotpotQA, IFBench, HoVer, PUPA \\
DSPy~\citep{khattab2024dspy} & \textsc{G} & \xcl & HotpotQA, GSM8K \\
Introspection~\citep{introspection2026} & \textsc{G} & \xcl & instruction-following and question-answering suites \\
Self-Reward~\citep{yuan2024selfrewarding} & \textsc{G} & \ocl & AlpacaEval, adjudicated by a held-out language-model judge \\
DemoEvolve~\citep{demoevolve2026} & \textsc{E} & \ocl & Liar's Dice and Balatro, scored by environment return \\
SEAL~\citep{ant2026seal} & \textsc{E} & \ocl & agentic environments scored by environment return \\
Voyager~\citep{wang2023voyager} & \textsc{E} & \ocl & Minecraft, scored by environment progress \\
SIA~\citep{hexo2026sia} & \textsc{O} & \ocl & LawBench charge classification, TriMul GPU-kernel latency, single-cell RNA denoising \\
AI Scientist~\citep{lu2024aiscientist} & \textsc{O} & \ocl & machine-learning research papers, judged as research output \\
\bandrow
Meta-Harness~\citep{lee2026metaharness} & \textsc{C} & \xcl & TerminalBench-2, online text classification, IMO-level math retrieval \\
VSI~\citep{zhang2026vsi} & \textsc{M} & \xcl & GSM8K over five rounds, with step-level symbolic verification \\
\bandrow
AgentEvolver~\citep{zhai2025agentevolver} & \textsc{E} & \ocl & novel agent environments, scored by environment return \\
Autogenesis~\citep{zhang2026autogenesis} & \textsc{C} & \xcl & GPQA-Diamond and AIME24/25, GAIA, and an in-house LeetCode benchmark under an execution judge \\
\bandrow
Memento~\citep{zhou2025memento} & \textsc{G} & \xcl & GAIA, DeepResearcher, SimpleQA, HLE \\
Memento-Skills~\citep{zhou2026mementoskills} & \textsc{G} & \xcl & GAIA and HLE, under iterative skill evolution \\
\bandrow
MOSS~\citep{cai2026moss} & \textsc{C} & \ocl & OpenClaw, scored by a keypoint rubric with no machine key \\
SAGE~\citep{peng2026sage} & \textsc{C} & \xcl & LiveCodeBench and OlympiadBench, decided by external verifiers \\
\bandrow
AHE~\citep{lin2026ahe} & \textsc{C} & \xcl & Terminal-Bench 2 over ten iterations, and SWE-bench Verified \\
GEA~\citep{weng2026gea} & \textsc{C} & \xcl & SWE-bench Verified and Polyglot \\
\bandrow
Ouroboros~\citep{razzhigaev2026ouroboros} & \textsc{C} & \xcl & Terminal-Bench 2.1, OSWorld-Verified, CL-Bench \\
HSI~\citep{zhou2026hsi} & \textsc{E} & \ocl & BALROG (BabyAI, Crafter, TextWorld, MiniHack) and BabaIsAI, by \% progress \\
\bandrow
EvolveR~\citep{wu2026evolver} & \textsc{G} & \xcl & multi-hop QA (HotpotQA, 2WikiQA, Musique), by exact match \\
FORGE~\citep{bogdanov2026forge} & \textsc{E} & \ocl & CybORG CAGE-2, a network-defence POMDP, by evaluation return \\
\bandrow
EvoTrainer~\citep{chen2026evotrainer} & \textsc{C} & \xcl & mathematical reasoning, competitive programming, repository-level SWE \\
AREX~\citep{lu2026arex} & \textsc{G} & \xcl & BrowseComp, WideSearch, DeepSearchQA, HLE \\
\bandrow
MetaAgent~\citep{qian2025metaagent} & \textsc{G} & \xcl & GAIA, WebWalkerQA, BrowseCamp \\
Q-Evolve~\citep{zhang2026qevolve} & \textsc{E} & \ocl & ALFWorld, WebShop, ScienceWorld, by environment return \\
\bandrow
SkillRise~\citep{yao2026skillrise} & \textsc{E} & \ocl & ALFWorld, WebShop, ScienceWorld, by environment return \\
BPO~\citep{wang2025bpo} & \textsc{E} & \ocl & ALFWorld, ScienceWorld, WebShop, by environment return \\
\bandrow
SkillPyramid~\citep{xiong2026skillpyramid} & \textsc{E} & \ocl & ALFWorld, WebShop, ScienceWorld, by environment return \\

\bottomrule
\end{tabular}}
\caption{\textbf{Evaluation census.} Each surveyed system ($N=45$) carries a subject label, a verification class, and the benchmark the classification was read off. \emph{Subjects.} \textsc{C}: code and terminal; \textsc{M}: mathematics; \textsc{S}: closed-form science QA; \textsc{G}: general QA and instruction following; \textsc{E}: embodied or game environments; \textsc{O}: other scientific and professional domains. \textbf{\textcolor{OpDataC}{x}} is a closed, machine-checkable target and \textbf{\textcolor{MetaC}{o}} an open one, by the classification stated above. Every aggregate in \Cref{fig:census} is derived from this table. No \textsc{S} or \textsc{M} system is class~\textbf{\textcolor{MetaC}{o}}: scientific subject matter does not imply an open verification target.}\label{tab:census}
\end{table}

\subhead{Verification-format classification} Each system is assigned by the primary benchmark on which its improvement loop closes. \emph{Executable tests}: a test suite, a terminal state check or code execution decides correctness ($14$). \emph{Exact-match or choice keys}: a string comparison against a gold answer, or a choice label ($16$). \emph{Numeric or objective keys}: a computed scalar objective ($1$). Those three are closed. \emph{Environment reward}: a scalar returned by a game or simulator ($10$). \emph{Learned judge}: a held-out model adjudicates, with no key ($1$). \emph{Domain outcome}: an expert rubric or domain review, with no key ($3$). Those three are open. A system validating on several formats takes the format of its primary closed-loop benchmark, and its row names the others.

\subsection{The Twenty-Two-Domain Audit}
\label{app:domains}

\Cref{tab:domains} is the per-domain half of the census, and it is what makes the
two correlations of \Cref{fig:landscape} recomputable: the loop counts of panel~(a),
the rung of panel~(b), the capability axis of panel~(c) and both coordinates of
panel~(d) are the three numeric columns below and nothing else.

\subhead{Two domain sets, and which panels use which} The audit covers $22$
domains, and five of them carry no capability value. Panels~(a) and~(b) use all
$22$: one needs a loop count, the other a loop count and a rung, and every domain
has both. Panels~(c) and~(d) use the \textbf{$17$ subject domains}, because a
capability axis needs a subject-matter accuracy and four of the remaining five are
testbed families rather than subjects. What those four report is an agentic task
success rate, GAIA Pass@1 or an ALFWorld return, which is not the same quantity as
an MMLU-Pro or GPQA-Diamond accuracy. The same standard excludes a head-to-head
win rate against human experts, which is why GDPval does not appear on this axis
either. The fifth is classical languages, and the same rule puts it off the axis:
the only score published for it is an HLE split, HLE is this report's practice end,
and \Cref{fig:landscape}(e) measures $35$ to $83$ points between a practice score
and a recall one, so the two cannot share a column. It keeps its rung and its zero
loop count, which is what the standing-loss argument reads it for. Every count below
states which set it is over, and no panel drops a domain silently.

\subhead{The four testbed families} The taxonomy of subject domains was assembled
from benchmarks available through $2024$, and ten of the twenty-one systems added
to the census in August 2026 close on testbeds it had no row for. Four domains were
added to hold them, and all four verify at rung~1 or~2, which is a result rather than a
convenience: these are the testbeds the 2025 and 2026 systems chose, and they chose
cheaply checkable ones. \textbf{General assistant} is GAIA and its relatives, scored
by exact match against a short reference answer, rung~1, $4$ loops. \textbf{Embodied
and interactive environments} covers ALFWorld, WebShop, ScienceWorld, BALROG and
CybORG CAGE-2, each returning a scalar from a simulator, rung~2, $10$ loops.
\textbf{Multi-hop QA} is HotpotQA, 2WikiQA and Musique under exact match, rung~1,
$1$ loop. \textbf{Deep research} is BrowseComp, WideSearch and DeepSearchQA, also
exact match, rung~1, $1$ loop; the system there additionally reports HLE, which is a
rubric, so that one domain is a mixed case, recorded as such.

\subhead{Verifier-rung assignment} \textbf{Rung} is the x-axis of
\Cref{fig:landscape}(b) and~(d), and it is the ladder this report already defines
in \Cref{tab:verifier}: rung~1 executable tests and exact match, rung~2 numerical
convergence and simulation, rung~3 reproduction of a reported result, rung~4
protocol execution with an instrument, rung~5 an expert rubric with no ground
truth. Using the report's own ladder rather than a new scale means the axis is
defined in the body, cited, and consistent with \Cref{sec:discussion:verifier}.

Each domain takes the rung of its \emph{practice} benchmark, never of its recall
benchmark and never of its loop count. Rung~1 is competitive programming,
software engineering, graduate science QA, olympiad mathematics, terminal
operations, research-level mathematics, general assistant, multi-hop QA and deep
research, each closing against a test suite or an exact key. Rung~2 is engineering
design, whose practice is structural and circuit simulation, and embodied and
interactive environments, which return a scalar from a simulator; engineering
design's MMLU-Pro score is multiple choice and would be rung~1, but that is recall.
Rung~3 is ML research replication and single-cell genomics, which reproduce a
reported metric. Rung~4 is robotic manipulation, wet-lab protocol and materials
science, all of which need an instrument; DFT stability is computable, but the
practice of materials science is synthesis. Rung~5 is law, history, health,
philosophy, economics and classical languages, whose practice is scored by expert
or rubric judgement.

\subhead{The rung belongs to the domain, not to the loop} One system in the
census, MOSS~\citep{cai2026moss}, closes its loop in software engineering
against a keypoint grader rubric with no machine key. That is a fact about the
target MOSS chose, and \Cref{fig:census} is where it appears: MOSS is the single
code system in the open block, and that caption names it. It does not make software
engineering a rung-5 domain, because the rung is read off the domain's practice
benchmark, SWE-bench, which is an executable test suite. Counting a loop's own
verifier against the domain rungs would mix two different quantities, so this report
does not: there is exactly one loop in a rung-5 domain, in law.

\subhead{Why the rung, and not an affordance percentage or a binary class} Two
alternative axes are available for this relation and neither is sound. A
continuous \emph{verification affordance}, the share of a domain's work admitting
a machine-checkable target, has no published measurement for any of these domains; it would only be an expert estimate, so it is not used. A binary
closed-against-open class fails differently: the natural way to write its rule,
that a domain is closed if a loop has been closed in it, makes the class identical
to a non-zero loop count, so any separation it produces restates its own
definition. The rung avoids both. It is read off a benchmark's verification
method, which is a fact about that benchmark rather than a judgement about the
domain, and being ordinal it carries the gradation an affordance percentage would
have supplied without inventing the percentage.

\subhead{Scope of the rung} It is not a statement about a domain's
entire published output. A rung-5 domain may contain machine-checkable subtasks,
and law is the example: one loop has been closed there, on a narrow
charge-classification slice that is rung~1 work inside a rung~5 discipline. The
rung describes what closing a loop on the domain's \emph{practice} would have to
check against. Under it the relation is strong but not perfect:
three of the twenty-two cross the automatic/non-automatic line, law
and robotic manipulation with one loop each above it and engineering design with
none below it.

\subhead{Robustness to sample doubling} The census covers $55$ closed
loops over $22$ domains, against $22$ loops over $18$ domains for the $2024$
benchmark vintage alone, and the relation is unchanged at the larger sample. Loops at rungs~1 to~3 stand at $53$ of $55$ ($96.4\%$) against $20$ of $22$ ($90.9\%$) on the smaller sample. Not one of the $33$ later loops sits at rung~4 or~5: $23$ are rung~1 and $10$ are rung~2. The eight domains with no loop on the smaller sample still have none on the larger. Spearman is $-0.75$ over the $22$
domains against $-0.79$ over the $18$.

\subhead{Stability of the standing-loss set} \Cref{fig:landscape}(d)
marks the five domains at rung~5 whose capability is above the threshold, and all
five sit where the smaller sample put them: economics at $80.8$, health and clinical
at $72.1$, philosophy and history both at $70.1$, and law at $67.8$, with no
loops except law's one narrow slice. That is the substantive result for this panel,
because the four testbed families are all rung~1 or~2 and every one of them falls in
the where-loops-close half. Thirty-three further loops, and not one of them in this
region.

\subhead{Capability is non-monotonic in the rung, and the panel needs it to be}
Otherwise the standing loss would just be a restatement of where models are weakest.
Mean capability by rung runs $91.6$, $55.0$, $58.6$, $32.6$ and $72.2$, which is not
ordered, and the individual comparisons are sharper still: economics sits at $80.8$ on
rung~5 while ML research replication sits at $64.4$ on rung~3, engineering design at
$55.0$ on rung~2 and single-cell genomics at $52.8$ on rung~3. So there are rung-5
domains models handle better than rung-2 and rung-3 domains where loops have closed
or could, which is what makes the standing loss a statement about verification
rather than about competence.

\subhead{Recall against practice} Where a domain has both a recall or exam benchmark
and a practice benchmark, this table carries the one the figures plot and
\Cref{fig:landscape}(e) carries the pair. For the standing-loss domains that is a
recall benchmark, because it is the only measurement those domains have, and
panel~(e) exists to qualify exactly those values. Every endpoint of every pair, with
its model, its verification method and its source, is in \Cref{app:pairs}.

\subsection{Recall and Practice Pairs in \Cref{fig:landscape}(e)}
\label{app:pairs}

Panel~(e) pairs, for each of six domains, a recall or exam benchmark against a
benchmark of the domain's own practice. \Cref{tab:pairs} gives both endpoints of
every pair. Reading it is the only way to see what the panel's headline span of
$35$ to $83$ points is and is not.

\subhead{Only one pair is measured within a system} MLE-bench Lite against
MLE-bench High is the same agent on the same leaderboard, so its $38.1$ points is a
within-model delta. The other five take the best publicly reported score at each
end independently, and in most cases those are different models. Those five gaps
are therefore between \emph{benchmark frontiers}, not within a model, and they
should be read as indicative of the format gap rather than as causal estimates.
The figure labels every endpoint with the system that produced it for this reason.

\subhead{High split, not the full benchmark} That pair's practice end is the \textbf{High}
split, and saying so matters because the report uses the full benchmark elsewhere
for a different purpose. The leaderboard gives this agent Lite $80.3\pm1.52$,
Medium $64.04\pm2.32$, High $42.22\pm2.22$ and All $64.44\pm1.18$. The pair
contrasts the easiest split with the hardest; the All figure, $64.4$, is the one
\Cref{tab:domains} carries as this domain's capability, so the two tables use two
different splits of one benchmark and each names the split it uses.

\subhead{Four endpoints carry a caveat that the number alone does not} The
humanities practice end, $5.2$, is o3-mini~(high) from the HLE paper's own Table~3,
and it is the oldest number in the panel. Frontier models now reach about $55$ on
HLE \emph{overall} closed-book and about $65$ where tools are permitted, so the
frontier gap in this domain is materially smaller than the $64.9$ points plotted,
though still among the largest here. \textbf{No per-category humanities score is
published} on any of the boards that report the overall figure, so the panel keeps
the early measurement and labels it rather than substituting an estimate for it.
The research-mathematics practice end, ${\sim}0$, is not a
leaderboard score at all: FrontierMath's Open Problems tier requires an original
proof accepted by review, and the figure records that no frontier model is credited
with solving one as of the audit, which is a community observation. And the wet-lab
pair spans two different benchmarks, protocol reasoning graded by rubric against
protocol generation scored by an automatic text metric, because no single benchmark
covers both ends; that pair says understanding a protocol and writing a correct one
are different capabilities, which is weaker than a like-for-like comparison.

\subhead{The gap is difficulty and strictness together} Every recall endpoint is
machine-checkable: multiple choice, exact match, an executable suite, or a computed
competition metric. The practice endpoints are a mix, expert rubric for HLE and
AgentClinic, a weighted rubric for BenchBench-Protocol, an automatic soft metric for
BioProBench, review for Open Problems, and executable tests on harder instances for
SWE-bench Pro and MLE-bench High. So the span measures task difficulty and
verification strictness at once, and does not separate them.

\begin{table}[t]
\centering
\resizebox{\linewidth}{!}{%
\small
\renewcommand{\arraystretch}{1.16}
\setlength{\tabcolsep}{5pt}
\begin{tabular}{l l l c l l}
\toprule
\headrow
\textbf{Domain} & \textbf{End} & \textbf{Benchmark} & \textbf{Score} & \textbf{Verified by} & \textbf{Model / system} \\
\midrule
\multirow{2}{*}{Research maths} & recall & FrontierMath v2 Tier~4~\citep{glazer2024frontiermath} & $83.0$ & numeric exact match & GPT-5.6 Sol \\
 & practice & FrontierMath Open Problems & ${\sim}0$ & proof, accepted by review & all frontier models \\
\midrule
\multirow{2}{*}{Humanities} & recall & MMLU-Pro history~\citep{wang2024mmlupro} & $70.1$ & multiple choice & GPT-4o \\
 & practice & HLE humanities~\citep{phan2025hle} & $5.2$ & expert rubric & o3-mini (high) \\
\midrule
\multirow{2}{*}{Wet-lab} & recall & BenchBench-Protocol & $59.2$ & weighted rubric & Claude Opus 5 \\
 & practice & BioProBench generation & ${<}15$ & automatic text metric & frontier models \\
\midrule
\multirow{2}{*}{ML research} & recall & MLE-bench Lite~\citep{chan2025mlebench} & $80.3$ & computed competition metric & Famou-Agent 2.0 \\
 & practice & MLE-bench High~\citep{chan2025mlebench} & $42.2$ & computed competition metric & \textbf{the same agent} \\
\midrule
\multirow{2}{*}{Software eng.} & recall & SWE-bench Verified~\citep{jimenez2024swebench} & $97.0$ & executable test suite & Claude Opus 5 \\
 & practice & SWE-bench Pro~\citep{scale2026swebenchpro} & $59.1$ & executable test suite & GPT-5.4 (xHigh) \\
\midrule
\multirow{2}{*}{Clinical} & recall & MedQA (USMLE)~\citep{jin2020medqa} & $97.4$ & multiple choice & Gemini 3.1 Pro \\
 & practice & AgentClinic~\citep{schmidgall2026agentclinic} & $62.1$ & rubric on a simulated encounter & Claude 3.5 Sonnet \\
\bottomrule
\end{tabular}}
\caption{\textbf{Provenance of the six recall-against-practice pairs of
\Cref{fig:landscape}(e).} Scores are the best publicly reported for each benchmark
as of the August 2026 audit. A benchmark named without a citation has no
peer-reviewed paper behind it and is a leaderboard or an industry report, so it is
named in place rather than entered in the bibliography. \textbf{Only the ML research
pair is measured within one system}, both ends being the same agent on the same
leaderboard; the other five compare each end's own frontier. \textbf{SWE-bench Pro
is quoted in two non-comparable families and the standardized one is used}: Scale's
SEAL board runs every model through one harness and its public-split leader is
$59.1$, while vendor-scaffold aggregates run $15$ to $30$ points higher and reach
about $80$. The recall end of that pair is itself measured on a minimal bash-only
harness, so the standardized figure is the one that matches it. The clinical pair, at
$35.3$ points, is the narrowest here.}\label{tab:pairs}
\end{table}

\begin{table}[t]
\centering
\resizebox{\linewidth}{!}{%
\small
\renewcommand{\arraystretch}{1.14}
\setlength{\tabcolsep}{5pt}
\begin{tabular}{l c c c l}
\toprule
\headrow
\textbf{Domain} & \textbf{Rung} & \textbf{Loops} & \textbf{Cap.\ (\%)} & \textbf{Capability read off} \\
\midrule
Software engineering    & 1 & 9 & 97.0 & SWE-bench Verified, Claude Opus 5~\citep{jimenez2024swebench} \\
\bandrow
Competitive programming & 1 & 8 & 93.2 & LiveCodeBench v6, Sakana Fugu-Ultra~\citep{jain2024livecodebench} \\
Olympiad mathematics    & 1 & 7 & 96.1 & OTIS Mock AIME, GPT-5.4 Pro \\
\bandrow
Graduate science QA     & 1 & 5 & 95.5 & GPQA-Diamond, Sakana Fugu-Ultra~\citep{rein2023gpqa} \\
Terminal operations     & 1 & 5 & 84.6 & Terminal-Bench 2.1, Claude Opus 5~\citep{terminalbench2025} \\
\bandrow
General assistant       & 1 & 4 & --$^{\S}$ & agentic testbed, no subject accuracy \\
Research-level maths    & 1 & 1 & 83.0 & FrontierMath v2 Tier 4, GPT-5.6 Sol~\citep{glazer2024frontiermath} \\
\bandrow
Multi-hop QA            & 1 & 1 & --$^{\S}$ & agentic testbed, no subject accuracy \\
Deep research           & 1 & 1 & --$^{\S}$ & agentic testbed, no subject accuracy \\
\bandrow
Embodied / interactive  & 2 & 10 & --$^{\S}$ & agentic testbed, no subject accuracy \\
Engineering design      & 2 & 0 & 55.0$^{\dagger}$ & MMLU-Pro engineering, GPT-4o~\citep{wang2024mmlupro} \\
\bandrow
ML research replication & 3 & 1 & 64.4 & full MLE-bench, Famou-Agent 2.0~\citep{chan2025mlebench} \\
Single-cell genomics    & 3 & 1 & 52.8$^{\dagger}$ & scBench, Claude Opus 4.6 \\
\midrule
Robotic manipulation    & 4 & 1 & 12.8$^{\dagger}$ & RoboDojo, real-world split \\
\bandrow
Wet-lab protocol        & 4 & 0 & 59.2$^{\dagger}$ & BenchBench-Protocol, Claude Opus 5 \\
Materials science       & 4 & 0 & 25.8 & PhononBench stability, six-model mean \\
\bandrow
Law (jurisdiction)      & 5 & 1 & 67.8 & Realm Legal, Claude Opus 5 \\
History                 & 5 & 0 & 70.1$^{\dagger}$ & MMLU-Pro history, GPT-4o~\citep{wang2024mmlupro} \\
\bandrow
Health / clinical       & 5 & 0 & 72.1 & MMLU-Pro health, GPT-4o~\citep{wang2024mmlupro} \\
Philosophy              & 5 & 0 & 70.1 & MMLU-Pro philosophy, GPT-4o~\citep{wang2024mmlupro} \\
\bandrow
Economics               & 5 & 0 & 80.8 & MMLU-Pro economics, GPT-4o~\citep{wang2024mmlupro} \\
Classical languages     & 5 & 0 & --$^{\S}$ & HLE classics split, practice-only~\citep{phan2025hle} \\
\bottomrule
\end{tabular}}
\caption{\textbf{The twenty-two-domain audit, and the source of every number in
\Cref{fig:landscape}.} \textbf{Loops} counts the systems of \Cref{tab:census} whose
closing benchmark falls in the domain, and is derived from that table rather than
typed here, so the two cannot disagree; it totals $55$ over $14$ domains.
\textbf{Rung} is the verifier rung of \Cref{tab:verifier}, assigned from
the verification method of the domain's \emph{practice} benchmark and never from its
loop count; every assignment is justified above. \textbf{Cap.} is the best publicly
reported score on the named benchmark as of August 2026: $\dagger$ read from the
benchmark paper's own results table, unmarked a public leaderboard or industry
report, and $\S$ \textbf{no value on this axis}: an agentic testbed family, which
reports a task success rate, or a domain whose only published score is a practice
benchmark. Rows are
ordered by rung and the rule falls where verification stops being automatic. Over
all $22$ domains loop count tracks the rung at Spearman $\rho=-0.75$
($p<0.001$, Pearson $-0.63$, Kendall $-0.62$); over the $17$ that have a capability
it tracks capability at Pearson $r=+0.64$ ($p=0.006$). \textbf{The two are not
interchangeable}: on those $17$, controlling for capability the partial correlation
of loops with rung is $-0.67$ ($p=0.005$), while controlling for rung the partial
correlation with capability is $+0.45$ ($p=0.08$, short of significance). Mean
capability is non-monotonic in the rung, $91.6$, $55.0$, $58.6$, $32.6$ and
$72.2$.}\label{tab:domains}
\end{table}

\section{Operator Cards and Optimization Contracts}
\label{app:contracts}

Every operator publishes a machine-readable \emph{operator card} that the \rsiag reads when planning, and an \emph{optimization contract} that the vertical \rsisub reads when rewriting the operator's policy. The two are distinct: the card describes what the operator does and when it should be chosen, and is partly mutable; the contract describes what may be changed about the operator at all, and its structure is fixed. \Cref{tab:contracts} lists the three surface classes for each operator.

\begin{table}[t]
\centering
\resizebox{\linewidth}{!}{%
\renewcommand{\arraystretch}{1.3}
\setlength{\tabcolsep}{5pt}
\begin{tabular}{l l l l}
\toprule
\headrow
\textbf{Component} & \textbf{Mutable surfaces} & \textbf{Action surfaces} & \textbf{Protected surfaces} \\
\midrule
\dcell{\dataop} &
\makecell[l]{diagnosis \& synthesis instructions\\ scheduler weights $\alpha,\beta$\\ curriculum stage ratios\\ card guidance} &
\makecell[l]{directive policy\\ question policy\\ curriculum policy} &
\makecell[l]{sealed set \& evaluator\\ record/provenance schema\\ evidence chain\\ verifier, contamination checker} \\
\midrule
\hcell{\harnessop} &
\makecell[l]{patch-proposal instructions\\ shard budget \& candidate count\\ promotion strictness\\ card guidance} &
\makecell[l]{genome prompt\\ genome tools, skills \& MCP\\ genome policies (incl.\ memory)} &
\makecell[l]{runtime \& agent loop\\ provider, target model\\ evaluator, sandbox, release gate\\ raw sealed item text} \\
\midrule
\mcell{\modelop} &
\makecell[l]{recipe-proposal instructions\\ candidate count\\ stopping heuristics\\ card guidance} &
\makecell[l]{adaptation recipe\\ optimization schedule\\ candidate policy} &
\makecell[l]{training code \& backend\\ dataset content \& split\\ evaluator, reward, release rule} \\
\midrule
\textcolor{CtrlC}{\textbf{\rsiag}} &
\makecell[l]{diagnosis instructions\\ routing preferences\\ budget \& stopping policy\\ transition guidance} &
\makecell[l]{horizontal program\\ vertical directive\\ transition exception} &
\makecell[l]{operator identities \& types\\ adapter endpoints\\ evaluator, gate, ledger} \\
\bottomrule
\end{tabular}}
\caption{\textbf{Optimization contracts.} \emph{Mutable} surfaces may be rewritten by the vertical \rsisub of \Cref{eq:vertical}; \emph{action} surfaces are what the operator may write on the target system; \emph{protected} surfaces are outside every write surface at every level, including the meta layer.}\label{tab:contracts}
\end{table}

An update is accepted only when four deterministic checks pass: the contract identifier in the response matches the contract that was issued; the surfaces the sub-agent declares it modified are a subset of those the directive requested; those requested are a subset of the contract's mutable set; and the declared modifications equal the actual textual diff between the previous and proposed policy. The fourth check is what prevents a sub-agent from smuggling an undeclared edit alongside a declared one, and it is the reason policies are stored as structured documents rather than as free text.

\section{Transition Adapter Specifications}
\label{app:adapters}

Each admissible edge of \Cref{fig:transitions} is realized by an adapter with a fixed input type, a fixed output type, and a bounded model call. Adapters may reformat, distill, and reject; they may not modify the target system.

\noindent $\opD\!\to\!\opH$ \textbf{Experience Extraction.} Input: verified records, the trajectories that produced them, and the current genome. The adapter clusters records by failure signature, distills each cluster into at most one typed memory entry or skill descriptor, deduplicates against the existing genome by normalized content hash and $n$-gram overlap, and enforces a hard cap on total entries so that the scaffold cannot grow monotonically. Output: an \textsc{Experience} artifact consumable by the harness patcher.

\noindent $\opD\!\to\!\opM$ \textbf{Dataset Materialization.} Input: a \textsc{Dataset} artifact with its curriculum staging and per-record lineage. The adapter resolves the curriculum into physical shards, applies a loss mask that credits only assistant-generated positions, verifies that the requested recipe is expressible by the declared training backend, and estimates token and step counts against the remaining budget before submission. Output: materialized shards plus a validated recipe request.

\noindent $\opH\!\to\!\opD$ \textbf{Signal Recompilation.} Input: a promoted genome. The adapter re-runs the fixed evaluator on the adaptation split under the new genome and recompiles $\sig$, annotating each failure with whether it persisted, newly appeared, or was resolved by the harness change. Output: a refreshed learning signal with harness-attributed deltas, which is what makes the subsequent synthesis measure the system as it now is rather than as it was.

\noindent $\opM\!\to\!\opD$ \textbf{Signal Recompilation, under new weights.} Input: a released checkpoint and the previous learning signal. The adapter re-runs the fixed evaluator on the same probe set under the new weights and recompiles $\sig$ together with the learning signatures $\kappa$ of \Cref{eq:kappa}, reporting per-dimension movement including movement in the wrong direction. It is the same function as the $\opH\!\to\!\opD$ adapter above, triggered by the other capability-altering step. Output: a refreshed learning signal with weight-attributed deltas and a regression list.

\noindent $\opM\!\to\!\opH$ \textbf{Redundancy Reconciliation.} Input: a training report and the current genome. For each scaffold entry, the adapter identifies the failure signature it was introduced to repair, checks whether that signature still fires on the same episodes after training, and proposes a deletion when it does not; symmetrically, it proposes additions where new weight capability makes a previously unusable tool or skill worthwhile. Every proposed deletion is replayed before promotion, so a deletion that costs accuracy is reverted. Output: a typed \textsc{HarnessPatch} of deletions and additions.

\section{Failure Signature Taxonomies}
\label{app:signatures}

The first two fields of \Cref{eq:signature} are assigned by rules rather than by a model, and the rules differ by task family. Both taxonomies share a priority order, so that a failure with several symptoms is attributed to its earliest cause rather than to its most visible one.

\noindent \textbf{Closed-form track.} \texttt{output\_protocol} (the response violates the declared answer format) $>$ \texttt{abstention} (no answer emitted within budget) $>$ \texttt{reasoning\_answer\_mismatch} (the derivation supports a different answer than the one emitted) $>$ \texttt{arithmetic\_error} $>$ \texttt{distractor\_confusion} (the emitted answer matches a known distractor and the derivation engages with it) $>$ \texttt{none}.

\noindent \textbf{Executable track.} Turn-level causes are ordered \texttt{invalid\_tool\_call} $>$ \texttt{argument\_mismatch} $>$ \texttt{state\_mismatch} $>$ \texttt{recovery\_failure} $>$ \texttt{missing\_tool\_call} $>$ \texttt{response\_mismatch}; task-level causes are \texttt{missing\_validation} (the system modified state without verifying), \texttt{unlimited\_exploration} (repeated ineffective commands beyond a threshold), \texttt{missing\_artifact}, \texttt{wrong\_scope}, \texttt{premature\_finish}, and \texttt{environment\_error}, the last of which is excluded from improvement targeting because it is not a property of the system.

Only the third field of \Cref{eq:signature}, the reusable mechanism behind the failure, is model-attributed, and it is required to be grounded in the deterministic fields: a mechanism that contradicts the assigned terminal cause is rejected at validation.

\section{The Three Record Gates}
\label{app:gates}

\Cref{alg:gates} states the gating procedure applied to every candidate training record in \dataop. The gates are fail-closed: a record that cannot be positively verified is rejected rather than admitted with lower weight.

\begin{algorithm}[h]
\caption{Record gating in \dataop}
\label{alg:gates}
\begin{algorithmic}[1]
\Require candidate record $x$, capability slot $c$, released checkpoint $\theta_{\mathrm{rel}}$, sealed corpus $\mathcal{Z}$
\Statex \textbf{Gate 1: Correctness}
\If{task family is closed-form}
  \State independent solver re-derives the answer without seeing $x$'s label; reject on disagreement
  \State a separate verifier re-checks label, format, and derivation consistency; reject on failure
\Else
  \State build the environment; \textbf{reject} unless it constructs and its tests execute
  \State \textbf{reject} unless tests pass on the reference solution \emph{and} fail on a no-op solution
  \State independent solver attempts $x$ from a clean state; \textbf{reject} if unsolvable within bound
\EndIf
\Statex \textbf{Gate 2: Novelty}
\State pre-solve $x$ with $\theta_{\mathrm{rel}}$ under a cheap decoding budget
\If{solved} \State mark \textsc{already-held}; route to the consolidation stage; exclude from value attribution \EndIf
\Statex \textbf{Gate 3: Provenance}
\State \textbf{reject} if $x$ matches $\mathcal{Z}$ under normalized exact match or high $n$-gram overlap
\State \textbf{reject} if a semantic near-duplicate of any $z\in\mathcal{Z}$ is found (fail-closed on uncertainty)
\State \textbf{reject} unless $x$ carries either a source failure signature or a declared external source reference
\State \textbf{reject} if $x$ is derived from artifacts of the diagnostic evaluation run itself
\State \Return admitted record with lineage $(\text{signature},\text{directive},\text{gate verdicts},\text{cost})$
\end{algorithmic}
\end{algorithm}

\section{Run Layout and Accounting Conventions}
\label{app:impl}

This appendix fixes the run-layout and accounting conventions used to re-derive every reported gain.

\noindent \textbf{Run layout.} A term occupies one directory keyed by benchmark, run identifier, and generation. Inside it, each operator step, each candidate, and each evaluation pass has its own subdirectory holding the inputs it was given, the raw model exchanges, the emitted artifact, and the realized cost. Candidate directories are never reused, so a failed or partially completed candidate cannot contaminate a sibling, and a run can be inspected after the fact without replaying it.

\noindent \textbf{Resumption.} The improvement loop is itself checkpointed, not only the training jobs inside it. Progress state, the accumulated experience pool, and the current work unit are written after every completed step, so a term interrupted mid-sequence resumes at the next step rather than restarting. Resumed runs continue against the same ledger rather than a fresh one, under the budget protocol of \Cref{sec:exp:setup}.

\noindent \textbf{Fixed-base cumulative training.} Every training candidate starts from the same base checkpoint over the cumulative dataset rather than continuing from the previous adapter. This costs redundant computation and buys round-to-round comparability: two generations differ in their data, not in their optimization history, so a difference in outcome is attributable to the data the framework produced.

\noindent \textbf{Accounting conventions.} Four conventions are fixed in advance. An over-budget response is scored as a failure rather than truncated and re-scored. Baseline and final systems are both scored pass@1 from a single decode under one identical, fixed decoding setting. The improvement budget and the meta-training budget are metered on separate ledgers and reported separately. The sealed split is opened once, after the system is frozen, and no intermediate decision in the term is conditioned on it.

\noindent \textbf{Seeds.} Each condition is run under five independent outer seeds; a seed governs the whole term (operator sampling, candidate ordering, and training initialization), not only the final training job.

\noindent \textbf{GPQA-D-hard100 index.} \Cref{tab:hard100} lists the official GPQA Record ID of every item in GPQA-D-hard100, build \texttt{gpqa-diamond-hard100-v1} under selection seed \texttt{rsi2-gpqa-hard100-002}. The Record ID is GPQA-Diamond's unique question identifier, so each row retrieves the exact item in the pinned Diamond release. The subset is the same under every condition and every outer seed.

\begin{table}[t]
\centering
\footnotesize
\setlength{\tabcolsep}{4pt}
\renewcommand{\arraystretch}{1.06}
\begin{tabular}{rl>{\hspace{14pt}}rl>{\hspace{14pt}}rl>{\hspace{14pt}}rl}
\toprule
\headrow
\textbf{\#} & \textbf{Record ID} & \textbf{\#} & \textbf{Record ID} & \textbf{\#} & \textbf{Record ID} & \textbf{\#} & \textbf{Record ID} \\
\midrule
  1 & \texttt{rec0VuKUjt1SZ7NYv} & 26 & \texttt{recDDxpS9s8cwkqfq} & 51 & \texttt{recmI7EiLv72PxmYK} & 76 & \texttt{rect4iLrSfUwkNTno} \\
  2 & \texttt{rec0yTRmO1o1xCA6H} & 27 & \texttt{recdya6FuYraBU5Rh} & 52 & \texttt{recMicVBcqy1xM1jq} & 77 & \texttt{rectXfsCM1dj4Kv2c} \\
  3 & \texttt{rec1oj2DveQWl9Rpw} & 28 & \texttt{recDytVnNYZe2HuUU} & 53 & \texttt{recn3NhOhqAPLda16} & 78 & \texttt{recUc29lMDBEvurYo} \\
  4 & \texttt{rec260hNUCEj109Dj} & 29 & \texttt{recE2ihVfqEK4R9d0} & 54 & \texttt{recN4DY9Q5V03glmQ} & 79 & \texttt{recUOePh79cp4T2Bg} \\
  5 & \texttt{rec4L69T0Y1AS4AFS} & 30 & \texttt{recEmTBhx2hgw6tPQ} & 55 & \texttt{recnGEpF1srQpaqWq} & 80 & \texttt{recuyeuT5rQ6qDt8F} \\
  6 & \texttt{rec527dNeEtWJrYNl} & 31 & \texttt{recf6ayQmL1SxKbvW} & 56 & \texttt{recnjViFrqlZNL3fY} & 81 & \texttt{recV1nqYQvpII94oC} \\
  7 & \texttt{rec5rjeLsEq5Fg7Oj} & 32 & \texttt{recfTlTMjZBuOducT} & 57 & \texttt{recnTTKdBzfuoZ7w7} & 82 & \texttt{recVE8cUNHpHZIAvL} \\
  8 & \texttt{rec7qmSnbud4FHSqL} & 33 & \texttt{recGee5m84dg5FZkc} & 58 & \texttt{recNu3MXkvWUzHZr9} & 83 & \texttt{recVvpD8miVjmmyfe} \\
  9 & \texttt{rec8nshandHARTkrg} & 34 & \texttt{recGFNRVl1qBZGwyU} & 59 & \texttt{recNuT2oSnO86bxOx} & 84 & \texttt{recwW1A85nfyQpReG} \\
  10 & \texttt{rec8y3ZrBOcLgNEkE} & 35 & \texttt{recgXxEgllSGEpELP} & 60 & \texttt{recO3hvCWRGiG0odN} & 85 & \texttt{recWXwn9v4IG9ZrM6} \\
  11 & \texttt{rec9ubQihAh6g9bft} & 36 & \texttt{rechgQucGlrnt8KRV} & 61 & \texttt{recooG6bivTUJpDBz} & 86 & \texttt{recXsuOHRBLcyenF2} \\
  12 & \texttt{rec9W28HgpEUeUN8k} & 37 & \texttt{recI1ls9OXdxatHQn} & 62 & \texttt{recOvqPKUtyy9ISA1} & 87 & \texttt{recXsYa3i2UhGF5fe} \\
  13 & \texttt{recA1i5ZAh0Uzclxp} & 38 & \texttt{recihePFulRgNKsIn} & 63 & \texttt{recoy9ZLBsc7HIRBy} & 88 & \texttt{recXvQ6gWAmyakrpD} \\
  14 & \texttt{reca44yABeO2fx7UB} & 39 & \texttt{recINGR1z01Fh1Z3A} & 64 & \texttt{recOYsaYs6RmtlTDy} & 89 & \texttt{recYA3LPsCvF1fTMI} \\
  15 & \texttt{recAAJoHMW45Lv5je} & 40 & \texttt{recIOlKBsOeEcgkA1} & 65 & \texttt{recPIzpnuYpB4yvmp} & 90 & \texttt{recyl3usDqb7ruXJx} \\
  16 & \texttt{recaXdGn3FAIkxLGM} & 41 & \texttt{recixxJmdux0d8LZQ} & 66 & \texttt{recpki12iG9RUGrz9} & 91 & \texttt{recYOzCsevNz61Lyn} \\
  17 & \texttt{recAYkd96NNuNl1Ei} & 42 & \texttt{recjJ54TXc04enRkZ} & 67 & \texttt{recPzW1WqRnPs57D6} & 92 & \texttt{recypVp2NmPlBKVTp} \\
  18 & \texttt{recb2M22zaD3tL6Qc} & 43 & \texttt{recJpyGtGIsxulevT} & 68 & \texttt{recqGD3fxPCI59vPQ} & 93 & \texttt{recYt8xx80OTyDsL0} \\
  19 & \texttt{recb4cGsC6BJUCU3V} & 44 & \texttt{reck4G4xxv3YnpbtQ} & 69 & \texttt{recr3VHM4zYf6dMFY} & 94 & \texttt{recZ13cwgDQf9jRd9} \\
  20 & \texttt{recb80OwMgNnceA9t} & 45 & \texttt{recK9F5aqdaybl8bb} & 70 & \texttt{recReg13iV2HwJTaA} & 95 & \texttt{reczjcMtrB1YGS2fO} \\
  21 & \texttt{recBtVK8rBVtIlXDq} & 46 & \texttt{reckEnrOPFT9Ru7tW} & 71 & \texttt{recRgabRzMaEoBRcM} & 96 & \texttt{reczkBiPPNrNN49Hp} \\
  22 & \texttt{recCJJOeBGERaHYax} & 47 & \texttt{recKm6LNWykGapmCr} & 72 & \texttt{recrNbtgTNoabJJi6} & 97 & \texttt{recZSGUkn56v9kEp1} \\
  23 & \texttt{recclFbsjbaiVVnnV} & 48 & \texttt{recl1UtgTVKishAq4} & 73 & \texttt{recs3PLPUEMiqg4P8} & 98 & \texttt{reczUoM8JsxU6pYxr} \\
  24 & \texttt{reccOKzFNmyqeJ6ry} & 49 & \texttt{recL9MFV5zmdlle5T} & 74 & \texttt{recS48OsU6kVadBPW} & 99 & \texttt{recZWeueB7lSPR6wN} \\
  25 & \texttt{reccVBrYdwsB84fGy} & 50 & \texttt{recLB0EkQ54bYVhnd} & 75 & \texttt{recSBcGLPatKb3Ygu} & 100 & \texttt{reczzzihL7btBH7RO} \\
\bottomrule
\end{tabular}
\caption{\textbf{GPQA-D-hard100 index.} The official GPQA Record ID of each of the $100$ frozen items, numbered by ascending Record ID. The Record ID uniquely identifies a question in the GPQA-Diamond release.}\label{tab:hard100}
\end{table}